\documentclass[10pt,english]{article}

\usepackage[margin=1in]{geometry}
\usepackage{babel}

\usepackage[T1]{fontenc}    %
\usepackage[utf8]{inputenc} %
\usepackage{hyperref}       %
\hypersetup{colorlinks,linkcolor=red,anchorcolor=blue,citecolor=blue}
\usepackage{amsfonts}
\usepackage{nicefrac}
\usepackage{microtype}
\usepackage{verbatim}

\usepackage{float}
\usepackage{bm}
\usepackage{amsthm}
\usepackage{amsmath}
\usepackage{amssymb}
\usepackage{graphicx} 
\usepackage{xcolor,wrapfig}
\usepackage{cases}

\usepackage{algorithm}
\usepackage{algorithmic}
\usepackage{multirow}
\usepackage{arydshln}

\usepackage{pifont}%
\newcommand{\cmark}{\ding{51}}%
\newcommand{\xmark}{\ding{55}}%

\definecolor{yxc}{RGB}{255,0,0}
\definecolor{yjc}{RGB}{125,0,0}
\definecolor{cm}{RGB}{0,0,200}
\definecolor{yly}{RGB}{0,150,0}
\definecolor{dacong}{RGB}{58,148,190}

\allowdisplaybreaks

\newcommand{\paren}[1]{\left(#1\right)}
\newcommand{\innprod}[1]{\left\langle#1\right\rangle}

\newcommand{\ba}{\bm{a}}
\newcommand{\bb}{\bm{b}}

\newcommand{\be}{\bm{e}}

\newcommand{\bs}{\bm{s}}

\newcommand{\bu}{\bm{u}}
\newcommand{\bv}{\bm{v}}

\newcommand{\bx}{\bm{x}}
\newcommand{\by}{\bm{y}}
\newcommand{\bz}{\bm{z}}

\newcommand{\bA}{\bm{A}}

\newcommand{\bG}{\bm{G}}
\newcommand{\bH}{\bm{H}}
\newcommand{\bI}{\bm{I}}

\newcommand{\bW}{\bm{W}}

\newcommand{\cM}{\mathcal{M}}

\newcommand{\EE}{\mathbb{E}}

\newcommand{\RR}{\mathbb{R}}

\newcommand{\bxi}{\bm{\xi}}

\newcommand{\bSigma}{\bm{\Sigma}}

\newcommand{\dd}{\mathrm{d}}
\newcommand{\argmin}{\mathop{\mathrm{argmin}}}

\newcommand{\norm}[1]{\|#1\|}

\newcommand{\inner}[2]{\left\langle #1,#2 \right\rangle}

\newcommand{\onet}{\bm{1}_n \otimes}
\newcommand{\ave}{\Big(\frac{1}{n} \bm{1}_n^\top \otimes \bI_d\Big)}

\newcommand{\WK}{(\bW^K \otimes \bI_d)}
\newcommand{\x}{\bx}
\newcommand{\y}{\by}

\newcommand{\bbx}{\overline{\bx}}
\newcommand{\bby}{\overline{\by}}

\newcommand{\bbH}{\overline{\bH}}

\newcommand{\mean}{( \frac1n \bm 1_n \bm 1_n^\top) \otimes \bI_d}

\newtheorem{lemma}{\textbf{Lemma}}
\newtheorem{theorem}{\textbf{Theorem}}
\newtheorem{corollary}{\textbf{Corollary}}
\newtheorem{remark}{\textbf{Remark}}
\newtheorem{assumption}{\textbf{Assumption}}
\newtheorem{definition}{\textbf{Definition}}
 
\newcommand{\ex}[1]{\mathbb{E}\left[#1\right]}

\newcommand{\opt}{{\mathsf{opt}}}

\title{Communication-Efficient Distributed Optimization in Networks with Gradient Tracking and Variance Reduction\footnotetext{Preliminary results in this paper appeared at The 23rd International Conference on Artificial Intelligence and Statistics (AISTATS), 2020. }}

\date{}
\author
{
	Boyue Li\thanks{Department of Electrical and Computer Engineering, Carnegie Mellon University, Pittsburgh, PA 15213, USA; Email:
		\texttt{\{boyuel,shicongc,yuejiec\}@andrew.cmu.edu}.} 
	\qquad\qquad Shicong Cen\footnotemark[1]
	\qquad\qquad Yuxin Chen\thanks{Department of Electrical Engineering, Princeton University, Princeton, NJ 08544, USA; Email:
		\texttt{yuxin.chen@princeton.edu}.}
	\qquad\qquad Yuejie Chi\footnotemark[1]   %
	\smallskip\\
	Carnegie Mellon University\footnotemark[1]  \qquad\qquad Princeton University\footnotemark[2]
	}

\begin{document}

\maketitle

\begin{abstract}
	There is growing interest in large-scale machine learning and optimization over decentralized networks, e.g.~in the context of multi-agent learning and federated learning.
Due to the imminent need to alleviate the communication burden, the investigation of communication-efficient distributed optimization algorithms --- particularly for empirical risk minimization --- has flourished in recent years. A large fraction of these algorithms have been developed for the master/slave setting, relying on the presence of a central parameter server that can communicate with all agents. 
	
This paper focuses on distributed optimization over networks, or decentralized optimization, where each agent is only allowed to aggregate information from its neighbors over a network (namely, no centralized coordination is present). By properly adjusting the global gradient estimate via local averaging in conjunction with proper correction, we develop a communication-efficient approximate Newton-type method, called \texttt{Network-DANE},  which generalizes DANE 
to accommodate decentralized scenarios. Our key ideas can be applied, in a systematic manner, to obtain decentralized versions of other master/slave distributed algorithms. A notable development is \texttt{Network-SVRG/SARAH}, which employs variance reduction 
	at each agent to further accelerate local computation. We establish linear convergence of \texttt{Network-DANE} and \texttt{Network-SVRG} for strongly  convex losses, and \texttt{Network-SARAH} for quadratic losses, which shed light on the impacts of data homogeneity, network connectivity, and local averaging 	
upon the rate of convergence. We further extend \texttt{Network-DANE} to composite optimization by allowing a nonsmooth penalty term. Numerical evidence is provided to demonstrate the appealing performance of our algorithms over competitive baselines, in terms of both communication and computation efficiency. Our work suggests that by performing a judiciously chosen amount of local communication and computation per iteration, the overall efficiency can be substantially improved.
\end{abstract}
 

\section{Introduction}

Distributed optimization has been a classic topic \cite{bertsekas1989parallel}, yet is attracting significant attention recently in machine learning due to its numerous applications such as distributed training \cite{boyd2011distributed}, multi-agent learning \cite{nedic2010constrained}, and federated learning \cite{konevcny2015federated,konevcny2016federated,mcmahan2017communication}. At least two facts contribute towards this resurgence of interest: (1) the scale of modern datasets has oftentimes far exceeded the capacity of a single machine and requires coordination across multiple machines; (2) privacy and communication constraints disfavor information sharing in a centralized manner and necessitates distributed infrastructures. 

Broadly speaking, there are two distributed settings that have received wide interest: 1) the {\em master/slave} setting, which assumes the existence of a central parameter server that can perform information aggregation and sharing with all agents; and 2) the {\em network} setting --- also known as the {\em decentralized} setting --- where each agent is only permitted to communicate with its neighbors over a locally connected network (in other words, no centralized coordination is present). Developing fast-convergent algorithms for the latter setting is in general more challenging.

Many algorithms have been developed for the master/slave setting to improve communication efficiency, including deterministic algorithms such as one-shot parameter averaging \cite{zhang2012communication}, CoCoA \cite{smith2018cocoa}, DANE \cite{shamir2014communication}, CEASE \cite{fan2019communication}, and stochastic algorithms like distributed SGD \cite{recht2011hogwild} and distributed SVRG \cite{lee2017distributed,konevcny2015federated,cen2019convergence}. In comparison, the network setting is substantially less explored. Recent work \cite{lian2017can} suggested that the network setting can effectively avoid traffic jams during communication on busy nodes, e.g.~the parameter server, and be more efficient in wall-clock time than the master/slave setting. It is therefore natural to ask whether one can adapt more appealing algorithmic ideas to the network setting --- particularly for the kind of network topology with a high degree of locality --- without compromising the convergence guarantees attainable in the master/slave counterpart.

\subsection{Our Contributions}

In this paper, we investigate the problem of empirical risk minimization in the network (decentralized) setting, with the aim of achieving communication and computation efficiency simultaneously. The main algorithmic contribution of this paper is the development of communication-efficient network-decentralized (stochastic) optimization algorithms with primal-only formulations, with the assistance of proper gradient tracking. The proposed algorithmic ideas accommodate both approximate Newton-type methods and stochastic variance-reduced methods, and come with theoretical convergence guarantees.

\paragraph{Algorithmic developments.} We start by studying an approximate Newton-type method called DANE \cite{shamir2014communication}, which is among the most popular communication-efficient algorithms to solve empirical risk minimization. However, DANE was only designed for the master/slave setting in its original form. The current paper develops \texttt{Network-DANE}, which generalizes DANE to the network setting. The main challenge in developing such an algorithm is to track and adapt a faithful estimate of the global gradient at each agent, despite the lack of centralized information aggregation. Towards this end, we leverage the powerful idea of {\em dynamic average consensus} (originally proposed in the control literature \cite{zhu2010discrete} and later adopted in decentralized optimization \cite{qu2018harnessing,nedic2017achieving,di2016next}) to track and correct the locally aggregated gradients at each agent --- a scheme commonly referred to as {\em gradient tracking}. We then employ the corrected gradient in local computation, according to the subroutine adapted from DANE. This simple idea allows one to adapt approximate Newton-type methods to network-distributed optimization, without the need of communicating the Hessians.

Our ideas for designing \texttt{Network-DANE} can be extended, in a systematic manner, to obtain decentralized versions of other  algorithms developed for the master/slave setting, by modifying the local computation step properly. As a notable example, we develop \texttt{Network-SVRG}, which performs variance-reduced stochastic optimization locally to enable further  computational savings \cite{johnson2013accelerating}. The same approach can be applied to other distribute stochastic variance-reduced methods such as SARAH \cite{nguyen2017sarah} to obtain \texttt{Network-SARAH}. We also demonstrate that \texttt{Network-DANE} can be extended to the proximal setting for nonsmooth composite optimization in a straightforward manner.

\paragraph{Performance analysis.} The proposed algorithms achieve an intriguing trade-off between communication and computation efficiency. During every iteration, each agent only communicates the parameter and the gradient estimate to its neighbors, and is therefore communication-efficient globally; moreover, the local subproblems at each agent can be solved efficiently with accelerated or variance-reduced gradient methods, and is thus computation-efficient locally. When the network exhibits a high degree of locality, we show that by allowing multiple rounds of local mixing within each iteration, an improved overall communication complexity can be achieved as it accelerates the rate of convergence. Theoretically, we establish the linear convergence of \texttt{Network-DANE} for strongly convex losses, with an improved rate for quadratic losses, both with and without extra averaging. For \texttt{Network-SVRG}, we establish its linear convergence for the case of smooth strongly convex losses with extra rounds of averaging. Similar results are obtained for \texttt{Network-SARAH} for quadratic losses. Our analysis is highly nontrivial, as it needs to deal with the tight couplings of optimization and network consensus errors through a carefully-designed linear system of Lyapunov functions, especially in the context of approximate Newton-type methods which are known be harder to handle than simple gradient-type methods. Our results shed light on the impacts of data homogeneity and network connectivity upon the rate of convergence; in particular, the proposed algorithms provably obtain fast convergence if the local data are sufficiently similar. Table~\ref{table:summary} summarizes the convergence rates of the proposed algorithms.

All in all, our work suggests that: by performing a judiciously chosen amount of local communication and computation per iteration, the overall efficiency can be remarkably improved. Extensive numerical experiments are provided to corroborate our theoretical findings, and to demonstrate the practical efficacy of the proposed algorithms over competitive baselines.

\renewcommand{\arraystretch}{2}
\begin{table}[ht]
\centering
\begin{tabular}{|c|c|c|c|c|}
\hline
Algorithm                              & \begin{tabular}[c]{@{}c@{}}Communication \\ Rounds\end{tabular}                                                                                                 & \begin{tabular}[c]{@{}c@{}}Extra\\ Averaging\end{tabular} & \begin{tabular}[c]{@{}c@{}}Loss\\ Functions\end{tabular} & $\beta$                                 \\ \hline\hline
\multirow{4}{*}{\texttt{Network-DANE}} & $O \left( \frac{ \kappa (\beta / \sigma + 1) \log({1}/{\varepsilon}) }{ (1 - \alpha_0)^2 } \right)$                & \xmark                                                    & \multirow{2}{*}{Quadratic}                               & \multirow{4}{*}{Arbitrary}              \\ \cline{2-3}
                                       & $O \left( \log\kappa \cdot \frac{ (\beta^2/\sigma^2 + 1) \log(1/\varepsilon )} { (1 - \alpha_0)^{1/2} } \right)$   & \cmark                                                    &                                                          &                                         \\ \cline{2-4}
                                       & $O \left( \frac{ \kappa^2  \log({1}/{\varepsilon}) }{ (1 - \alpha_0)^2 } \right)$                                  & \xmark                                                    & \multirow{2}{*}{Strongly convex}                         &                                         \\ \cline{2-3}
                                       & $O \left( \log\kappa \cdot \frac{ \kappa (\beta/\sigma + 1) \log(1/\varepsilon )} { (1 - \alpha_0)^{1/2} } \right)$ & \cmark                                                    &                                                          &                                         \\ \hline
\texttt{Network-SVRG}                  & $O \left( \log\kappa \cdot \frac{ \log(1/\varepsilon)}{(1-\alpha_0)^{1/2}} \right)$                                 & \cmark                                                    & Strongly convex & \multirow{2}{*}{$\beta\leq \sigma/200$} \\ \cline{1-4}
\texttt{Network-SARAH}                 & $O \left( \log\kappa \cdot \frac{ \log(1/\varepsilon)}{(1-\alpha_0)^{1/2}} \right)$                                 & \cmark                                                    & Quadratic &                                         \\ \hline
 EXTRA                                 &    $O \left( \kappa^2 \log(1/\varepsilon) \right)$                                                                                                                 & \xmark                                                    &    Strongly convex                                                      & \multirow{2}{*}{Arbitrary}              \\ \cline{1-4}
DGD                                   &     $O\left(\frac{\kappa^2 \log(1/\varepsilon)}{(1-\alpha_0)^2}\right)$                                                                                                                & \xmark                                                    &   Strongly convex                                                       &                                         \\ \hline
\end{tabular}
\caption{Communication complexity of the proposed algorithms for quadratic and strongly convex losses to reach $\varepsilon$-accuracy. Here, $\sigma$, $L$ and $\kappa=L/\sigma$ are the strong convexity, smoothness, and condition number of the local loss functions $f_j$, $j=1,\ldots,n$, $\beta\leq L$ is the homogeneity parameter gauging the similarities of the local loss functions, and $\alpha_0 := \|\bm{W} -\tfrac{1}{n}\bm{1}_n\bm{1}_n^{\top}\|$ is the mixing rate over the network topology. Here, we assume the extra averaging step is implemented via the Chebyshev acceleration scheme \cite{arioli2014chebyshev}.
EXTRA \cite{shi2015extra} and DGD \cite{qu2018harnessing} are listed as baselines.
}   
	\label{table:summary}
\end{table} 
\renewcommand{\arraystretch}{1}

\subsection{Related Work}

First-order methods, which rely mainly on gradient information, are of core interest to big data analytics, due to their superior scalability. However, it is well-known that distributed gradient descent (DGD) suffers from a ``speed'' versus ``accuracy'' dilemma when na\"ively implemented in a decentralized setting \cite{nedic2018network}. Various fixes (see e.g.~the pioneering approaches such as EXTRA \cite{shi2015extra} and NEXT \cite{di2016next}) have been proposed to address this issue. Similar gradient tracking ideas \cite{zhu2010discrete} have been incorporated to adjust DGD to ensure its linear convergence using a constant step size \cite{nedic2017achieving,qu2018harnessing, li2019decentralized,xi2017add,yuan2018exact,scutari2019distributed,xin2019distributed}. The current paper is inspired by the use of gradient tracking in these early results. Our paper implements, and verifies the effectiveness of, gradient tracking for  algorithms that involve approximate Newton and variance reduction steps, which are far from straightforward and require significant efforts.   

\cite{scaman2017optimal} proposed a multi-step dual accelerated (MSDA) method for network-distributed optimization, which is optimal within a class of black-box procedures that satisfy the span assumption ---  the parameter updates fall in the span of the previous estimates and their gradients.   Further optimal algorithms are proposed in \cite{uribe2017optimal} and \cite{scaman2018optimal} for loss functions that are not necessarily convex or smooth. Their algorithms require knowledge of the dual formulation. In contrast, our algorithms are directly applied to the primal problem, which are more friendly for problems whose dual formulations are hard to obtain. Our algorithms also do not require the span assumption and therefore do not fall into the class of procedures studied in \cite{scaman2017optimal}. The recent work \cite{hannah2018breaking} suggested that algorithms that break the span assumption such as SVRG can be fundamentally faster than those that do not, and it is of future interest to study if similar conclusions hold in the distributed/decentralized setting.
  
The \texttt{Network-DANE} algorithm is closely related to DANE \cite{shamir2014communication}, which exhibits appealing performance  in both theory and practice. Another recent work further extended DANE with an additional proximal term in the objective function and strengthened its analysis \cite{fan2019communication}. The proposed \texttt{Network-DANE} adapts DANE to the network setting with the aid of gradient tracking. During the preparation of this paper, it was brought to our attention that the SONATA algorithm \cite{sun2019convergence}, which also applies gradient tracking and subsumes many existing algorithms as special cases with convergence guarantees, can be specialized to obtain the same local sub-problem studied in \texttt{Network-DANE}, up to different mixing approaches.   
 The connections between DANE and SVRG observed in \cite{konevcny2015federated} motivate the development of \texttt{Network-SVRG} in this paper, which can be viewed as implementing the local optimization of \texttt{Network-DANE} with variance-reduced stochastic gradient methods. The same idea can be easily applied to obtain network-distributed versions of other algorithms such as Katyusha \cite{allen2017katyusha}, GIANT \cite{wang2018giant}, AIDE \cite{reddi2016aide}, among others. Compared with decentralized SGD \cite{lan2017communication,lian2017can}, the proposed \texttt{Network-SVRG/SARAH} employ variance reduction to achieve much faster convergence. 
 
We note that variance-reduced methods have been adapted to the network setting recently in \cite{mokhtari2016dsa,yuan2018variance,xin2019variance,sun2019improving}; however, they either have a large memory complexity or impose substantial communication burdens. To be more specific, to decentralize SVRG-type algorithms, these papers \cite{yuan2018variance,xin2019variance,sun2019improving} all require communication at every step of the inner loop; in contrast, the proposed \texttt{Network-SVRG} algorithm only requires communication at the end of the inner loop, allowing each agent to perform the inner loop efficiently without synchronization, and is therefore  more communication-efficient.

\paragraph{Paper organization and notations.} Section~\ref{sec:background} introduces the formulation of distributed optimization in the decentralized setting, in addition to some preliminary facts. Section~\ref{sec:network_dane} presents the proposed \texttt{Network-DANE} together with its theoretical guarantees, and briefly discusses its extension to nonsmooth composite optimization. Section~\ref{sec:network_svrg} introduces \texttt{Network-SVRG/SARAH}, which invokes the variance reduction idea to further reduce local computation, together with their theoretical guarantees. We provide numerical experiments in Section~\ref{sec:numerical} and conclude in Section~\ref{sec:conclusions}. The details of the proofs are deferred to the appendix. Throughout this paper, we use boldface letters to represent vectors and matrices. In addition, $\|\bA\|$ denotes the spectral norm of a matrix $\bA$, $\|\ba\|_2$ represents the $\ell_2$ norm of a vector $\ba$, $\otimes$ stands for the Kronecker product, and $\bm{I}_n$ denotes the identity matrix of dimension $n$.

\section{Problem Formulation and Preliminaries} \label{sec:background}

\subsection{Network-Distributed Optimization}

Consider the following empirical risk minimization problem:
\begin{equation}
    \underset{{\bm{x}\,\in\, \mathbb{R}^d}}{\text{minimize}} \quad f(\x) \triangleq \frac{1}{N} \sum_{i=1}^{N} \ell(\bm{x}; \bm{z}_i),
\end{equation}
where $\bx \in \RR^d$ represents the parameter to optimize, $\ell(\bm{x}; \bm{z}_i)$ encodes certain empirical loss of $\bm{x}$ w.r.t.~the $i$th sample $\bm{z}_i$ and $N$ denotes the total number of samples we have available.
This paper primarily focuses on the case where the function $\ell(\cdot; \bm{z})$ is both convex and smooth for any given $\bm{z}$, although we shall also study nonconvex problems in numerical experiments.

In a decentralized optimization framework, data samples are distributed over $n$ agents. For simplicity, we assume throughout that data samples are split into disjoint subsets of equal size.  The $j$th local data set, represented by $\mathcal{M}_j$, thus contains $m  \triangleq N / n$ samples. 
As such, the global loss function can alternatively be represented by
\begin{equation}\label{eq:global_decom}
    f(\bm{x}) =  \frac{1}{n}\sum_{j=1}^n f_j(\bm{x}) , \qquad \text{with }~ f_j (\bm{x}) \triangleq \frac{1}{m} \sum_{\bm{z} \in \cM_j}\ell(\bm{x}; \bm{z}).
\end{equation}
Here, $f_j(\bm{x})$ denotes 
the local loss function at the $j$th agent $(1\leq j\leq n)$. In addition, there exists a network --- represented by an undirected graph $\mathcal{G}$ of $n$ nodes --- that captures the local connectivity across all agents. More specifically, each node in $\mathcal{G}$ represents an agent, and two agents are allowed to exchange information only if there is an edge connecting them in $\mathcal{G}$. Throughout this paper, we denote by $\mathcal{N}_j$ the set of all neighbors of the $j$th agent over $\mathcal{G}$. The goal is to minimize $f(\cdot)$ in a decentralized manner, subject to the aforementioned network-based communication constraints.

\subsection{Preliminaries}\label{sec:preliminary}

Before continuing, we find it helpful to introduce and explain two important concepts. 

\paragraph{Mixing.} Mathematically, the information mixing between neighboring nodes is often characterized by a mixing or gossiping matrix, denoted by $\bm{W} = [w_{ij}]_{1\leq i,j\leq n}\in\mathbb{R}^{n\times n}$.
More specifically, $w_{ij} = 0$ if agent $i$ and $j$ are not connected, and $\bW$ satisfies 
\begin{align}
	\bm{W}^{\top} \bm{1}_n=\bm{1}_n \qquad  \text{and} \qquad \bm{W}\bm{1}_n=\bm{1}_n,	\label{eq:mixing-matrix}
\end{align}
where $\bm{1}_n\in \mathbb{R}^n$ is the all-one vector. The spectral quantity, which we call the {\em mixing rate}, 
\begin{align}
    \alpha_0 \triangleq \|\bm{W} -\tfrac{1}{n}\bm{1}_n\bm{1}_n^{\top}\| \in [0,1)
    \label{eq:def_alpha0}
\end{align}
dictates how fast information mixes over the network.  As an example, in a fully-connected network, one can attain $\alpha_0=0$ by setting $\bm{W} = \tfrac{1}{n}\bm{1}_n\bm{1}_n^{\top}$. The paper \cite{nedic2018network} provides comprehensive bounds on $1/(1-\alpha_0)$ for various graphs. For instance, one has $\alpha_0 \asymp 1$ with high probability in an Erd\"os-R\'enyi random graph, as long as the graph is connected.

\paragraph{Dynamic average consensus.} 
Assume that each agent generates some {\em time-varying} quantity $r_j^{(t)}$ (e.g.~the current local parameter or gradient estimates). We are interested in tracking the dynamic average $$\tfrac{1}{n}\sum_{j=1}^n r_j^{(t)} =\tfrac{1}{n}\bm{1}_n^{\top} \bm{r}^{(t)}$$ in each of the agents, where $\bm{r}^{(t)}=[r_1^{(t)},\cdots, r_n^{(t)}]^\top$. To accomplish this, the paper \cite{zhu2010discrete} proposed a simple tracking algorithm: suppose each agent maintains an estimate $q_j^{(t)}$ in the $t$th iteration, and the network collectively adopts the following update rule
\begin{equation} \label{eq:dynamic_averaging}
\bm{q}^{(t)} = \bm{W} \bm{q}^{(t-1)}  + \bm{r}^{(t)} - \bm{r}^{(t-1)},
\end{equation}
where  $\bm{q}^{(t)} = [q_1^{(t)},\cdots, q_n^{(t)}]^\top$. 
The first term $\bm{W} \bm{q}^{(t-1)}$ represents the  standard local information mixing operation (meaning that each agent updates its own estimate by a weighted average of its neighbors' estimates), the second term  $\bm{r}^{(t)} - \bm{r}^{(t-1)}$ tracks the temporal difference. A crucial property of \eqref{eq:dynamic_averaging} is  
\begin{align}\label{eq:property_dac}
	\bm{1}_n^{\top}\bm{q}^{(t)} = \bm{1}_n^{\top}\bm{r}^{(t)} ,
\end{align}
which indicates that the average of $\{q^{(t)}_i\}_{1\leq i\leq n}$ dynamically tracks the average of  $\{r^{(t)}_i\}_{1\leq i\leq n}$. We shall adapt this procedure  in our algorithmic development, in the hope of reliably tracking the global gradients (i.e.~the average of the local, and often time-varying, gradients at all agents).

\section{\texttt{Network-DANE}: Algorithm and Convergence}
\label{sec:network_dane}

In this section, we propose an algorithm called \texttt{Network-DANE} (cf.~Alg.~\ref{alg:network_dane}),  which generalizes DANE \cite{shamir2014communication} to the network/decentralized setting. This is accomplished by carefully coordinating the information sharing mechanism and employing dynamic average consensus for gradient tracking.

\subsection{The DANE Algorithm} \label{sec:dane}

The DANE algorithm is a popular communication-efficient approximate Newton method developed for the master/slave model, initially proposed by \cite{shamir2014communication}. Here, we review some key features of DANE. (i) Each agent performs an update using both the local loss function $f_j(\cdot)$ and the gradient $\nabla f(\cdot)$ of the global loss function (obtained via the parameter server). (ii) In the $t$th iteration, the $j$th agent solves the following problem to update its local estimate $\bm{x}^{(t)}_j$:
\begin{equation} \label{eq:local_approximation}
	\bm{x}^{(t)}_j =  \underset{\bm{x}\,\in\, \mathbb{R}^d}{\arg \min}~~ \Big\{ f_j(\bm{x}) - \Big\langle \nabla f_j\big( \bbx^{(t)}\big) - \nabla  f\big(\bbx^{(t)}\big), \bm{x} \Big\rangle + \frac{\mu}{2} \big\| \bm{x} - \bbx^{(t)} \big\|_2^2 \Big\} ,
\end{equation}
where $\mu \geq 0$ is the regularization parameter.\footnote{In \cite{shamir2014communication}, the second term in \eqref{eq:local_approximation} takes the form $  \nabla f_j ( \bbx^{(t)} )  - \tilde{\eta} \nabla  f (\bbx^{(t)} )$. We set $\tilde{\eta}=1$ without loss of generality following the analysis in \cite{fan2019communication}.}
Implementing this algorithm requires two rounds of communications per iteration. 
\begin{itemize}
	\item[(a)] The parameter server first collects all local estimates $\{\bx_j^{(t-1)}\}_{1\leq j\leq n}$ and computes the average global parameter estimate $\bbx^{(t)} = \tfrac{1}{n} \sum_{j=1}^n \bm{x}_j^{(t-1)}$; this is then sent back to all agents; 
	\item[(b)] The parameter server collects all local gradients evaluated at the point $\bbx^{(t)}$, computes the global gradient $\nabla f(\bbx^{(t)}) = \tfrac{1}{n} \sum_{j=1}^n \nabla f_j(\bbx^{(t)})$, and shares it with all agents. 
\end{itemize}
The DANE algorithm has been demonstrated as a competitive baseline whose communication efficiency improves, in some sense, with the increase of data size \cite{shamir2014communication}; see \cite{fan2019communication} for its proximal variation and improved theoratical analysis. To see the reason why DANE is an approximate Newton-type algorithm, consider the case when the local loss functions in all agents are quadratic and takes the form 
\begin{align} \label{eq:quadratic-problem}
 f_j(\bm x) = \frac{1}{2}\bm{x}^{\top}  \bm{H}_j \bm{x} +  \bm{b}_j^{\top} \bm{x} + c_j, 
\end{align}
where each $\bm H_j = \nabla^2 f_j(\bm x) \in \mathbb{R}^{d\times d}$ 
is a fixed symmetric and positive semidefinite matrix. The local optimization subproblem \eqref{eq:local_approximation} in DANE can be solved in closed form, with $\bm{x}_j^{(t)}$ given by\footnote{See \cite{shamir2014communication} or Appendix~\ref{sec:derivation_dane} for a short derivation.} 
\begin{equation}
	\bm{x}_j^{(t)}   = \bbx^{(t)}-   \big( \underset{\text{local Hessian}}{\underbrace{ \bm H_j + \mu \bI_d }} \big)^{-1} \nabla  f\big(\bbx^{(t)}\big) . 
    \label{eq:xt-expression-quadratic}
\end{equation}     
Clearly, this can be interpreted as 
\begin{align*}
\bm{x}_{j}^{(t)}=\text{local parameter estimate}-\big(\text{local Hessian}\big)^{-1}\big(\text{global gradient}\big),	
\end{align*}
which is an approximate Newton-type update rule (since we invoke the local Hessian to approximate the true global Hessian).
It is worth noting that the algorithm proceeds without actually communicating the local Hessians.

\subsection{Algorithm Development} 

The DANE algorithm was originally developed for the master/slave setting. 
In the network setting, however, agents can no longer compute \eqref{eq:local_approximation} locally, due to the absence of centralization enabled by the parameter server; more specifically, agents have access to neither $\bbx^{(t)}$ nor $\nabla f(\bbx^{(t)})$, both of which are required when solving \eqref{eq:local_approximation}. To address this lack of global information, one might naturally wonder whether we can simply replace global averaging by local averaging; that is, replacing $\bbx^{(t)}$ and $\nabla f(\bbx^{(t)})$ by $\frac{1}{|\mathcal{N}_j|}\sum_{i\in \mathcal{N}_j} \bm{x}_i^{(t-1)}$ and $\frac{1}{|\mathcal{N}_j|} \sum_{i\in \mathcal{N}_j} \nabla f_i(\bm{x}_i^{(t-1)})$, respectively, in the $j$th agent. However, this simple idea fails to guarantee convergence in local agents. For instance, the local estimation errors may stay flat (but nonvanishing) --- as opposed to converging to zero --- as the iterations progress, primarily due to imperfect information sharing.

\begin{algorithm}[t]
\caption{\texttt{Network-DANE}}
\label{alg:network_dane}
\begin{algorithmic}[1]
    \STATE {\textbf{input:} initial parameter estimate $\bm{x}_j^{(0)}\in \mathbb{R}^d$ ($1\leq j\leq n$), regularization parameter $\mu$.}

    \STATE {\textbf{initialization:} set $\by^{(0)}_j = \bx^{(0)}_j$,
    $\bm{s}_j^{(0)} = \nabla f_j(\by_j^{(0)})$ for all agents $1\leq j\leq n$. }

    \FOR {$t = 1, 2, \cdots$}

        \FOR{\textbf{Agents} $1\leq j \leq n$ in parallel}

            \STATE {Set $\y_j^{(t), 0} = \x_j^{(t-1)}$ and $ \bs_j^{(t), 0} = \bs_j^{(t-1)}$.}
            \FOR {$k = 1, 2, \ldots, K$}
                \STATE{Receive  information $\y_i^{(t), k-1}$ and $\bs_i^{(t), k-1}$ from its neighbors $i \in \mathcal{N}_j$.}

                \STATE{Aggregate parameter estimates from neighbors:}
                \begin{equation}
                    \y_j^{(t), k} = \sum\nolimits_{i \in\mathcal{N}_j} w_{ji} \y_i^{(t), k-1}, \quad
                    \bs_j^{(t), k} = \sum\nolimits_{i \in\mathcal{N}_j} w_{ji} \bs_i^{(t), k-1}
                    \label{eq:network_dane_mixing}
                \end{equation} 
                \vspace{-1em}
        \ENDFOR

        \STATE {Set the local parameter estimate to $\y_j^{(t)} = \y_j^{(t), K}$.}

        \STATE{Update the global gradient estimate by aggregated local information and gradient tracking:}
            \begin{equation} 
                \bs_j^{(t)} =  \bs_j^{(t), K}  + \underbrace{ \nabla f_j \big( \bm{y}_j^{(t)} \big) - \nabla f_j \big( \bm{y}_j^{(t-1)} \big)}_{\text{gradient tracking}}.   
                \label{eq:network_dane_s_update}
            \end{equation} 
            \vspace{-1em}

            \STATE{Update the parameter estimate by solving: }
            \begin{equation}
                \x_j^{(t)}
                = \argmin_{\bz \in \RR^d}\; \left\{ f_j(\bz) - \big\langle \nabla f_j(\y_j^{(t)}) - \bs_j^{(t)}, \bz \big\rangle + \frac\mu2 \big\| \bz - \y_j^{(t)} \big\|_2^2 \right\}.
                \label{eq:local_optimization}
            \end{equation}
            \vspace{-1em}
        \ENDFOR

    \ENDFOR

\end{algorithmic}
\end{algorithm}

With this convergence issue in mind, our key idea is composed of the following components.
\begin{itemize}
\item The first ingredient is to maintain an additional estimate of  the global gradient in each agent --- denoted by $\bm{s}_j^{(t)}$ in the $j$th agent.
This additional gradient estimate is updated via dynamic average consensus \eqref{eq:network_dane_s_update},
in the hope of tracking the global gradient evaluated at $\y_j^{(t)}$ in the $j$th agent ($1\leq j\leq n$), i.e.~$\bs_j^{(t)}$ attempts to track $\nabla f(\y_j^{(t)})$. Here, $\bm{y}_{j}^{(t)}$ stands for the parameter estimate obtained by local neighborly averaging in the $t$th iteration (see Alg.~\ref{alg:network_dane} for details).
		As the algorithm converges, $\{\bm{y}_j^{(t)}\}_{1\leq j\leq n}$ is expected to reach consensus, allowing $\bm{s}_j^{(t)}$ $(1\leq j\leq n)$ to converge to the true global gradient as well.
\item In addition, we also allow multiple rounds of mixing within each iteration, i.e.~\eqref{eq:network_dane_mixing}, which is helpful in accelerating convergence when the network exhibits a high degree of locality. In essence, by applying $K$ rounds of mixing, we improve the mixing rate from $\alpha_0$ to
\begin{equation}\label{eq:def_alpha}
\alpha = \alpha_0^K.
\end{equation} 
As we shall see later, choosing a proper (but not too large) $K$ suffices to achieve the desired trade-off between the rate of information sharing and iteration complexity, which helps reduce the overall communication and computation cost. This step of extra averaging can be implemented in an efficient manner via the Chebyshev acceleration scheme \cite{arioli2014chebyshev,scaman2017optimal}.
\end{itemize}

Armed with such improved global gradient estimates, we propose to solve a modified local optimization subproblem \eqref{eq:local_optimization} in \texttt{Network-DANE}, which approximates the original Newton-type problem \eqref{eq:local_approximation} by replacing $\nabla f(\bbx^{(t)})$ with the local surrogate $\bm{s}_j^{(t)}$.  
The proposed local subproblem \eqref{eq:local_optimization} is convex  and  can be solved efficiently via, say, Nesterov's accelerated gradient methods. The whole algorithm is presented in Alg.~\ref{alg:network_dane}.

\begin{remark} It is certainly possible to employ more general mixing matrices in \eqref{eq:network_dane_mixing}. For instance, in mobile computing scenarios with moving agents, one might prefer using time-varying mixing matrices in order to accommodate the topology changes over time. We omit such extensions for brevity.
\end{remark}

\subsection{Assumptions and Key Parameters}

Before stating theoretical convergence guarantees of \texttt{Network-DANE}, we formally introduce a few assumptions, key parameters, and error metrics.

\begin{assumption}[strongly convex loss]
    \label{assumption:strongly_convex_risk}
    The loss function $f_j(\bx)$ at each agent is strongly convex and smooth, namely, $\sigma \bI \preceq \nabla^2 f_j(\x) \preceq L \bI$ $(1\leq j\leq n)$ for some quantities $0<\sigma\leq L$, where $\kappa =L/\sigma$ is the condition number.
\end{assumption}

\begin{assumption}[quadratic loss]
    \label{assumption:quadratic_risk}
    The loss function $f_j(\bx)$ at each agent is quadratic w.r.t. $\bx$, i.e. taking the form of \eqref{eq:quadratic-problem}.
\end{assumption}

In the strongly convex setting, let the unique global optimizer of $f(\bm{x})$ be
\begin{align}
	\by^{\mathsf{opt}} := \underset{{\bm{x}\in \mathbb{R}^d}}{\arg\min} ~f(\bm{x}).
	\label{eq:defn-global-opt}
\end{align}

In the following definition, we further define the homogeneity parameter \cite{cen2019convergence,fan2019communication}.
\begin{definition}[Homogeneity parameter]
    \label{definition:beta}
    Let $f(\cdot)$ and $f_j(\cdot)$ be as defined in \eqref{eq:global_decom}.
    The homogeneity parameter $\beta$ is defined as
\begin{equation} \label{eq:homogeneity}
	\beta  :=\max_{1\leq j\leq n} \beta_j \qquad \text{with }\; \beta_j := \sup_{\x \in \RR^d} \big\| \nabla^2 f_j(\x) - \nabla^2 f(\x) \big\|.
\end{equation}
\end{definition}

As it turns out, $\beta$ is bounded by the smoothness parameter of $f(\bx)$, i.e.~$\beta\leq L$.\footnote{To see this, we note from the minimax theorem of eigenvalues and the triangle inequality that
\begin{align}
\beta \,\leq\,& \max_j \Bigg\{ \sup_{\x \in \RR^d,\| \bv \|_2 = 1} \bv^\top \Big( \tfrac{n-1}{n} \nabla^2 f_j(\x) \Big) \bv -  \inf_{\x \in \RR^d,\| \bv \|_2 = 1} \bv^\top \Big( \tfrac{1}{n} \sum_{i: i\neq j} \nabla^2 f_i(\x) \Big) \bv \Bigg\} 
	= \big(1 - \tfrac{1}{n} \big) (L - \sigma) \leq L.
\label{lemma:bound_on_beta}
\end{align}
} On the other end, as the local loss functions $f_j$'s become similar with each other, $\beta$ will become smaller. Therefore, $\beta$ is a key quantity measuring the similarity of data across agents.

\begin{remark} \label{remark:beta_vs_m} If the local data follow certain statistical models, it is possible to show that $\beta$ decreases as the local data size $m$ grows. For example, \cite{shamir2014communication} shows that if the data samples at all agents are i.i.d.~(with $\ell(\bx;\bz)$ defined in \eqref{eq:global_decom} satisfying $0\preceq  \nabla^2 \ell(\bx;\bz) \preceq L \bI$ for all $\bz$), then with probability at least $1 - \delta$ over the samples, we have $   \beta < \sqrt{\frac{32 L^2}{m} \log \frac{nd}{\delta}}$ -- implying $\beta$ decreases at the rate of $1/\sqrt{m}$. 
\end{remark}

\paragraph{Metrics and convergence.}
We define the following $(nd)$-dimensional vectors
\begin{align}
	\label{eq:defn-xt-yt-st}
    	\bm{x}^{(t)} :=\big[\bm{x}_{1}^{(t)\top},\cdots,\bm{x}_{n}^{(t)\top}\big]^{\top},\quad
	\bm{y}^{(t)} :=\big[\bm{y}_{1}^{(t)\top},\cdots,\bm{y}_{n}^{(t)\top}\big]^{\top},\quad
	\bm{s}^{(t)} :=\big[\bm{s}_{1}^{(t)\top},\cdots,\bm{s}_{n}^{(t)\top}\big]^{\top}.
\end{align}
The average of each $(nd)$-dimensional vector is defined by $\bbx = \frac1n \sum_{j=1}^n \x_j \in \RR^d$. In addition, we introduce the distributed gradient $\nabla F(\x) \in \RR^{nd}$
and the global gradient $\nabla f(\x) \in \RR^{nd}$ of an $(nd)$-dimensional vector $\x$ as follows
\begin{align}
	\label{eq:def_gradient}
    \nabla F(\x) := [\nabla f_1(\x_1)^\top, \cdots, \nabla f_n(\x_n)^\top]^\top,\quad
    \nabla f(\x) := [\nabla f(\x_1)^\top, \cdots, \nabla f(\x_n)^\top]^\top .
\end{align}

To characterize the convergence behavior of our algorithm, we need to simultaneously track several interrelated error metrics as follows 
\begin{itemize}
	\item[(1)] the convergence error: $\big\| \overline \by^{(t)} - \by^{\mathsf{opt}} \big\|_2$;
	\item[(2)] the parameter consensus error: $\big\| \by^{(t)} -  \bm{1}_n \otimes \overline{\by}^{(t)} \big\|_2$; 
    \item[(3)] the gradient estimation error: $\big\| \bs^{(t)} - \onet \nabla f(\y^{(t)}) \big\|_2$. 
\end{itemize}
In this paper, an algorithm is said to converge linearly at a rate $\rho\in (0,1)$ if there exists some constant $C>0$ such that the following holds for all $t\geq 1$:
\begin{align*}
\max\left\{  \sqrt{n} \big\Vert \overline \by^{(t)}   - \by^{\mathsf{opt}} \big\Vert_2 ,
\big\Vert \by^{(t)} - \onet \overline \by^{(t)} \big\Vert_2 ,  L^{-1} \big\Vert \bs^{(t)} - \nabla f(\y^{(t)}) \big\Vert_2  \right\} \leq C \rho^t .
\end{align*}
In addition, an algorithm is said to reach $\varepsilon$-accuracy if the left-hand side of the above expression is bounded by $\varepsilon$.
 
\subsection{Theoretical Guarantees of \texttt{Network-DANE} for Quadratic Losses}

This subsection establishes linear convergence of \texttt{Network-DANE} when the objective functions are quadratic. The proofs are postponed to Appendix~\ref{proof:convergence_quadratic}.
\begin{theorem}[\texttt{Network-DANE} under quadratic loss, arbitrary $K$]
\label{theorem:dane_convergence_quadratic}
Suppose that Assumptions~\ref{assumption:strongly_convex_risk} and \ref{assumption:quadratic_risk} hold. Set $\alpha = \alpha_0^K$, and 
take $\mu$ large enough so that $\sigma + \mu \geq \frac{140 L}{(1 - \alpha)^2} \left( \frac{\beta}{\sigma} + 1 \right)$. 
Then \texttt{Network-DANE} converges linearly at a rate $\rho_{1}$ obeying
\begin{align}
 \rho_{1} :=   \max \left\{
        \frac{1 + \theta_1}{2}, \,
        \alpha + \frac{140 \kappa}{1-\alpha} \left( \frac{\sigma+\beta}{\sigma+\mu}  \right),
        \frac{1 + \alpha}{2} + \frac{2 \beta}{\sigma + \mu}
    \right\},
    \label{eq:rho_quadratic_bound}
\end{align}
where $\theta_1$ is defined by
\begin{align}
    \theta_1
    :=& 1 - \frac{\sigma}{\sigma + \mu} + \frac{L}{L + \mu} \frac{\beta^2}{(\sigma + \mu)(\sigma + \mu - \beta)} . \label{eq:dane_quadratic_theta_def}
\end{align}
\end{theorem}
\begin{remark}
It turns out that $\theta_1 \in (0,1)$ is the convergence rate of DANE in the master/slave setting under quadratic losses \cite[Theorem 1]{shamir2014communication}. 
\end{remark}

It is worth noting that we have spent no effort in optimizing the pre-constants in the above theorem.
If the regularization parameter $\mu$ is sufficiently large,
one can guarantee that $\theta_1 < 1$ and hence DANE converges at a linear rate when optimizing quadratic losses \cite{shamir2014communication}.
We can clearly see that \eqref{eq:rho_quadratic_bound} is always greater than $\theta_1$,
which is the price we pay for consensus under the network setting.
Fortunately, by properly setting $\mu$,
we can still guarantee that $\rho_1<1$,
which in turn enables linear convergence of \texttt{Network-DANE}.

In view of \eqref{eq:rho_quadratic_bound}, if the network is sufficiently connected (i.e.~$\alpha$ is small),
or if the data are sufficiently homogeneous (i.e.~$\beta$ is small), we can use a smaller parameter $\mu$,
which makes $\theta_1$ (defined in \eqref{eq:dane_quadratic_theta_def}) smaller and results in faster convergence.
In summary, \texttt{Network-DANE} takes fewer iterations to converge when $\alpha$ and $\beta$ are both small.
After some basic calculations, the complexity of \texttt{Network-DANE} for quadratic losses is formalized in the following corollary.
\begin{corollary} \label{corollary:dane_quadratic_rate}
Set $\mu + \sigma = \frac{180 L}{(1 - \alpha)^2} (\frac\beta\sigma + 1)$.
Under the assumptions of Theorem~\ref{theorem:dane_convergence_quadratic},
one has
\begin{align}   \label{eq:dane_quadratic_rate_1}
    \rho_1
    \leq 1 - \left(\frac{1 - \alpha}{20}\right)^2 \frac{1}{\kappa} \frac{1}{(\beta/\sigma + 1)}.
\end{align}
To reach $\varepsilon$-accuracy, 
\texttt{Network-DANE} takes at most $O \left( \frac{ \kappa (\beta / \sigma + 1) \log({1}/{\varepsilon}) }{ (1 - \alpha)^2 } \right)$ iterations,\\
and $O \left( K \cdot \frac{ \kappa (\beta / \sigma + 1) \log({1}/{\varepsilon}) }{ (1 - \alpha)^2 } \right)$ communication rounds.
\end{corollary}
Recall that if we set the number of local averaging rounds to be $K=1$, then one has $\alpha=\alpha_0$, and hence our iteration complexity can be readily compared with other existing results.
If the homogeneous parameter $\beta$ obeys $\beta = O(\sigma)$, then the convergence rate can be improved to $O \big(\kappa \log(1/\varepsilon) / (1 - \alpha_0)^2 \big)$; this is much faster than the corrected DGD \cite{qu2018harnessing} with gradient tracking,
which converges in $O(\kappa^2\log(1/\varepsilon) /(1-\alpha_0)^2 )$ iterations. The convergence rate of \texttt{Network-DANE} degenerates to that of DGD \cite{qu2018harnessing} with gradient tracking under the worst condition $\beta = \Theta(L)$.
This observation highlights the communication efficiency of \texttt{Network-DANE} by harnessing the homogeneity of data across different agents. We emphasize that this is an important feature of our analysis, where the convergence rate adapts with respect to the data homogeneity.
 
\paragraph{Benefits of extra local averaging (i.e.~$K>1$).} The careful reader might have noticed that the rate  established above scales poorly with respect to the network parameter, namely, $1-\alpha_0$, when $K=1$. One remedy is to consider the case with $K > 1$, where \texttt{Network-DANE} performs $K$ rounds of communications per iteration. On the one hand, the effective network parameter $\alpha = \alpha_0^K$ can be made arbitrarily small by taking $K$ sufficiently large,
thus leading to faster convergence;
on the other hand, the total number of communications is $K$ times larger than the number of iterations,
meaning that we might end up with a higher communication complexity. As an example, invoking Corollary~\ref{corollary:dane_quadratic_rate}, we see that: the total communication cost to reach $\varepsilon$-accuracy,
in terms of the native network parameter $\alpha_0$,
is given by
$$O \big(K \cdot \kappa (1 + \beta/\sigma) \log(1/\varepsilon) / (1 - \alpha_0^K)^2 \big).$$ Therefore, by judiciously choosing $K$, it is possible to significantly improve the overall communication complexity, especially when $\alpha_0$ is close to $1$. 
For example, by setting $K \asymp 1 / \log(1/\alpha_0) = O(1 / (1-\alpha_0))$,
we can ensure $\alpha_0^K \asymp 1 / 2$ and reduce the communication complexity to $O \big( \kappa \cdot (\beta/\sigma + 1) \log(1/\varepsilon) / (1 - \alpha_0) \big)$, thus improving the dependence with the graph topology.

The following theorem shows an improved result following a refined analysis, which improves the dependence simultaneously with respect to both $\kappa$ and $1-\alpha_0$.  

\begin{theorem}[\texttt{Network-DANE} under quadratic loss, optimized $K$] \label{corollary:dane_rate_varyingK_quadratic}
Instate the assumptions of Theorem~\ref{theorem:dane_convergence_quadratic}. Set $K$ and  $\mu$ large enough so that $\alpha = \alpha_0^K \leq 1/ (2\kappa)$ and $\sigma + \mu \geq 360 \sigma \left( \frac{\beta^2}{\sigma^2} + 1 \right)$. To reach $\varepsilon$-accuracy, 
\texttt{Network-DANE} takes at most $O \left((\beta^2/\sigma^2 + 1 ) \log (1/\varepsilon) \right)$ iterations,
and $O \left( \log\kappa \cdot \frac{ (\beta^2/\sigma^2 + 1) \log(1/\varepsilon )} { 1 - \alpha_0 } \right)$ communications rounds.
\end{theorem}
When we set $K$ as suggested in Theorem~\ref{corollary:dane_rate_varyingK_quadratic},
the iteration complexity becomes independent of the network topology. Moreover, it matches the rate of DANE in the master/slave setting \cite{shamir2014communication} when $\beta=O(\sigma)$, which is $O(\log(1/\varepsilon))$ and further independent of the condition number $\kappa$. 

In terms of network dependence, the communication complexity improves from $O \big( 1/(1 - \alpha_0)^2 \big)$ to $O \big( 1 / (1 - \alpha_0) \big)$. By implementing the extra averaging step in an efficient manner via the well-known Chebyshev acceleration scheme \cite{arioli2014chebyshev,scaman2017optimal}, the dependence of the communication complexity with respect to $1-\alpha_0$ can be further improved to $O \left((1-\alpha_0)^{-1/2}\right)$. The final communication complexity of \texttt{Network-DANE} for quadratic losses thus becomes 
$$O \left( \log\kappa \cdot \frac{ (\beta^2/\sigma^2 + 1) \log(1/\varepsilon )} { (1 - \alpha_0)^{1/2} } \right) .$$
Therefore, the total amount of communication is significantly reduced using extra averaging, where it scales only logarithmically with respect to $\kappa$.

\subsection{Theoretical Guarantees of \texttt{Network-DANE} for Strongly Convex Losses} 

This subsection establishes the linear convergence of \texttt{Network-DANE} for general smooth and strongly convex loss functions, where the rate is worse than that for  quadratic losses.
The proof can be found in Appendix~\ref{proof:convergence}.
\begin{theorem}
\label{theorem:dane_convergence}
Suppose that Assumption~\ref{assumption:strongly_convex_risk} holds.
Set $\alpha = \alpha_0^K$, and take $\mu$ large enough so that $\sigma + \mu \geq \frac{170 \kappa L}{(1 - \alpha)^2}$. 
Then \texttt{Network-DANE} converges linearly at a rate $\rho_2$ obeying
\begin{align} 
 \rho_2 :=   \max \left\{
        \frac{1 + \theta_2}{2},
        \alpha + \frac{170 \kappa }{1 - \alpha}\left(\frac{L}{\sigma+\mu}\right),
        \frac{1+\alpha}{2} +\frac{ 2\beta}{\sigma+\mu}
    \right\},
    \label{eq:rho_bound}
\end{align}
where $\theta_2$ is given by
\begin{align}
    \label{eq:defn-alpha-beta-theta}
    \theta_2 :=& 1 - \frac{\sigma}{\sigma + \mu} + \frac{\beta}{\sigma + \mu} \sqrt{1 - \Big( \frac{\mu}{\sigma + \mu} \Big)^2} .
\end{align}
\end{theorem}
\begin{remark}
Note that $\theta_2 \in (0,1)$ is precisely the convergence rate of DANE in the master/slave
	setting (see \cite[Theorem 3.1]{fan2019communication}). 
\end{remark}

Similar to Theorem~\ref{theorem:dane_convergence_quadratic},
one can guarantee $\theta_2 < 1$ and $\rho_2 < 1$ by setting the regularization parameter $\mu$ sufficiently large. Therefore, 
 \texttt{Network-DANE} can converge at a linear rate for a general class of smooth and strongly convex problems. Comparing the convergence rates of \texttt{Network-DANE} derived for the above two different losses (i.e.~comparing \eqref{eq:dane_quadratic_theta_def} with \eqref{eq:defn-alpha-beta-theta}), we see that: when the loss functions are non-quadratic, $\theta_2$ is generally greater than $\theta_1$\footnote{This is because $\sqrt{\frac{\sigma^2 + 2 \sigma \mu}{(\sigma + \mu)^2}} \geq \frac{\sigma}{\sigma + \mu}$.}. 
This happens since the Hessian matrices associated with the non-quadratic loss functions may vary across different points, 
which is also the reason why the convergence rate of \texttt{Network-DANE} derived for the general case degenerates to the worst-case rate.
After some basic calculations, the complexity of \texttt{Network-DANE} under strongly convex losses is formalized by the following corollary.
\begin{corollary} \label{corollary:dane_rate}
Set $\sigma + \mu = \frac{180\kappa L}{(1 - \alpha)^2}$. Under the assumptions of Theorem~\ref{theorem:dane_convergence}, 
one has
\begin{align}   \label{eq:dane_rate_1}
    \rho_2
    \leq 1 - \left(\frac{1 - \alpha}{20}\right)^2 \frac{1}{\kappa^2} .
\end{align}
To reach $\varepsilon$-accuracy, 
\texttt{Network-DANE} takes at most $O \left( \frac{ \kappa^2 \log({1}/{\varepsilon} ) }{ (1 - \alpha)^2 } \right)$ iterations and $O \left( K \cdot \frac{ \kappa^2 \log({1}/{\varepsilon} )  }{ (1 - \alpha)^2 } \right)$ communication rounds.
\end{corollary}

 When $K=1$, the communication complexity of \texttt{Network-DANE} is $ O \left( \frac{ \kappa^2 \log({1}/{\varepsilon} )  }{ (1 - \alpha)^2 } \right)$, which is rather pessimistic and does not improve with data homogeneity. Similar to Theorem~\ref{corollary:dane_rate_varyingK_quadratic}, we can improve this by optimizing $K$ properly. 
We have the following theorem, which is parallel to Theorem~\ref{corollary:dane_rate_varyingK_quadratic}.
\begin{theorem}[\texttt{Network-DANE} under strongly convex loss, optimized $K$]  \label{corollary:dane_rate_varyingK}
Instate the assumptions of Theorem~\ref{theorem:dane_convergence}. Set $K$ and $\mu$ large enough so that $\alpha = \alpha_0^K \leq 1/ (2\kappa)$ and $\sigma + \mu \geq 360 L \left( \frac{\beta}{\sigma} + 1 \right)$. 
To reach $\varepsilon$-accuracy, \texttt{Network-DANE} takes at most $O \left(\kappa (\beta/\sigma + 1 ) \log (1/\varepsilon) \right)$ iterations
and $O \left( \log\kappa \cdot \frac{ \kappa (\beta/\sigma + 1)\log (1/\varepsilon) } { 1 - \alpha_0 } \right)$ communication rounds.

\end{theorem}

The improved rate in Theorem~\ref{corollary:dane_rate_varyingK} improves as the local data become more homogeneous, recovering a feature that has been highlighted previously. Similar to earlier discussions, by using the Chebyshev acceleration scheme \cite{arioli2014chebyshev,scaman2017optimal}, the final communication complexity of \texttt{Network-DANE} for strongly convex losses becomes 
$$O \left( \log\kappa \cdot \frac{ \kappa (\beta/\sigma + 1) \log(1/\varepsilon )} { (1 - \alpha_0)^{1/2} } \right) .$$

\begin{remark}
    The homogeneity parameter $\beta$ defined in Definition~\ref{definition:beta} measures the largest deviation of local Hessians from the global Hessian. A refined analysis using local deviation $\beta_j$ is possible by permitting different regularization parameters $\mu_j$ in \eqref{eq:local_optimization} for different agents.
\end{remark}

\subsection{Extension to Nonsmooth Composite Optimization}%
\label{sub:non-smooth_loss}

The proposed algorithms can be extended for nonsmooth composite optimization, by properly adjusting the local optimization step, leveraging proximal variants of DANE \cite{fan2019communication} and SVRG \cite{xiao2014proximal}. For simplicity, we present the proximal variant of \texttt{Network-DANE} and leave its theoretical analysis to future work.

Consider the following regularized empirical risk minimization problem:
\begin{equation}
    \underset{\x \, \in \, \RR^d}{\text{minimize}} \quad f(\x) + g(\x)
    \triangleq \frac{1}{N} \sum_{i=1}^{N} \ell(\x; \bz_i) + g(\x),
    \label{eq:non-smooth_problem}
\end{equation}
where $f(\cdot)$ and $f_j(\cdot)$ are defined as in \eqref{eq:global_decom},
and $g(\cdot)$ is a deterministic convex regularizer that can be nonsmooth. This type of problem has wide applications, where it is desirable to promote additional structures or incorporate prior knowledge about the solution through adding a deterministic regularization term $g(\x)$. We can extend \texttt{Network-DANE} to solve \eqref{eq:non-smooth_problem} by adding the proximal term into the local optimization step, as detailed in  Algorithm~\ref{alg:network_dane_nonsmooth}, which is a direct extension of Algorithm~\ref{alg:network_dane}. Section~\ref{sec:numerical} numerically verifies the effectiveness of Algorithm~\ref{alg:network_dane_nonsmooth}.
\begin{algorithm}[htbp]
    \caption{\texttt{Network-DANE} for nonsmooth composite optimization}
	\label{alg:network_dane_nonsmooth}
	\begin{algorithmic}[1]
		   \STATE {Replace the local optimization sub-problem \eqref{eq:local_optimization} of \texttt{Network-DANE} by the following:}

		\STATE {\textbf{Input:} $\y_j^{(t)}$, $\bs_j^{(t)}$, regularization parameter $\mu$.}
		
        \STATE{Update the parameter estimate by solving: }

            \begin{equation}
                \x_j^{(t)}
                = \argmin_{\bz \in \RR^d}\; \left\{ f_j(\bz) + g(\bz) - \big\langle \nabla f_j(\y_j^{(t)}) - \bs_j^{(t)}, \bz \big\rangle + \frac\mu2 \big\| \bz - \y_j^{(t)} \big\|_2^2 \right\}.
                \label{eq:local_optimization_nonsmooth}
            \end{equation}
	\end{algorithmic}
\end{algorithm}

\section{Generalizing the Algorithm Design with Variance Reduction}
\label{sec:network_svrg}

The design of \texttt{Network-DANE} suggests a systematic approach to obtain decentralized versions of other algorithms.
We illustrate this by reducing local computation of \texttt{Network-DANE} using variance reduction. 
Stochastic variance reduction methods are a popular class of stochastic optimization algorithms, developed to allow for constant step sizes and faster convergence in finite-sum optimization \cite{johnson2013accelerating,xiao2014proximal,nguyen2017sarah}. It is therefore natural to ask whether such variance reduction techniques can be leveraged in a network setting to further save local computation without compromising communication. 

\begin{algorithm}[ht]
	\caption{\texttt{Network-SVRG/SARAH}}
	\label{alg:network_svrg}
	\begin{algorithmic}[1]
		   \STATE {Replace the local optimization subproblem \eqref{eq:local_optimization} of \texttt{Network-DANE} by the following:}

		\STATE {\textbf{Input:} $\y_j^{(t)}$, $\bs_j^{(t)}$, step size $\delta$, number of local iterations $S$.}
		
        \STATE{\textbf{Initialization:} set $\bu_j^{(t), 0} = \y_j^{(t)}$, $\bv_j^{(t), 0} = \bs_j^{(t)}$.}
		\FOR{$s=1,...,S$}
	\STATE{ $
                    \bu_j^{(t), s} = \bu_j^{(t), s-1} - \delta \bv_j^{(t), s-1}  $.}
                
		\STATE{Sample $\bz$ from $\cM_j$ uniformly at random, then,
            \begin{subnumcases}{\bv_j^{(t), s} = }
                \nabla \ell(\bu_j^{(t), s}; \bz) - \nabla \ell(\bu_j^{(t), 0}; \bz)   + \bv_j^{(t), 0}; & \texttt{(SVRG)} \label{eq:svrg_local_v} \\
                \nabla \ell(\bu_j^{(t), s}; \bz) - \nabla \ell(\bu_j^{(t), s-1}; \bz)  + \bv_j^{(t), s-1}. & \texttt{(SARAH)}  \label{eq:sarah_local_v}
            \end{subnumcases}
            }
		\ENDFOR	
        \STATE{Choose the new parameter estimate $\x_j^{(t)}$ from $\{\bu_j^{(t),1},\cdots,\bu_j^{(t),S}\}$ uniformly at random.}	
		
	\end{algorithmic}
\end{algorithm}

Inspired by the connection between DANE and SVRG \cite{konevcny2015federated},
we introduce \texttt{Network-SVRG/SARAH} in Alg.~\ref{alg:network_svrg}, a decentralized version of SVRG \cite{johnson2013accelerating} and SARAH \cite{nguyen2017sarah} tailored to the network setting, with the assistance of gradient tracking.
In particular, the inner loops of SVRG \cite{johnson2013accelerating} or SARAH \cite{nguyen2017sarah} are adopted to replace the local computation subproblem \eqref{eq:local_optimization} of \texttt{Network-DANE}, where the reference to the global gradient is replaced by $\bm{s}_j^{(t)}$ to calculate the variance-reduced stochastic gradient.

 The convergence analysis of Alg.~\ref{alg:network_svrg}
is more challenging due to the biased stochastic gradient involved in each local iteration. Encouragingly, the theorem below establishes the linear convergence of \texttt{Network-SVRG} for strongly convex losses, and of \texttt{Network-SARAH} for quadratic losses, as long as $\beta$ is sufficiently small and the number of mixing rounds $K$ is sufficiently large. Again, we have not strived to improve the pre-constants specified in the theorem.

\begin{theorem}
    \label{theorem:svrg}
    Assume that the sample loss $\ell( \bx; \bz)$ is convex and $L$-smooth w.r.t. $\bx$ for all $\bz$. If $ \beta / \sigma \leq 1/200$,
    set $K$ large enough such that $\alpha =\alpha_0^K \asymp 1/\kappa$ and $S$ large enough, \texttt{Network-SVRG} converges linearly under Assumption~\ref{assumption:strongly_convex_risk}; and \texttt{Network-SARAH} converges linearly under Assumptions~\ref{assumption:strongly_convex_risk} and \ref{assumption:quadratic_risk}. In particular, to reach $\varepsilon$-accuracy,
\texttt{Network-SVRG} and \texttt{Network-SARAH} take at most $O\left(\log (1/\varepsilon) \right)$ iterations
and $O \left( \log\kappa \cdot \frac{ \log(1/\varepsilon)}{1-\alpha_0} \right)$ communication rounds under the aforementioned assumptions.
\end{theorem}
The proof of Theorem~\ref{theorem:svrg} can be found in Appendix~\ref{proof:svrg_and_sarah}. Theorem~\ref{theorem:svrg} implies that: as long as the local data are sufficiently similar (so that $\beta$ does not exceed the order of $\sigma$), by performing $O\left(\log \kappa / (1-\alpha_0) \right)$ rounds of local communication per iteration, \texttt{Network-SVRG} and \texttt{Network-SARAH} converge in $O \left( \log(1/\varepsilon) \right)$ iterations independent of $\kappa$. This performance guarantee matches its counterpart in the master/slave setting \cite{cen2019convergence}. Altogether, \texttt{Network-SVRG/SARAH} achieves appealing computation and communication complexities simultaneously. By further adopting the Chebyshev acceleration scheme \cite{arioli2014chebyshev,scaman2017optimal}, the final communication complexity of \texttt{Network-SVRG/SARAH} is at most
$$O \left( \log\kappa \cdot \frac{ \log(1/\varepsilon)}{(1-\alpha_0)^{1/2}} \right).$$

It is straightforward to extend this idea to obtain decentralized variants of other stochastic variance reduced algorithms such as Katyusha \cite{allen2017katyusha}, basically by replacing the local computation step  \eqref{eq:local_optimization}
by the inner loop update rules of the stochastic methods of interest. For the sake of brevity, this paper does not pursue such ``plug-and-play'' extensions.

\begin{remark}
Our convergence theory of \texttt{Network-SVRG} requires $\beta \lesssim \sigma$, which is consistent with its counterpart in the master/slave setting \cite{cen2019convergence}. In contrast, \texttt{Network-DANE} is guaranteed to converge linearly in the entire range of $\beta$ by setting $\mu$ sufficiently large. One scheme to relax this requirement, as analyzed in \cite{cen2019convergence}, is to add a regularization term, similar to the last term in \eqref{eq:local_optimization}, that penalizes the distance to the previous estimate. However, this might come at a price of slower convergence. We leave this to future investigation.  
\end{remark}

\section{Numerical Experiments}
\label{sec:numerical}

We evaluate the performance of the proposed algorithms\footnote{In our experiments of \texttt{Network-SVRG/SARAH}, we use the last iterate $\bu_j^{(t), S}$ as the new parameter estimate locally, which is more practical; our analysis only handles the case where the new parameter estimate is selected uniformly at random from previous iterates, though.} for solving both strongly convex and nonconvex problems, in order to demonstrate the appealing performance in terms of communication-computation trade-offs. Code for our experiments can be found at 
\begin{center}
\url{https://github.com/liboyue/Network-Distributed-Algorithm}.
\end{center}

Throughout this section, we set the number of agents $n = 20$.
We use symmetric fastest distributed linear averaging (FDLA) matrices \cite{Xiao2004} generated according to the communication graph as the mixing matrix $\bW$ for aggregating $\bx^{(t)}_j$ in \eqref{eq:network_dane_mixing}.
For aggregating $\bs^{(t)}_j$ in \eqref{eq:network_dane_mixing}, we use a convex combination of $\bI$ and $\bW$ such that its diagonal elements are greater than $0.1$, which makes the algorithm more stable in practice.
The same regularization parameter $\mu$ is used for DANE and \texttt{Network-DANE}.
We generate connected random communication graphs using an Erd\"{o}s-R\`{e}nyi graph with the probability of connectivity $p=0.3$ (if not specified).
For each experiment, we use the same random starting point $\bx^{(0)}$ and mixing matrix $\bW$ for all algorithms.
To solve the local optimization subproblems, we use Nesterov's accelerated gradient descent for at most $100$ iterations for DANE and \texttt{Network-DANE}.

\begin{figure}[htb]
\centering
\includegraphics[width=1\textwidth]{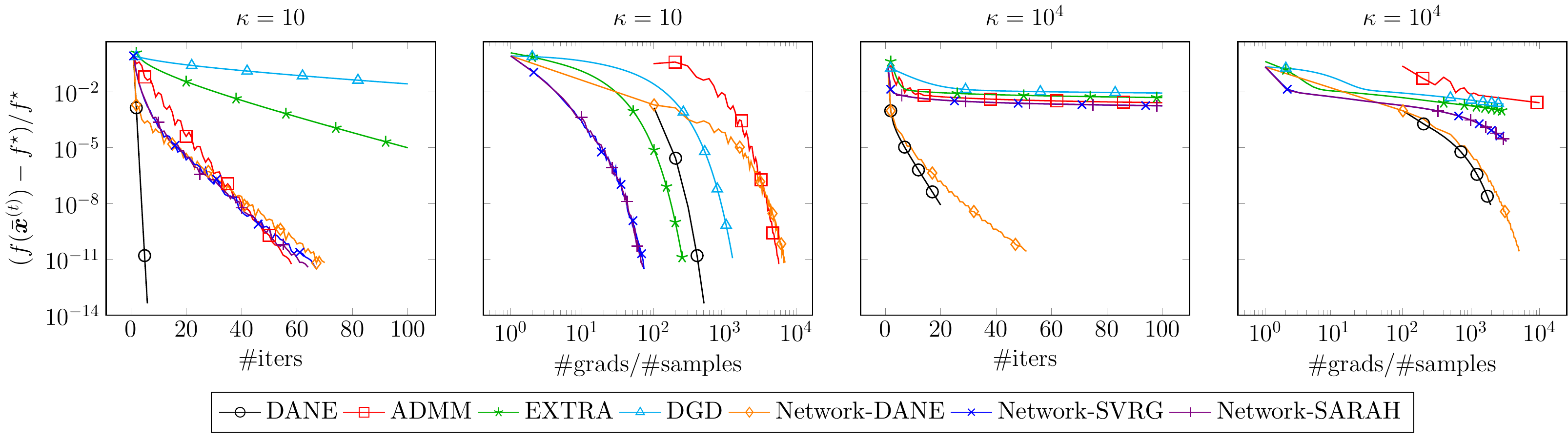}
\caption{The relative optimality gap with respect to the number of iterations and gradient evaluations under different conditioning $\kappa=10$ (left two panels) and $\kappa=10^4$ (right two panels) for linear regression.}
\label{fig:linear_regression}
\end{figure}

\subsection{Experiments On Synthetic Data}%
\label{sub:experiments_on_synthetic_data}
We conduct five synthetic numerical experiments based on linear regression to investigate the performance of our algorithms.
The same data generation method is used for all synthetic experiments.
We generate $m = 1000$ samples of dimension $d=40$, denoted by $\bA_i$, randomly from $\mathcal N (\bm 0, \bSigma)$ i.i.d. for each agent, where $\bSigma$ is a diagonal matrix with $\bSigma_{ii} = i^{-\varrho}$.
By changing $\varrho$, we can change the condition number $\kappa$. 
Data samples are generated according the linear model $\bb_i = \bA_i \bx_0 + \bxi_i$,
with a random signal $\bx_0$ and i.i.d. noise $\bxi_i \sim \mathcal N(\bm 0, \bI)$. For DANE and \texttt{Network-DANE}, we set $\mu = 5 \times 10^{-10}$ when $\kappa = 10$ and $\mu = 5 \times 10^{-4}$ when $\kappa = 10^4$.
For \texttt{Network-SVRG/SARAH},
we set the step size $\delta = 0.1 / (L + \sigma + 2\mu)$,
the number of local iterations $S = 0.05m$.

\paragraph{Comparison with existing algorithms.}
To make a fair comparison with other algorithms, no extra local averaging is adopted in this experiment, i.e. the number of mixing rounds is set to $K=1$.
The loss function at each agent is given as $f_i(\bx) = \frac{1}{2m}\|\bA_i \bx - \bb_i \|_2^2$.
We plot the relative optimality gap, given as $ ( f(\overline\bx^{(t)}) - f^\star) / f^\star$, where $\overline \bx^{(t)}$ is the average parameter of all agents at the $t$th iteration, and $f^\star$ is the optimal value.
We compare the proposed \texttt{Network-DANE} (Alg.~\ref{alg:network_dane}) and \texttt{Network-SVRG/SARAH} (Alg.~\ref{alg:network_svrg}) with the master/slave algorithm DANE \cite{shamir2014communication} and ADMM \cite{boyd2011distributed},\footnote{We apply ADMM to the constrained optimization problem, which amounts to the centrally-distributed setting,
$ \min_{\x_i} \frac1n \sum f_i(\x_i) ~
\text{s.t.} ~ \x_i = \x $. Note that ADMM can also be applied to the network-distributed setting, which is not shown here since our network algorithms already outperform ADMM in the centrally-distributed setting. }
and two popular network-distributed gradient descent algorithms,
referred to as DGD \cite{qu2018harnessing} and EXTRA \cite{shi2015extra}.

Fig.~\ref{fig:linear_regression} shows the relative optimality gap with respect to the number of iterations as well as the number of gradient evaluations under different condition numbers $\kappa=10$ and $\kappa =10^4$ for linear regression.
In both experiments,  \texttt{Network-DANE} and \texttt{Network-SVRG/SARAH} significantly outperform DGD and EXTRA in terms of the numbers of communication rounds. \texttt{Network-SVRG/SARAH} has similar communication rounds with ADMM but only communicates locally. \texttt{Network-DANE} is quite insensitive to the condition number, performing almost as well as the DANE algorithm in the ill-conditioned case, but operates in a fully decentralized setting.
\texttt{Network-SVRG/SARAH} further outperforms other algorithms in terms of gradient evaluations in most settings, especially for well-conditioned cases.
\texttt{Network-SVRG} and \texttt{Network-SARAH} are almost indistinguishable.

\begin{figure}[htb]
	\centering 
    \includegraphics[width=\textwidth]{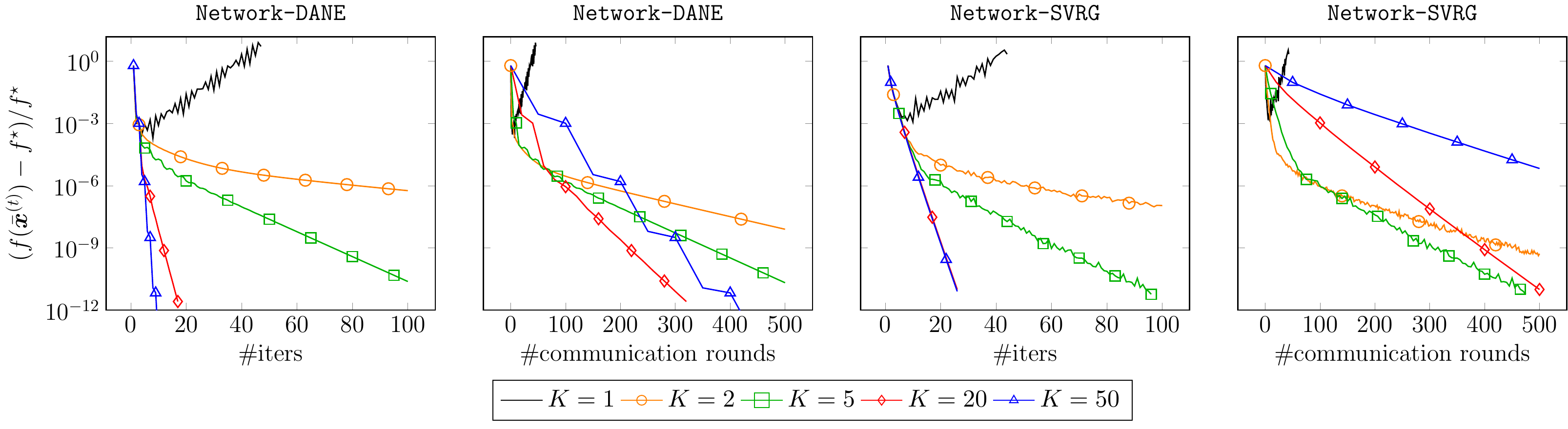}
	\caption{The relative optimality gap with respect to the number of iterations and communication rounds under different rounds of mixing $K$ for \texttt{Network-DANE} (left two panels) and \texttt{Network-SVRG} (right two panels) over a poorly-connected graph.}
	\label{fig:extra_communication}
\end{figure}

\paragraph{Benefits of extra local mixing (communication) per iteration.}
We conduct synthetic experiments to investigate the communication-computation trade-off observed in Corollary~\ref{corollary:dane_rate_varyingK} when employing multiple rounds of mixing within every iteration. Following the suggestion of the theory, we use a poorly-connected network with mixing rate $\alpha_0 = 0.944$ for communication, which is generated by an Erd\"{o}s-R\`{e}nyi graph with $p=0.2$. For illustration, we consider the relative optimality gap for a linear regression problem with $\kappa = 10$,
with respect to the number of iterations and communication rounds for \texttt{Network-DANE} and \texttt{Network-SVRG}, under different values of $K$ (no Chebyshev acceleration is employed), shown in Fig.~\ref{fig:extra_communication}.  
Due to poor connectivity, \texttt{Network-DANE} and \texttt{Network-SVRG} fail to converge when using moderate parameters. However, by using a larger $K$, due to improvement in consensus, both algorithms converge faster in terms of the number of iterations.
Notice that after certain threshold, further increasing $K$ will not improve the convergence rate in terms of communication rounds.

\begin{figure}[htb]
	\centering 
    \includegraphics[width=0.35\linewidth]{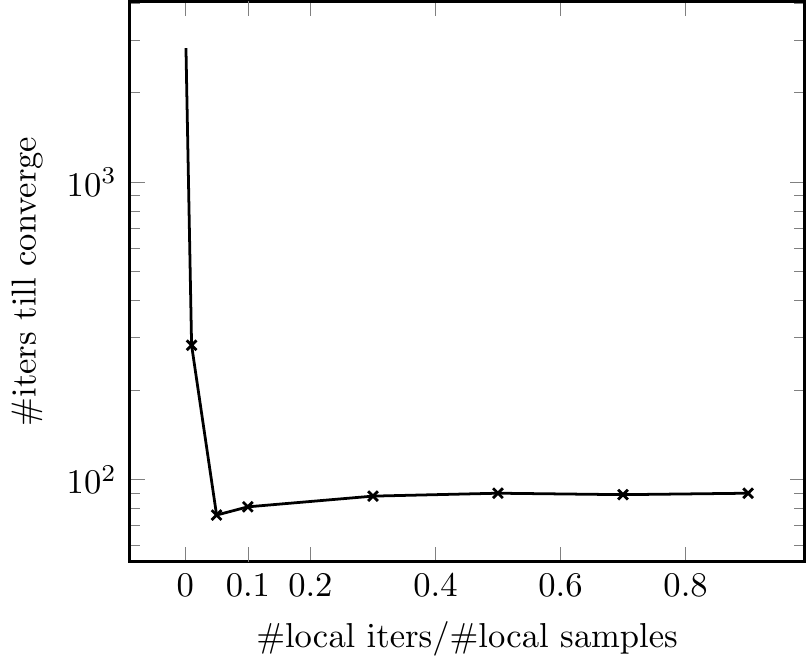}
	\caption{Number of communication rounds and number of gradient evaluations till converge with respect to different numbers of local iterations.}
	\label{fig:svrg_local_iter}
\end{figure}

\paragraph{Effects of local computation for \texttt{Network-SVRG}.}
We conduct an experiment to analyze the effect of different numbers of local stochastic iterations for \texttt{Network-SVRG}. Throughout this experiment,
we run our algorithms on a linear regression problem with $\kappa = 10$ and Erd\"{o}s-R\`{e}nyi graph ($p=0.2$) as the communication graph.
Fig.~\ref{fig:svrg_local_iter} shows the number of communication rounds and the number of gradient evaluations till converge for different numbers of local iterations.
It is clear that with too few local iterations,
\texttt{Network-SVRG} converges very slow and requires more communication.
As soon as $S$ is above a threshold, i.e. around
$0.05m$ local iterations, the communication rounds no longer decreases. Therefore, in our experiments, we set the number of local iterations as $S=0.05m$ to ensure satisfactory convergence rate while using an economical amount of local computation.

\begin{figure}[htb]
	\centering
    \includegraphics[width=0.68\linewidth]{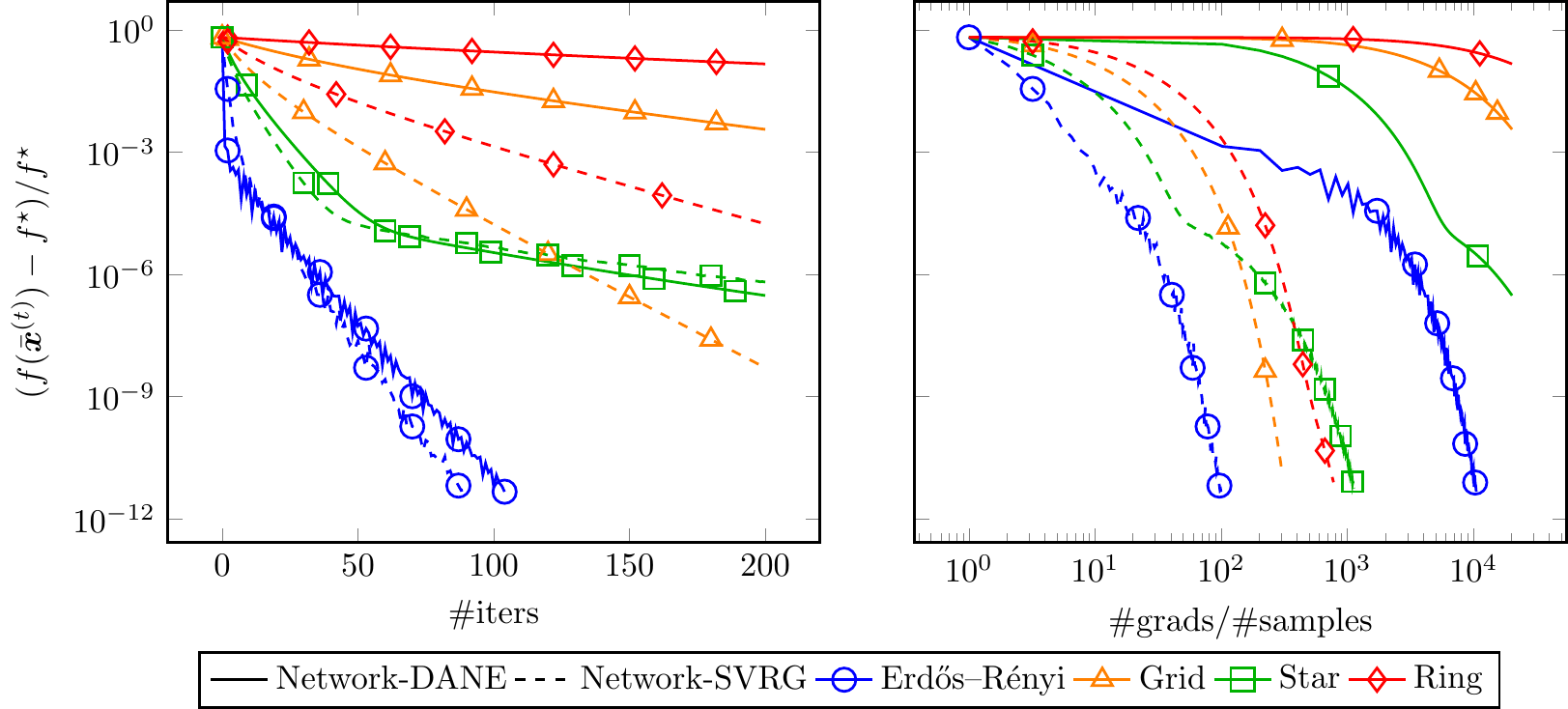}
	\caption{Performance of the proposed algorithms under different network topologies.} 
	\label{fig:topology}
\end{figure}

\paragraph{Effects of network topology.}
We conduct another experiment to compare the effect of network topology on linear regression problem with $\kappa = 10$.
We generate communication graphs with different topology settings.
Fig.~\ref{fig:topology} shows the relative optimality gap with respect to the number of iterations and gradient evaluations for \texttt{Network-DANE} and \texttt{Network-SVRG/SARAH} for Erd\"{o}s-R\`{e}nyi graph ($p=0.3$), a $4 \times 5$ grid graph, a star graph, and a ring graph. The performance degrades as the network becomes less connected (where $1-\alpha_0$ becomes small) \cite{nedic2018network}.

\begin{figure}[htb]
    \centering
    \includegraphics[width=\linewidth]{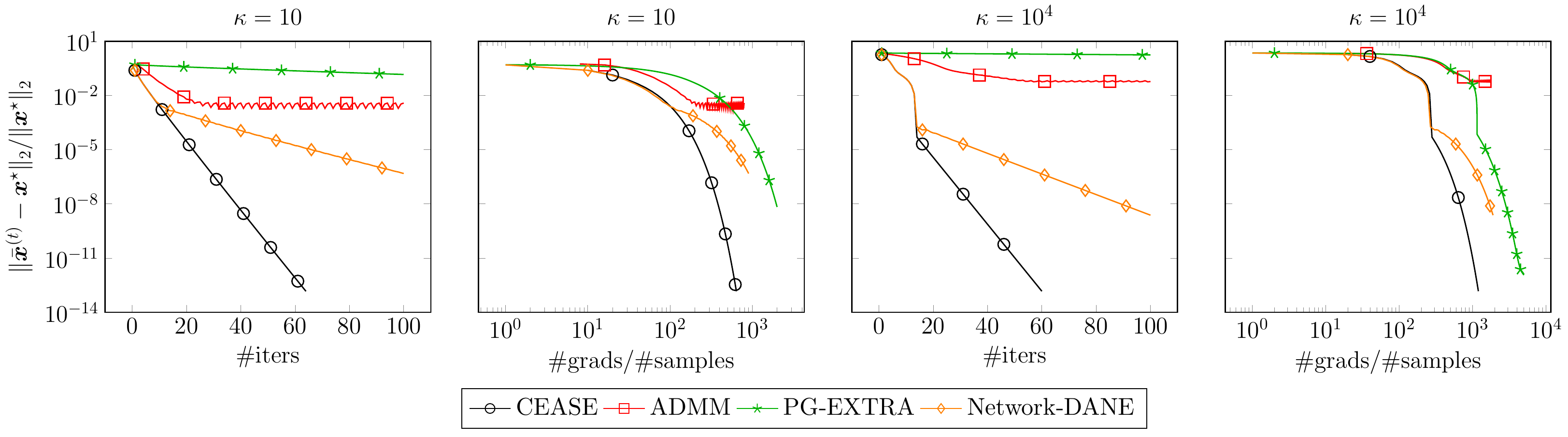}
    \caption{The relative optimality gap with respect to the number of iterations and gradient evaluations under different conditioning $\kappa=10$ (left two panels) and $\kappa=10^4$ (right two panels) for linear regression with $\ell_1$-norm regularization.}
    \label{fig:linear_regression_l1_regularization}
\end{figure}

\paragraph{Experiments for nonsmooth composite optimization} We consider the $\ell_1$-norm regularized linear regression, where the loss function of each agent is given as $\tilde{f}_i(\bx) = f_i(\bx) + g(\bx) = \frac{1}{2m}\|\bA_i \bx - \bb_i \|_2^2 + 0.01 \|\x\|_1$,
and the communication graph are generated in the same way as Fig.~\ref{fig:linear_regression}. The condition number $\kappa$ is also defined in the same way as earlier.
We compare the performance of \texttt{Network-DANE} with CEASE \cite{fan2019communication}, which is the proximal version of DANE in the master/slave setting, ADMM, and PG-EXTRA, which is the proximal version of EXTRA \cite{shi2015proximal}.
For CEASE and \texttt{Network-DANE}, we set $\mu = 10^{-4}$ when $\kappa = 10$ and $\mu = 10^{-1}$ when $\kappa = 10^4$,
and use FISTA \cite{beck2009fast} to solve the $\ell_1$-norm regularized local problems for computation efficiency.
Fig.~\ref{fig:linear_regression_l1_regularization} plots the relative optimality gap $\| \bbx^{(t)} - \x^\opt \|_2 / \| \x^\opt \|_2$ with respect to the number of iterations and the number of gradient evaluations for different algorithms under different condition numbers. In both experiments, \texttt{Network-DANE} outperformed ADMM and PG-EXTRA in both metrics, and achieves similar convergence behavior as CEASE, though at a slower rate due to optimizing over a decentralized topology.

\subsection{Experiments On Real Data}%
We perform two experiments on real data to further evaluate the performance of the proposed algorithms for both convex and nonconvex problems.

\paragraph{Binary classification using logistic regression.}
We use regularized logistic regression to solve a binary classification problem using the Gisette dataset.\footnote{The dataset can be found at \href{https://archive.ics.uci.edu/ml/datasets/Gisette}{https://archive.ics.uci.edu/ml/datasets/Gisette}.} 
We split the Gisette dataset to $n=20$ agents, where
each agent receives $m=300$ training samples of dimension $d=5000$.
The loss function at each agent is given as
\[
    f_i(\bx)
    = - \frac{1}{m} \sum_{j=1}^{m}
    \Big[ b_i^{(j)} \log \Big(\frac{1}{1 + \exp( \bx^\top \ba_i^{(j)})} \Big)
    + (1 - b_i^{(j)}) \log \Big(  \frac{\exp(\bx^\top \ba_i^{(j)})}{1 + \exp(\bx^\top \ba_i^{(j)})} \Big) \Big]
    + \frac{\lambda}{2} \Vert \bx \Vert_2^2 ,
\]
where $\ba_i^{(j)} \in \RR^d$ and $b_i^{(j)}\in \{0, 1\}$ are samples stored at agent $i$.
For DANE and \texttt{Network-DANE}, we set $\mu = 5 \times 10^{-9}$ when $\kappa = 2$ and $\mu = 5 \times 10^{-1}$ when $\kappa = 100$.
The condition number is controlled by changing the regularization $\lambda$.
In both cases, our algorithms exhibit compelling performance over other decentralized optimization algorithms especially in terms of communication efficiency. 

\begin{figure}[htb]
\centering
\includegraphics[width=1\textwidth]{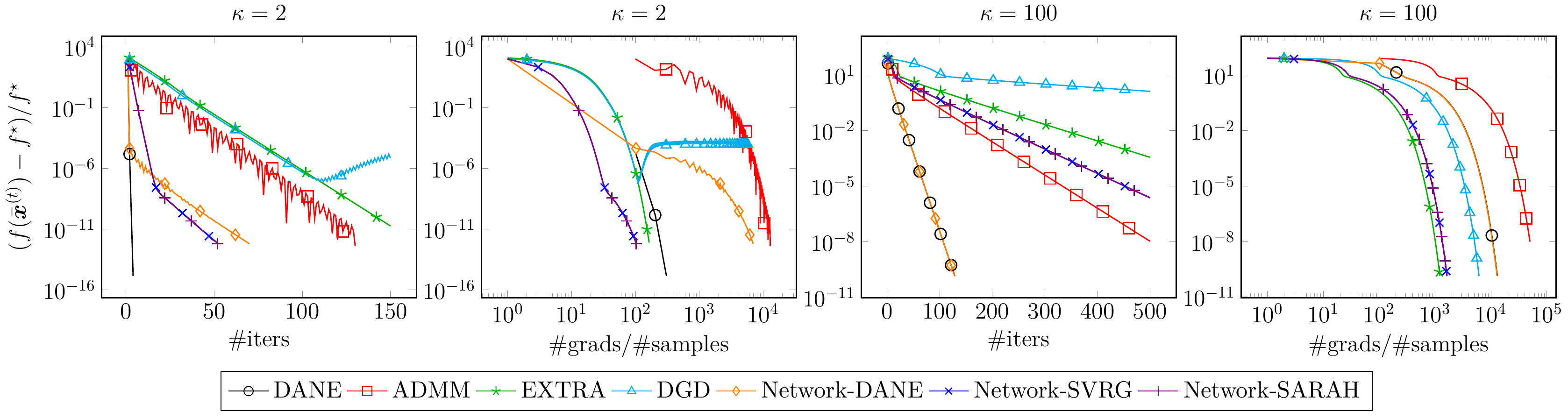}
\caption{The relative optimality gap with respect to the number of iterations and gradient evaluations under different conditioning $\kappa=2$ (left two panels) and $\kappa=100$ (right two panels) for logistic regression using the Gisette dataset.}
\label{fig:logistic_regression}
\end{figure}

\begin{figure}[htb]
	\centering
    \includegraphics[width=\linewidth]{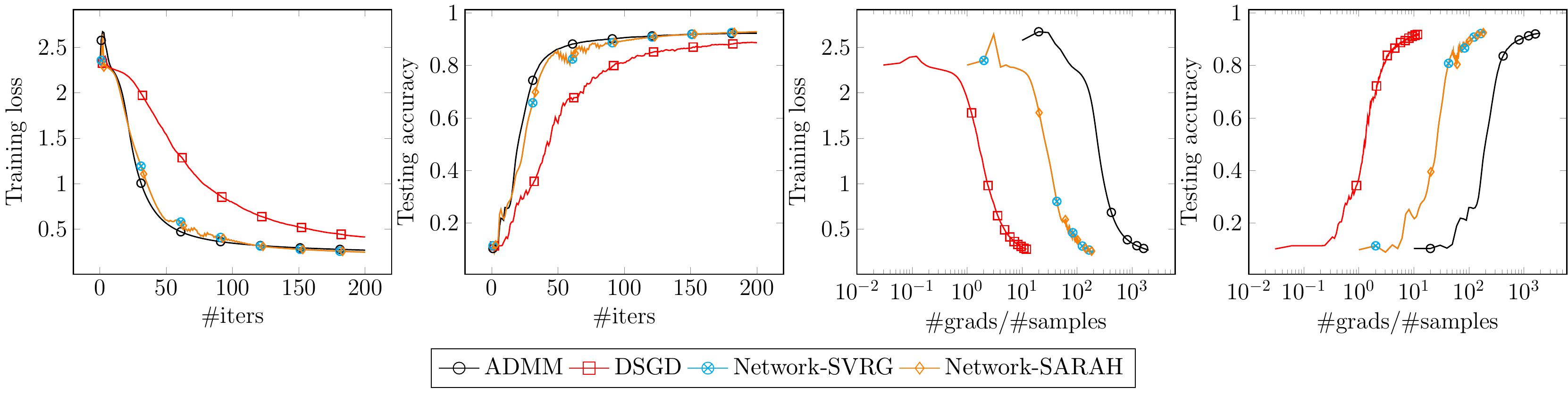}
	\caption{The training loss and testing accuracy with respect to the number of iterations (left two panels) and gradient evaluations (right two panels) for different algorithms on the MNIST dataset.}
	\label{fig:nn}
\end{figure}

\paragraph{Neural network training.} Though our theory only applies to the strongly convex case, we examine \texttt{Network-SVRG/SARAH} in the nonconvex case, by training a one-hidden-layer neural network with $64$ hidden neurons and sigmoid activations for a classification task using the MNIST dataset. We split $60,000$ training samples to $20$ agents and use an Erd\"{o}s-R\`{e}nyi graph with $p=0.3$ for communications. Fig.~\ref{fig:nn} plots the training loss and testing accuracy against the number of iterations and gradient evaluations for different algorithms, where centralized ADMM and decentralized stochastic algorithm (DSGD) are plotted as baselines.
Being more communication-efficient than DSGD, and more computation-efficient than ADMM,  \texttt{Network-SVRG/SARAH} reach a desirable balance between computation and communication efficacies.  

\section{Conclusions}
\label{sec:conclusions}

This paper proposes decentralized (stochastic) optimization algorithms that are communication-efficient over a network: (i) \texttt{Network-DANE} based on an approximate Newton-type local update, and (ii) \texttt{Network-SVRG/SARAH} based on stochastic variance-reduced local gradient updates. Theoretical convergence guarantees are developed for the proposed algorithms, highlighting the impact of network topology, data homogeneity across agents, and refined trade-offs between global communication and local computation.
Moreover, extensive numerical experiments are conducted to verify the superior performance of the proposed algorithms. The idea can be easily extended to obtain decentralized versions of other master/slave distributed algorithms in a systematic manner. This work opens up many exciting directions for future investigation, including but not limited to establishing the convergence of \texttt{Network-DANE} and \texttt{Network-SVRG/SARAH} under general loss functions for both convex and nonconvex settings, with the possibility of asynchronous updates across agents.

\section*{Acknowledgments}

The work of B.~Li, S.~Cen and Y.~Chi is supported in part by ONR under the grants N00014-18-1-2142 and N00014-19-1-2404, by ARO under the grant W911NF-18-1-0303, and by NSF under the grants CAREER ECCS-1818571, CCF-1806154, CCF-1901199 and CCF-2007911.
The work of Y. Chen is supported in part by the grants AFOSR YIP award FA9550-19-1-0030,
ONR N00014-19-1-2120, ARO YIP award W911NF-20-1-0097, ARO W911NF-18-1-0303, NSF CCF-1907661, DMS-2014279 and IIS-1900140, and the Princeton SEAS Innovation Award.

\bibliographystyle{alphaabbr}
\newcommand{\etalchar}[1]{$^{#1}$}

\newpage
 
\appendix

\section{Derivation of Equation~(\ref{eq:xt-expression-quadratic})} \label{sec:derivation_dane}

We make the observation that 
\[
f_{j}(\bm{x})-\big\langle\nabla f_{j}(\bbx^{(t)}),\bm{x}\big\rangle=\tfrac{1}{2}\bm{x}^{\top}\bm{H}_{j}\bm{x}-\bm{x}^{\top}\bm{H}_{j}\bbx^{(t)}+\text{constant}=\tfrac{1}{2}\big(\bm{x}-\bbx^{(t)}\big)^{\top}\bm{H}_{j}\big(\bm{x}-\bbx^{(t)}\big)+\text{constant},
\]
which allows us to derive a closed-form expression for $\bm{x}_{j}^{(t)}$ as follows
\begin{align*}
\bm{x}_{j}^{(t)} & =\arg\min_{\bm{x}\in\mathbb{R}^{d}}\;\left\{ \frac{1}{2}\big(\bm{x}-\bbx^{(t)}\big)^{\top}\bm{H}_{j}\big(\bm{x}-\bbx^{(t)}\big)+  \big\langle \nabla  f\big(\bbx^{(t)}\big) ,\bm{x}-\bbx^{(t)} \big\rangle+\frac{\mu}{2}\big\|\bm{x}-\bbx^{(t)}\big\|_{2}^{2}\right\} \\
 & =\arg\min_{\bm{x}\in\mathbb{R}^{d}}\;\left\{ \frac{1}{2}\big(\bm{x}- \bbx^{(t)} \big)^{\top}\left(\bm{H}_{j}+\mu\bm{I}\right)\big(\bm{x}-\bbx^{(t)} \big)+ \big\langle \nabla  f\big(\bbx^{(t)}\big) ,\bm{x}-\bbx^{(t)} \big\rangle \right\}  \\
 & = \bbx^{(t)}-  \left(\bm{H}_{j}+\mu\bm{I}_d\right)^{-1} \nabla  f\big(\bbx^{(t)}\big)  .
\end{align*}

\section{Proof of Theorem \ref{theorem:dane_convergence_quadratic} and Theorem \ref{corollary:dane_rate_varyingK_quadratic}}
\label{proof:convergence_quadratic}
This sections proves the convergence rate of \texttt{Network-DANE} for quadratic losses. When local and global loss functions are quadratic,
we can solve \eqref{eq:local_optimization} explicitly.
Specifically, Alg.~\ref{alg:network_dane} can be alternatively written as Alg.~\ref{alg:network_dane_quadratic} below.

\begin{algorithm}[htb]
\caption{\texttt{Network-DANE} for quadratic losses \eqref{eq:quadratic-problem}}
\label{alg:network_dane_quadratic}
\begin{algorithmic}[1]
	
    \FOR {$t = 1, 2, \cdots$} 
    \STATE
	\begin{subequations}
        \label{eq:quadratic_network_dane}
        \begin{align}
            \by^{(t)} & = \WK \bx^{(t-1)} ,
            \label{eq:quadratic_dane_update_y} \\
            \bs^{(t)} & =  \WK \bs^{(t-1)} + \bH\big( \by^{(t)}- \by^{(t-1)} \big) ,
            \label{eq:quadratic_dane_update_s} \\
            \bx^{(t)} & = \by^{(t-1)} - (\bH + \mu \bI_{nd})^{-1}\bs^{(t-1)} ,
            \label{eq:quadratic_dane_update_x}
        \end{align}
    \end{subequations} 
	where $\by^{(t)}$ and $\bs^{(t)}$ are defined in \eqref{eq:defn-xt-yt-st},
    $\bm H := \mbox{diag}(\bm{H}_{1},\cdots,\bm{H}_{n})\in\mathbb{R}^{nd\times nd}$,
    and $\bm{H}_i$ is defined in \eqref{eq:quadratic-problem}.
    \ENDFOR
    
\end{algorithmic}
\end{algorithm}
 
For notational convenience, we let $\overline{\bH} = \nabla^2 f(\bx) = \frac{1}{n}\sum_{j=1}^n \bH_j$ be the Hessian of the global loss function. From the definition of the homogeneity parameter $\beta$, we have $\| \overline{\bH}- \bH_j\|_2 \leq \beta $ for all $j=1,\ldots,n$. 
In addition, we recall the notations in \eqref{eq:defn-global-opt}, \eqref{eq:defn-xt-yt-st} and \eqref{eq:def_gradient}, and define the error vector as follows
\begin{align}
    \be^{(t)}
    = \begin{bmatrix}
    \sqrt n \| \bby^{(t)} - \by^\opt \|_2 \\ 
    \| \y^{(t)} - \onet \bby^{(t)} \|_2 \\ 
    L^{-1}\| \bs^{(t)} - \nabla f (\y^{(t)}) \|_2
\end{bmatrix}. \label{eq:dane_error_vector}
\end{align}

Establishing the convergence of \texttt{Network-DANE} relies on characterization of the per-iteration dynamics of $ \be^{(t)}$ for quadratic losses. Towards this end, 
we state the following key lemma --- which is established in Appendix~\ref{proof:dynamic_system_qudratic} --- that plays a crucial role in the analysis. 
\begin{lemma} \label{lemma:dynamic_system_qudratic}
Let $\eta = \frac{1}{\sigma + \mu}$ and $\gamma = \frac{L}{L + \mu}$. Suppose that Assumptions~\ref{assumption:strongly_convex_risk} and \ref{assumption:quadratic_risk} hold. Then one has
\begin{equation} \label{eq:dane_quadratic_original_g}
\be^{(t)}
\leq
\underbrace{ \begin{bmatrix}
    \theta_1 & \gamma\eta\beta + \eta\beta &  \eta^2 L \beta \\
    \alpha \gamma \eta \beta & \alpha + \alpha \eta L  & \alpha \eta L \\
    \frac\beta L + \theta_1 \frac\beta L + \alpha \gamma \eta \beta \frac\beta L & \alpha \frac\beta L + \alpha + 1 + \gamma\eta\beta \frac\beta L + \eta \beta \frac\beta L + \alpha\frac\beta L + \alpha\eta\beta & \alpha + \gamma\eta\beta \frac\beta L + \alpha\eta\beta
\end{bmatrix}}_{=: \,\bG}
\be^{(t-1)} .
\end{equation}
Here, $\bm{a}\leq \bm{b}$ indicates that $a_i\leq b_i$ for all entries $i$. 
\end{lemma}
In what follows, we invoke this result to establish  Theorem \ref{theorem:dane_convergence_quadratic} and Theorem \ref{corollary:dane_rate_varyingK_quadratic} separately. 

\subsection{Proof of Theorem \ref{theorem:dane_convergence_quadratic}}
\label{proof_dane_convergence_quadratic}
 
By the choice of $\mu$ stated in Theorem \ref{theorem:dane_convergence_quadratic}, we can show that
\begin{align} \label{eq:conditon_qudratic}
	\gamma < 1 \qquad \text{and} \qquad \eta \beta \leq \eta L < 1 .
\end{align}
In view of Lemma~\ref{lemma:dynamic_system_qudratic},
we can obtain $$ \be^{(t)} \leq \bG_1 \be^{(t-1)}$$ with a simplified matrix
\begin{equation}\label{eq:quadratic_convergence-linear}
\bG_1: = \begin{bmatrix}
    \theta_1 & 2\eta\beta &  \eta^2 L \beta \\
    \alpha\gamma\eta\beta  & \alpha + \alpha\eta L  & \alpha \eta L\\
    3 \frac\beta L & 7 & \alpha + 2 \eta\beta 
\end{bmatrix} ,
\end{equation}
where $\be^{(t)}$ is defined in \eqref{eq:dane_error_vector}. 
We first invoke an argument from \cite{wai2018multi} to show that $\be^{(t)}$ converges linearly at a rate not exceeding $\rho(\bG_1)$. Given that $\bG_1$ is a positive matrix (i.e.~all of its entries are strictly greater than zero),  one can invoke the Perron-Frobenius Theorem to show that: there exists a real-valued positive number $\rho(\bG_1) \in \mathbb R$ --- the spectral radius of $\bG_1$ --- such that
(i) $\rho(\bG_1)$ is an algebraically simple eigenvalue of $\bG_1$ associated with a strictly positive eigenvector $\boldsymbol{\chi}$,
(ii) all other eigenvalues of $\bG_1$ are strictly smaller in magnitude than $\rho(\bG_1)$. Therefore, there exists some constant $C>0$ such that  $\be_0 \leq C \boldsymbol{\chi} $, and consequently,
\begin{align}
	\be^{(1)} \leq \bG_1 \be^{(0)} \leq C \bG_1  \boldsymbol{\chi} = C \rho(\bG_1)\boldsymbol{\chi} .
\end{align}
Invoking this argument recursively for all $t$, we arrive at 
\begin{align}
	\be^{(t)} \leq C \big(\rho(\bG_1)\big)^t \boldsymbol{\chi} .
\end{align}
Therefore, the rest of this proof boils down to upper bounding $\rho(\bG_1)$. 
Rearrange the characteristic polynomial of $\bG_1$, given by
\begin{align}\label{eq:characteristic_quadratic}
    f_1(\lambda)
    =& \det\big( \lambda \bI - \bG_1 \big) \notag \\
    =& (\lambda - \theta_1) p_1(\lambda)
     + \alpha\gamma\eta^2\beta^2 ( 2 \alpha + 4 \eta \beta - 2 \theta_1 - 7 \eta L )
     - 3 \eta^2 \beta^2 ( \alpha - \alpha \eta L + \theta_1 ) ,
\end{align}
where $p_1(\lambda)$ is the following function obtained by direct computation
\begin{align} \label{eq:def_p1}
p_1(\lambda)
= (\lambda - \alpha - \alpha \eta L ) ( \lambda - \alpha - 2  \eta \beta ) - 7 \alpha \eta L - 2 \alpha\gamma\eta^2\beta^2 - 3 \eta^2\beta^2 .
\end{align}
From the Perron-Frobenius Theorem, we know that $\rho(\bG_1)$ is a simple positive root of $f_1(\lambda)$ (so that $f_1(\rho(\bG_1))=0$).
However, it is difficult to compute it directly.
In what follows, we seek to first upper bound $\rho(\bG_1)$ by 
\begin{equation} \label{eq:lambda_0_quadratic} 
  \rho_1 := \lambda_0 = \max \Bigg\{
        \frac{1 + \theta_1}{2},
        \alpha + \frac{140 \eta L}{1-\alpha} \Big( \frac{\beta}{\sigma} + 1 \Big),
        \frac{1 + \alpha}{2} + 2\eta\beta
    \Bigg\},
\end{equation}
and then demonstrate that $\lambda_0 < 1$, which in turn ensures linear convergence.

\paragraph{Step 1: bounding $\rho(\bG_1)$ by $\lambda_0$.} The following calculation
aims to verify the fact that: for all $\lambda \geq \lambda_0$, one has
$f_1(\lambda) > 0$,
and hence $\rho(\bG_1) \leq \lambda_0$. Recall the definition of $\theta_1$ in \eqref{eq:dane_quadratic_theta_def}.
When $\lambda \geq \lambda_0 \geq \frac{1 + \theta_1}{2}$, one has
\begin{align}
    \lambda - \theta_1
    \geq& \frac{1 - \theta_1}{2} \notag \\
    =& ~ \frac12 \frac{\sigma}{\sigma + \mu} \Big( 1 - \frac{L}{L + \mu} \frac{\beta}{\sigma + \mu - \beta} \frac{\beta}{\sigma} \Big) \notag \\
    \geq& ~ \frac14 \frac{\sigma}{\sigma + \mu} . \label{eq:lower_bound_quadratic_1}
\end{align}
In order for the last inequality to hold, we must make sure that
\begin{align}
\begin{cases}
	\sigma + \mu \geq \frac{3 \beta^2}{\sigma}, \qquad & \text{if }\beta \geq \sigma; \\
    \sigma + \mu \geq 3 \sigma, & \mbox{otherwise}.
\end{cases}
\label{eq:dane_condition_quadratic_1}
\end{align}
Note that the above relationship is guaranteed by the condition $\sigma + \mu \geq \frac{140 L}{(1 - \alpha)^2} \Big( \frac{\beta}{ \sigma } + 1 \Big)$.
When $\lambda \geq \lambda_0$,
using \eqref{eq:conditon_qudratic}, we can lower bound
the first term of $p_1(\lambda)$ by
\begin{align}
    (\lambda - \alpha - \alpha \eta L ) (\lambda - \alpha - 2 \eta \beta) 
    \geq& \frac{1-\alpha}{2} \Big(\frac{140 \eta L}{1 - \alpha} \Big( \frac{\beta}{ \sigma } + 1 \Big) - \alpha \eta L \Big) \notag \\
    >& 69 \eta L \Big( \frac{\beta}{ \sigma } + 1 \Big) . \notag
\end{align}
We can lower bound $p_1 (\lambda)$ by incorporating \eqref{eq:conditon_qudratic} as
\begin{align}
    p_1(\lambda)
    =&~ ( \lambda - \alpha - \alpha \eta L ) ( \lambda - \alpha - 2 \eta \beta ) - 7 \alpha \eta L - 2 \alpha\gamma\eta^2\beta^2 - 3 \eta^2\beta^2 \notag \\
    >&~ 69 \eta L \Big( \frac{\beta}{ \sigma } + 1 \Big) - 12 \eta L \notag \\
    >&~ 68 \kappa \eta \beta. 
    \label{eq:lower_bound_quadratic_2}
\end{align}

As a result of \eqref{eq:lower_bound_quadratic_1} and \eqref{eq:lower_bound_quadratic_2},
when $\lambda\geq \lambda_0$, 
the characteristic polynomial \eqref{eq:characteristic_quadratic} satisfies
\begin{align}
    f_1(\lambda)
    \geq&~ (\lambda - \theta_1) p_1(\lambda)
    + \alpha\gamma\eta^2\beta^2 ( 2 \alpha + 4 \eta \beta - 2 \theta_1 - 7 \eta L )
    - 3 \eta^2\beta^2 ( \alpha - \alpha \eta L + \theta_1 ) \notag \\
    >& ~\frac14 \eta\sigma \cdot 68 \kappa \eta \beta 
    - 9 \alpha\gamma\eta^2\beta^2
    - 3 \eta^2\beta^2 ( \alpha + \theta_1 ) \notag \\
    >&~ 17 \eta\beta \eta L
    - 9 \alpha\gamma\eta^2\beta^2
    - 6 \eta^2\beta^2
    > 0. \notag 
\end{align}
Therefore, any $\lambda$ that exceeds $\lambda_0$ cannot be a root of $f_1(\cdot)$.
This implies that the spectral radius $\rho(\bm{G}_1)$, of necessity, obeys $\rho(\bm{G}_1)< \lambda_0$.

\paragraph{Step 2: bounding $\lambda_0$.}
This step verifies that all three terms in \eqref{eq:lambda_0_quadratic} are smaller than $1$, thus leading to the conclusion $\lambda_0 < 1$.
\begin{itemize}
\item
First, observe that if \eqref{eq:dane_condition_quadratic_1} is satisfied,
we have
$\frac{1 + \theta_1}{2}
\leq 1 - \frac14 \eta \sigma
< 1$.
\item
When $\sigma + \mu \geq \frac{140 L}{(1 - \alpha)^2} \Big( \frac{\beta}{ \sigma } + 1 \Big)$,
the second term in \eqref{eq:lambda_0_quadratic} obeys $\alpha + \frac{140 \eta L}{1 - \alpha} \Big( \frac{\beta}{ \sigma } + 1 \Big) \leq 1$.
\item 
Finally, the third term in \eqref{eq:lambda_0_quadratic} is also less than $1$, since
$$\frac{1+\alpha}{2} + 2 \eta \beta
\leq \frac{1+\alpha}{2} + \frac{(1 - \alpha)^2}{70} \frac{\beta}{\frac{\beta}{ \sigma } + 1 } \frac1L
\leq \frac{1+\alpha}{2} + \frac{(1 - \alpha)^2}{70}
\leq 1 - \frac{1 - \alpha}{2} + \frac{1 - \alpha}{70} < 1.$$
\end{itemize}

\subsection{Proof of Theorem \ref{corollary:dane_rate_varyingK_quadratic}}
\label{proof:dane_rate_varying_k}

By the assumption $\sigma + \mu \geq 360 \sigma \Big( \frac{\beta^2}{\sigma^2} + 1 \Big)$ and $\alpha \leq \frac{1}{2\kappa}$,
we can prove that $\eta \beta < 1$ and $\alpha \eta L \leq \frac12$.
The characteristic polynomial \eqref{eq:characteristic_quadratic} in Appendix \ref{proof_dane_convergence_quadratic} can then be lower bounded by
\begin{align}
    f_1(\lambda)
    =& ~\det\left( \lambda \bI - \bG_1 \right) \notag \\
    =&  (\lambda - \theta_1) \Big( (\lambda - \alpha -  {\alpha \eta L} ) ( \lambda - \alpha - 2  \eta \beta ) -  {7 \alpha \eta L} -  {2 \alpha\gamma\eta^2\beta^2} - 3 \eta^2\beta^2 \Big) \notag \\
    &~+ \alpha\gamma\eta^2\beta^2 ( 2 \alpha + 4 \eta \beta - 2 \theta_1 - 7 \eta L )
    - 3 \eta^2 \beta^2 ( \alpha - \alpha \eta L + \theta_1 ) \notag \\
    \geq& ~(\lambda - \theta_1) \Big( ( \lambda - \alpha -  {\frac12 \eta\sigma} ) ( \lambda - \alpha - 2  \eta \beta ) -  {\frac72 \eta\sigma} -  {\eta\sigma \eta^2\beta^2} - 3 \eta^2 \beta^2 \Big) \notag \\
    &~+ \alpha\gamma\eta^2\beta^2 ( 2 \alpha + 4 \eta \beta - 2 \theta_1 - 7 \eta L )
    - 3 \eta^2 \beta^2 ( \alpha - \alpha \eta L + \theta_1 ) ,
    \label{eq:characteristic_3}
\end{align}
provided that $\lambda$ obeys
\[
    \lambda \geq \max \Bigg\{
        \frac{1 + \theta_1}{2},
        \alpha + 180 \eta\sigma \Big( \frac{\beta^2}{\sigma^2} + 1 \Big),
        \frac{1 + \alpha}{2} + 2\eta\beta
    \Bigg\}.
\]
Given that all conditions in \eqref{eq:dane_condition_quadratic_1} are satisfied,
we can show $\eta^2 \beta^2 \leq \eta\sigma \cdot \frac{\beta^2}{360 \sigma^2 (\beta^2 / \sigma^2 + 1)} < \eta \sigma < 1$.
One can thus continue to lower bound \eqref{eq:characteristic_3}  by
\begin{align*}
    f_1(\lambda)
    &>  (\lambda - \theta_1) \Big( ( \lambda - \alpha - \frac12 \eta\sigma ) ( \lambda - \alpha - 2  \eta \beta ) - 8 \eta\sigma \Big)
    -11 \eta^2\beta^2  \\
    & > \frac14 \eta\sigma \Big\{  \frac{1}{4} \Big[ 180 \eta\sigma \Big( \frac{\beta^2}{\sigma^2} + 1 \Big) - \frac12\eta\sigma \Big] - 8 \eta\sigma \Big\}
    -11 \eta^2\beta^2 \\
    & > \frac14 \eta\sigma \Big\{  45 \eta\beta \frac{\beta}{\sigma} + 44 \eta \sigma - 8 \eta\sigma \Big\}
    -11 \eta^2\beta^2 \\
    & > \frac{45}{4} \eta\beta - 11 \eta^2\sigma^2  \\
    & > 0 .
\end{align*}

Consequently, following similar arguments as in Appendix~\ref{proof_dane_convergence_quadratic},
we can show that: under the conditions of Theorem \ref{corollary:dane_rate_varyingK_quadratic},
the spectral radius of $\bG_1$ can be upper bounded by 
$$\rho(\bG_1) \leq 1 - \frac{C }{  \frac{\beta^2}{\sigma^2} + 1  },$$ 
where $C$ is some sufficiently small positive constant.
This immediately tells us that: to reach $\varepsilon$-accuracy, \texttt{Network-DANE} takes at most $O\left( \big( \frac{\beta^2}{\sigma^2} + 1 \big) \log (1/\varepsilon) \right)$ iterations.
For each iteration, \texttt{Network-DANE} needs $$K \asymp  \frac{ \log(1/2\kappa) }{ \log \alpha_0 } \lesssim  \frac{\log \kappa }{ 1 - \alpha_0 } $$ rounds of communication, where we have used the elementary inequality $1-\alpha_0 <\log(1/\alpha_0)$.
Putting all this together leads to a communication complexity at most $O\left( \log\kappa \cdot \frac{( \beta^2 / \sigma^2 + 1 ) \log (1/\varepsilon) }{ 1 - \alpha_0} \right)$.

\section{Proofs of Theorem \ref{theorem:dane_convergence} and Theorem \ref{corollary:dane_rate_varyingK}}
\label{proof:convergence}
This sections establishes the convergence rate of \texttt{Network-DANE} for smooth and strongly convex loss functions, following the analysis approach adopted in the proof of Theorem~\ref{theorem:dane_convergence_quadratic}. In particular, the following key lemma plays a crucial role, which characterizes the per-iteration dynamics of the proposed \texttt{Network-DANE} for general smooth strongly convex losses. The proof of this lemma is deferred to Appendix~\ref{proof:dynamic_system}.
\begin{lemma} \label{lemma:dynamic_system}
    Recall the notations in Lemma~\ref{lemma:dynamic_system_qudratic}. Suppose that Assumption~\ref{assumption:strongly_convex_risk} holds, and $\big( \frac{\beta}{\sigma + \mu} \big)^2 \leq \frac{\sigma}{\sigma + 2 \mu}$.
One has
 \begin{equation} 
    \label{eq:dane_original_g}
\be^{(t)}
\leq
\underbrace{    \begin{bmatrix}
        \theta_2 & \eta L & \gamma \eta L \\
        \alpha \gamma \eta L & \alpha + \alpha \eta L & \alpha \eta L \\ 
        \frac\beta L + \theta_2 \frac\beta L + \alpha \gamma \eta \beta
        & \alpha + 1 + \alpha \frac\beta L + \eta\beta + \alpha \frac\beta L + \alpha \eta\beta 
        & \alpha + \gamma \eta\beta + \alpha \eta\beta
    \end{bmatrix}}_{=: \,\bG^{\prime}}
\be^{(t-1)} .
\end{equation}
Here, $\be^{(t)}$ is the error vector defined in \eqref{eq:dane_error_vector}, and the notation $\bm{a}\leq \bm{b}$ indicates that $a_i\leq b_i$ for all entries $i$.
 \end{lemma}

\subsection{Proof of Theorem~\ref{theorem:dane_convergence}}
\label{proof:theorem_dane_convergence}

Under the conditions of Theorem \ref{theorem:dane_convergence}, the inequalities stated in  \eqref{eq:conditon_qudratic} remain valid.
In addition,
when $\sigma + \mu = \frac{170 \kappa L}{(1 - \alpha)^2}$,
we can verify that
\[
    \Big( \frac{\beta}{\sigma + \mu} \Big)^2
    = \frac{(1 - \alpha)^4 \beta^2}{170^2 \kappa^2 L^2}
    \leq \frac{(1 - \alpha)^2 }{170^2 \kappa^2}
    < \frac12 \cdot \frac{(1 - \alpha)^2}{170 \kappa^2}
    = \frac12 \cdot \frac{\sigma}{\sigma + \mu}
    < \frac{\sigma}{\sigma + 2 \mu}.
\]
When $\sigma + \mu \geq \frac{170 \kappa L}{(1 - \alpha)^2}$, the LHS decreases faster than the RHS,
thus the requirement of Lemma \ref{lemma:dynamic_system} is met.
In view of Lemma~\ref{lemma:dynamic_system} as well as the fact $\theta_2 \leq 1$, we can replace $\bG'$ by a simplified matrix that dominates $\bG'$:
\begin{equation}
\label{eq:convergence-linear}
\bG_2 := \begin{bmatrix}
    \theta_2 & 2\eta L &  \gamma\eta L \\
    \alpha\gamma\eta L& \alpha + \alpha\eta L  & \alpha \eta L\\
    3 \frac\beta L & 7 & \alpha + 2 \eta\beta 
\end{bmatrix} .
\end{equation} 
The above matrix $\bG_2$ is similar to $\bG_1$ in \eqref{eq:quadratic_convergence-linear} in the quadratic case,
except that the quantity $\beta$ in the first two  rows of $\bG_1$ is replaced by $L$ (thus leading to a worse convergence rate).

Similar to the proof of Theorem~\ref{theorem:dane_convergence_quadratic},
we shall upper bound $\rho(\bG_2)$ --- the spectral radius of $\bG_2$. To locate the eigenvalues of $\bG_2$, we rearrange the characteristic polynomial of $\bG_2$ as follows
\begin{align}\label{eq:characteristic}
    f_2(\lambda)
    =& \det\left( \lambda \bI - \bG_2 \right) \notag \\
    =& (\lambda - \theta_2) p_2(\lambda)
    + \alpha\gamma\eta^2 L^2 \left( 2 \alpha + 4 \eta \beta - 2 \theta_2 - 7 \gamma \right)
    - 3 \eta\beta \left( 2\alpha\eta L - \gamma ( \alpha + \alpha \eta L - \theta_2 ) \right),
\end{align}
where $p_2(\lambda)$ is the following function obtained by direct computation
$$p_2(\lambda)
= (\lambda - \alpha - \alpha \eta L) (\lambda - \alpha - 2  \eta \beta) - 7 \alpha \eta L - 2 \alpha\gamma\eta^2 L^2 - 3 \gamma\eta\beta .$$
From the Perron-Frobenius Theorem, $\rho(\bG_2)$ is a simple positive root of the equation $f_2(\lambda) = 0$. However, it is hard to calculate it directly.
In what follows, we seek to first upper bound $\rho(\bG_2)$ by 
\begin{equation}\label{eq:lambda_0_strongly_convex}
 \rho_2 :=   \lambda_0 = \max \left\{
        \frac{1 + \theta_2}{2}, ~
        \alpha + \frac{170 \kappa \eta L}{1 - \alpha},
        \frac{1+\alpha}{2} + 2\eta\beta
    \right\},
\end{equation}
and then demonstrate that $\lambda_0 < 1$, which in turn ensures linear convergence.

\paragraph{Step 1: bounding $\rho(\bG_2)$ by $\lambda_0$.} The following calculation
aims to verify the fact that $f_2(\lambda) > 0$ holds for all $ \lambda \geq \lambda_0$ ,
so that $\rho(\bG_2) \leq \lambda_0$.
Recalling the definition of $\theta_2$ in Lemma \ref{lemma:dynamic_system}, we see that
when $\lambda \geq \lambda_0 \geq \frac{1 + \theta_2}{2}$,
\begin{align}
    \lambda - \theta_2
    \geq&~ \frac{1 - \theta_2}{2} \notag \\
    =&~ \frac12 \eta \Big(\sigma - \beta \sqrt{ (1 - \eta\mu)(1 + \eta \mu)} \Big) \notag \\
    \geq& ~\frac12 \eta \Big(\sigma - \beta \sqrt{ 2 (1 - \eta\mu)} \Big)
    > \frac14 \eta \sigma, \label{eq:lower_bound_1}
\end{align}
where we have used the fact $\eta\mu < 1$ to reach the second inequality.
For the last inequality to hold, we need to make sure 
\begin{align}
	\begin{cases}
    \sigma + \mu \geq \frac{10\beta^2}{\sigma}, \qquad & \beta \geq \sigma \\
    \sigma + \mu \geq 10 \sigma, & \mbox{otherwise}
	\end{cases}
\label{eq:dane_condition_1}
\end{align}
which is guaranteed by the assumption $\sigma + \mu \geq \frac{170 \kappa L}{(1 - \alpha)^2}$.

Similarly, when $\lambda \geq \lambda_0$, the first term of $p_2(\lambda)$ can be lower bounded by
\begin{align}
    (\lambda - \alpha - \alpha \eta L ) (\lambda - \alpha - 2 \eta \beta) 
    \geq& \frac{1-\alpha}{2} \Big(\frac{170 \kappa \eta L}{1 - \alpha} - \alpha \eta L \Big) 
    > 80 \kappa \eta L . \notag
\end{align}
Then, using \eqref{eq:conditon_qudratic}
we can bound $p_2 (\lambda)$ by
\begin{align}
    p_2(\lambda)
    =&~ ( \lambda - \alpha - \alpha \eta L ) ( \lambda - \alpha - 2 \eta \beta ) - 7 \alpha \eta L - 2 \alpha\gamma\eta^2 L^2 - 3 \gamma\eta\beta \notag \\
    >&~ 80 \kappa \eta L - 12 \eta L  
    \geq 68 \kappa \eta L .
    \label{eq:lower_bound_2}
\end{align}
By virtue of  \eqref{eq:lower_bound_1} and \eqref{eq:lower_bound_2}, it is seen that
when $\lambda\geq \lambda_0$, the characteristic polynomial $ f_2(\lambda)$ in \eqref{eq:characteristic} satisfies
\begin{align*}
    f_2(\lambda)
    >& \frac14 \eta \sigma \cdot 68 \kappa \eta L
    - 8 \alpha \gamma \eta^2 L^2 
    - 9 \eta\beta \eta L > 0 .
\end{align*}
Therefore, any $\lambda$ that exceeds $\lambda_0$ cannot possibly be a root of $f_2(\cdot)$.
This implies that the spectral radius necessarily obeys $\rho(\bG_2)< \lambda_0$.

\paragraph{Step 2: bounding $\lambda_0$.}
This step verifies that the three terms in the expression of $\lambda_0$ in \eqref{eq:lambda_0_strongly_convex} is smaller than $1$, allowing us to conclude that $\lambda_0 < 1$.
\begin{itemize}
\item First, observe that if \eqref{eq:dane_condition_1} is satisfied,
then we have
$\frac{1 + \theta_2}{2}
\leq 1 - \frac14 \eta \sigma
< 1$.

\item When $\sigma + \mu \geq \frac{170 \kappa L}{(1 - \alpha)^2}$,
the second term is $\alpha + \frac{170 \kappa \eta L}{1 - \alpha} \leq 1$.

\item We conclude the proof by checking that the third term is also less than $1$, namely,  
$$\frac{1+\alpha}{2} + 2 \eta \beta
\leq \frac{1+\alpha}{2} + \frac{(1 - \alpha)^2}{85} \frac1\kappa \frac\beta L
\leq \frac{1+\alpha}{2} + \frac{(1 - \alpha)^2}{85}
\leq 1 - \frac{1 - \alpha}{2} + \frac{1 - \alpha}{85}.$$
\end{itemize}

\subsection{Proof of Theorem \ref{corollary:dane_rate_varyingK}}\label{proof:dane_rate_varyingK}

We first verify the assumption of Lemma \ref{lemma:dynamic_system}.
When $\sigma + \mu = 360 L \left( \frac{\beta}{\sigma} + 1 \right)$,
\[
    \Big( \frac{\beta}{\sigma + \mu} \Big)^2
    = \frac{\beta^2}{360^2 L^2 (\frac\beta\sigma + 1)^2}
    \leq \frac{\beta}{360^2 \kappa L (\frac\beta\sigma + 1)}
    < \frac12 \cdot \frac{1}{360 \kappa (\frac\beta\sigma + 1)}
    = \frac12 \cdot \frac{\sigma}{\sigma + \mu}
    < \frac{\sigma}{\sigma + 2 \mu}.
\]
Therefore, Lemma \ref{lemma:dynamic_system} still holds.

By the assumption $\alpha \leq \frac{1}{2\kappa}$,
we can further lower bound the characteristic polynomial \eqref{eq:characteristic} in Appendix \ref{proof:theorem_dane_convergence} as follows:
\begin{align}
    f_2(\lambda)
    =&~ \det\big( \lambda \bI - \bG_2 \big) \notag \\
    =&~ (\lambda - \theta_2) \left((\lambda - \alpha - \alpha \eta L) (\lambda - \alpha - 2  \eta \beta) - 7 \alpha \eta L - 2 \alpha\gamma\eta^2 L^2 - 3 \gamma\eta\beta\right) \notag\\
    &~ + \alpha\gamma\eta^2 L^2 \left( 2 \alpha + 4 \eta \beta - 2 \theta_2 - 7 \gamma \right)
    - 3 \eta\beta \left( 2\alpha\eta L - \gamma ( \alpha + \alpha \eta L - \theta_2 ) \right) \notag \\
    \geq&~ (\lambda - \theta_2) \Big( (\lambda - \alpha - \frac12 \eta\sigma ) ( \lambda - \alpha - 2  \eta \beta ) - \frac72 \eta\sigma - \eta\sigma\eta^2 L^2 - 3 \gamma\eta\beta \Big) \notag \\
    &~ - \eta\sigma\eta^2 L^2 ( \theta_2 + \frac72 \gamma ) \notag
    - 3 \eta\beta \Big( \eta\sigma + \gamma \theta_2 \Big) \notag \\
    >&~ (\lambda - \theta_2) \Big( (\lambda - \alpha - \frac12 \eta\sigma ) ( \lambda - \alpha - 2  \eta \beta ) - 8 \eta\sigma \Big)
    - 5 \eta\sigma \eta^2 L^2
    - 6 \eta \beta \eta L , \label{eq:corollary3_2}
\end{align}
providing $\lambda$ obeys
\[
    \lambda \geq \max \Bigg\{
        \frac{1 + \theta_2}{2},
        \alpha + 180 \eta L \Big( \frac{\beta}{\sigma} + 1 \Big),
        \frac{1 + \alpha}{2} + 2\eta\beta
    \Bigg\}.
\]
We can further lower bound \eqref{eq:corollary3_2} by
\begin{align*}
    f_2(\lambda)
    & \geq \frac14 \eta\sigma \left\{  \frac{1}{4} \Big[ 180 \eta L \Big( \frac{\beta}{\sigma} + 1 \Big) - \frac12\eta\sigma \Big] - 8 \eta\sigma \right\}
    - 5 \eta\sigma \eta^2 L^2
    - 6 \eta \beta \eta L > 0 ,
\end{align*}
as long as $\mu$ satisfies $\sigma + \mu \geq 360 L \big( \frac{\beta}{\sigma} + 1 \big)$. Therefore, following similar arguments as adopted in Appendix~\ref{proof:theorem_dane_convergence}, the spectral radius of $\bG_2$ can be upper bounded by $$\rho(\bG_2) \leq 1 - \frac{C}{   \kappa (\frac{\beta}{\sigma} + 1) },$$ where $C$ is a small positive constant.
Consequently, to reach $\varepsilon$-accuracy, \texttt{Network-DANE} takes at most $O\left( \kappa \big( \frac\beta\sigma + 1 \big) \log (1/\varepsilon) \right)$ iterations
and $O\left( \log\kappa \cdot \frac{\kappa ( \beta / \sigma + 1 ) \log (1/\varepsilon) }{ 1 - \alpha_0} \right)$ communication rounds.

\renewcommand{\norm}[1]{\|#1\|_2}
\section{Proof of Theorem \ref{theorem:svrg}}
\label{proof:svrg_and_sarah}

The proof strategy of Theorem \ref{theorem:svrg} is similar in spirit to the convergence proof of \texttt{Network-DANE}, where we will carefully build a linear system that tracks the coupling of the consensus error and the optimization error. Under the assumptions in Theorem \ref{theorem:svrg}, we can assume that $1-3\alpha\kappa-3 \beta / \sigma > 0$. Let  
$$\zeta = 1/(1-3\alpha\kappa-3 \beta / \sigma).$$ 

In what follows, we first introduce two key lemmas that connect the convergence behavior of \texttt{Network-SVRG} and \texttt{Network-SARAH} in the network setting to their master/slave counterparts (namely, D-SVRG and D-SARAH) studied in \cite{cen2019convergence}. 
Lemma \ref{lemma:svrg_dynamic},
proved in Appendix \ref{proof:svrg_dynamic},
creates the linear system characterizing the iteration dynamics
of \texttt{Network-SVRG}. Similarly, Lemma \ref{lemma:sarah_dynamic} describes the dynamics of \texttt{Network-SARAH}, whose proof can be found in Appendix \ref{proof:sarah_dynamic}.

\begin{lemma}
    \label{lemma:svrg_dynamic}
    Under the assumptions in Theorem \ref{theorem:svrg},
  \texttt{Network-SVRG} satisfies
    \begin{align}
        \EE [\be^{(t)}]
        \leq
        \underbrace{
        \begin{bmatrix}
            \paren{\nu(1+3\alpha\kappa+4\frac\beta\sigma) + \frac\beta\sigma}\zeta & 8 \frac\beta\sigma \zeta& \alpha\zeta/\kappa & \zeta/16 \\
            1/2 & 0 & 0 & 0 \\
            8 \big(\frac\beta\sigma\big)^2 & 64 \big(\frac\beta\sigma\big)^2 & 4\alpha^2& \alpha \kappa/2 \\
            64\alpha\kappa & 0 & 0 & 0 \\
        \end{bmatrix}
    }_{:= \bG_3}
        \EE [\be^{(t-1)}],
        \end{align}
    where the error vector is defined as
    \[
        \be^{(t)}
        = \begin{bmatrix}
            \sum_{j=1}^n \big( f(\bx_j^{(t)}) - f(\y^\opt) \big) \\
            \sum_{j=1}^n \big( f(\by_j^{(t)}) - f(\y^\opt) \big)/2 \\
            \| \bs^{(t)} - \nabla f(\by^{(t)}) \|_2^2 /\sigma \\
            32L \|\by^{(t)}-\onet \bby^{(t)} \|_2^2 / \alpha
        \end{bmatrix}.
    \]
Here, $\nu \leq \frac12 \frac{\sigma  - 2\beta}{\sigma  - 3\beta}$ is the convergence rate of D-SVRG in the master/slave setting under the same assumptions \cite[Theorem 1]{cen2019convergence}.
\end{lemma}

\begin{lemma}
    \label{lemma:sarah_dynamic}
    Under the assumptions of Theorem \ref{theorem:svrg}, and the loss functions are quadratic, \texttt{Network-SARAH} satisfies
	\begin{align}
    \EE [\be^{(t)}]
	\leq
    \underbrace{
	\begin{bmatrix}
	\paren{\nu(1+3\alpha\kappa+4 \frac\beta\sigma ) + \frac\beta\sigma}\zeta & 8\frac\beta\sigma  \zeta& 2\alpha\zeta/\kappa & \zeta/8 \\
	1/2 & 0 & 0 & 0 \\
	4 \big(\frac\beta\sigma\big)^2 & 32 \big(\frac\beta\sigma\big)^2 & 4\alpha^2& \alpha\kappa/2 \\
	32\alpha\kappa & 0 & 0 & 0 \\
	\end{bmatrix}
    }_{:=\bG_4}
    \EE[\be^{(t-1)}],
	\end{align}
    where the error vector is defined as
    \[
        \be^{(t)} 
        =\begin{bmatrix}
            \norm{\nabla f(\bx^{(t)})}^2\\
            \norm{\nabla f(\by^{(t)})}^2/2 \\
            \norm{\bs^{(t)} - \nabla f(\by^{(t)})}^2 \\
            32L^2\norm{\by^{(t)}-\onet \bby^{(t)}}^2/(\alpha\kappa)
        \end{bmatrix}.
    \]
 Here,  $\nu\leq \frac12 \frac{1}{1- 4 \beta^2/\sigma^2}$ is the convergence rate of D-SARAH in the master/slave setting under the same assumptions \cite[Theorem 2]{cen2019convergence}.
\end{lemma}

Since every term in the matrices of linear systems of Lemma \ref{lemma:svrg_dynamic} and Lemma \ref{lemma:sarah_dynamic} is non-negative,
all eigenvalues of $\bG_3$ and $\bG_4$ are bounded by the maximum of the sum of rows according to the Gershgorin circle theorem.
For \texttt{Network-SVRG}, by setting $\alpha = \frac{1}{70\kappa}$, which needs $K \asymp O(\log_{\alpha_0} 1/\kappa) = O \big(\log\kappa / (1-\alpha_0) \big)$, we can ensure that the sum of the first row is bounded by $5/6$,
and the sums of other rows are also bounded by a constant smaller than $1$, under the assumption $\beta\leq \sigma/200$. Therefore, invoking the Gershgorin circle theorem, the spectral radius is bounded by a constant smaller than $1$. To achieve $\varepsilon$-accuracy,
the total number of iterations needed is $O\left(\log(1/\varepsilon) \right)$
and thus the communication complexity is $O \left(\log \kappa \cdot \frac{\log(1/\varepsilon) }{1 - \alpha_0} \right)$. Similar arguments hold true for \texttt{Network-SARAH}, which we omit for simplicity.

\section{Proof of Lemma \ref{lemma:dynamic_system_qudratic}}
\label{proof:dynamic_system_qudratic}

The proof is divided into several steps.
(i) In Appendix \ref{sub:dane_quadratic_convergence_error}, we bound the convergence error $\sqrt n \| \bbx^{(t)} - \y^\opt \|_2$;
(ii) in Appendix~\ref{sub:dane_quadratic_consensus}, we bound the parameter consensus error $\| \x^{(t)} - \onet\bbx^{(t)} \|_2$;
(iii) in Appendix \ref{sub:dane_gradient_error}, we bound the gradient estimation error $\| \bs_j^{(t)} - \nabla f(\y^{(t)}) \|_2$;
(iv) finally, we create induction inequalities of $\| \y^{(t)} - \onet\bby^{(t)} \|_2$,
$\sqrt n \|\bby^{(t)} - \y^\opt \|_2$ and $\| \bs_j^{(t)} - \nabla f(\y^{(t)}) \|_2$
in Appendix \ref{sub:dane_quadratic_linear_system} to conclude the proof.

\subsection{Convergence error}
\label{sub:dane_quadratic_convergence_error}

We begin by defining an auxiliary variable $\x_j^+$,
which can be seen as the result of one local iterate \eqref{eq:local_optimization} of the original DANE algorithm initialized at $\bby^{(t-1)}$:
\begin{align}\label{eq:dane_definition_y+}
    & \x_j^+ = \argmin_{\x}\; \left\{ f_j(\x) - \left\langle \nabla f_j( \bby^{(t-1)} ) - \nabla f( \bby^{(t-1)} ), \x \right\rangle + \frac\mu2 \| \x - \bby^{(t-1)} \|_2^2 \right\} .
\end{align}
Following the same convention as in previous definitions,
we also define 
\begin{equation}\label{eq:def_x_+}
\bbx^+ = \frac1n \sum_j \x_j^+.
\end{equation}
Given that the function we optimize at each agent is strongly convex,
the local optimality conditions of \eqref{eq:dane_definition_y+} and \eqref{eq:local_optimization} are as follows:
\begin{subequations}
\label{eq:dane_opt_cond}
\begin{align}
    \nabla f_j (\x_j^+) + \mu (\bx_j^+ - \y^\opt)
    =& \nabla (f_j - f) (\bby^{(t-1)}) + \mu (\bby^{(t-1)} - \y^\opt) \label{eq:dane_opt_cond_x} ,\\
    \nabla f_j (\x_j^{(t-1)}) + \mu (\x_j^{(t-1)} - \y^\opt)
    =& \nabla f_j (\y_j^{(t-1)}) - \bs_j^{(t-1)} + \mu (\by_j^{(t-1)} - \y^\opt) .\label{eq:dane_opt_cond_y+}
\end{align}
\end{subequations}

Taking the average of \eqref{eq:dane_opt_cond} over $j=1,\ldots,n$,
we obtain another set of optimality conditions:
\begin{subequations}
\label{eq:dane_opt_cond_sum}
\begin{align}
    \frac1n \sum_j \nabla f_j (\x_j^+) + \mu (\bbx^+ - \y^\opt)
    =& \mu (\bby^{(t-1)} - \y^\opt) \label{eq:dane_opt_cond_sum_x}, \\
    \frac1n \sum_j \nabla f_j(\x_j^{(t-1)}) + \mu (\bbx^{(t-1)} - \y^\opt)
    =& \mu (\bby^{(t-1)} - \y^\opt) \label{eq:dane_opt_cond_sum_y+},
\end{align}
\end{subequations}
where we use the fact $\sum_j \bs_j^{(t-1)} = \sum_j \nabla f_j(\y_j^{(t-1)})$ due to the property of gradient tracking \eqref{eq:property_dac}.
    
In view of the triangle inequality, the convergence error can be decomposed as 
\begin{equation}
\label{eq:quadratic_convergence_error_tri}
\| \bbx^{(t-1)} - \y^\opt \|_2 \leq \| \bbx^{(t-1)} - \bbx^+ \|_2 + \| \bbx^+ - \y^\opt \|_2,
\end{equation}
where the first term is the error caused by inaccurate gradient estimate, and
the second term is the progress of DANE initialized at $\bby^{(t-1)}$.

\begin{enumerate}
    \item For the first term $ \| \bbx^{(t-1)} - \bbx^+ \|_2$,
        we first plug in the Hessian of the quadratic losses to solve for $\x_j^{(t-1)}$ and $\x_j^+$ explicitly as
\begin{subequations}
\begin{align}
    \x_j^{(t-1)} 
    =& \y_j^{(t-1)} - (\bH_j + \mu \bI_d)^{-1} \bs_j^{(t-1)},
    \label{eq:dane_quadratic_x_opt_cond} \\
    \x_j^+
    =& \bby^{(t-1)} - (\bH_j + \mu \bI_d)^{-1} \nabla f(\bby^{(t-1)}) .
    \label{eq:dane_quadratic_aux_opt_cond}
\end{align}
\end{subequations}
The first error term $ \| \bbx^{(t-1)} - \bbx^+ \|_2$ can be written as
\begin{align*}
    & \| \bbx^{(t-1)} - \bbx^+ \|_2 \notag \\
    =& \Big\| \ave ( \x^{(t-1)} - \x^+ ) \Big\|_2 \\
    =& \Big\| \ave \Big( \y^{(t-1)} - \onet\bby^{(t-1)} - (\bH + \mu \bI_{nd})^{-1} \nabla f(\onet\bby^{(t-1)}) + (\bH + \mu \bI_{nd})^{-1} \bs^{(t-1)} \Big) \Big\|_2 \\
    =& \Big\| \ave (\bH + \mu \bI_{nd})^{-1} \big( \bs^{(t-1)} - \nabla f(\onet\bby^{(t-1)}) \big) \Big\|_2 ,
\end{align*}
where the last line follows from the definition of $\bby^{(t-1)}$. Then, we add and subtract $( \bI_n\otimes\bbH + \mu \bI_{nd})^{-1}$ and rearrange terms, obtaining
\begin{align}
    & \| \bbx^{(t-1)} - \bbx^+ \|_2 \notag \\
    =& \Big\| \ave \Big( ( \bH + \mu \bI_{nd})^{-1} - ( \bI_n\otimes\bbH + \mu \bI_{nd})^{-1} \Big) \big(\bs^{(t-1)} - \nabla f(\onet\bby^{(t-1)}) \big) \notag \\
    & + \ave ( \bI_n\otimes\bbH + \mu \bI_{nd})^{-1} \big(\bs^{(t-1)} - \nabla f(\onet\bby^{(t-1)}) \big) \Big\|_2 \notag \\
    =& \Big\| \ave ( \bH + \mu \bI_{nd})^{-1} (\bI_n \otimes \bbH - \bH) ( \bI_n\otimes\bbH + \mu \bI_{nd})^{-1} \big(\bs^{(t-1)} - \nabla f(\y^{(t-1)}) \big) \notag \\
    &+ \ave ( \bH + \mu \bI_{nd})^{-1} (\bI_n \otimes \bbH - \bH) ( \bI_n\otimes\bbH + \mu \bI_{nd})^{-1} \big(\nabla f(\y^{(t-1)}) - \nabla f(\onet\bby^{(t-1)}) \big) \notag \\
    &+ \ave ( \bI_n\otimes\bbH + \mu \bI_{nd})^{-1} (\bH - \bI_n \otimes \bbH) (\y^{(t-1)} - \onet\bby^{(t-1)}) \Big\|_2  \label{eq:quadratic_consensus_y_W_3} \\
    \leq& \Big\| \ave \Big\|_2 \Big\| ( \bH + \mu \bI_{nd})^{-1} (\bI_n \otimes \bbH - \bH) ( \bI_n\otimes\bbH + \mu \bI_{nd})^{-1} \Big\|_2  \| \bs^{(t-1)} - \nabla f(\y^{(t-1)}) \|_2 \notag \\
    &+ \Big\| \ave \Big\|_2 \Big\| ( \bH + \mu \bI_{nd})^{-1} (\bI_n \otimes \bbH - \bH) \Big( \bI_{nd} + \mu \bI_{n}\otimes\bbH^{-1} \Big)^{-1} \Big\|_2  \| \y^{(t-1)} - \onet\bby^{(t-1)} \|_2 \notag \\
    &+ \Big\| \ave \Big\|_2 \Big\| ( \bI_n\otimes\bbH + \mu \bI_{nd})^{-1} ( \bH - \bI_n \otimes \bbH ) \Big\|_2 \| \y^{(t-1)} - \onet\bby^{(t-1)} \|_2 . \notag
\end{align}
The last term in \eqref{eq:quadratic_consensus_y_W_3} follows from the identity
\begin{align*}
    & \ave ( \bI_n\otimes\bbH + \mu \bI_{nd})^{-1} \big(\bs^{(t-1)} - \nabla f(\onet\bby^{(t-1)}) \big) \\
    =& ( \bbH + \mu \bI_{d})^{-1} \ave \big(\bs^{(t-1)} - \nabla f(\onet\bby^{(t-1)}) \big) \\
    =& ( \bbH + \mu \bI_{d})^{-1} \ave \big(\bH\y^{(t-1)} - \onet\bbH\bby^{(t-1)} \big) \\
    =& ( \bbH + \mu \bI_{d})^{-1} \ave \big(\bH\y^{(t-1)} - \onet\bH\bby^{(t-1)} \big) \\
    = & \ave ( \bI_n\otimes\bbH + \mu \bI_{nd})^{-1} \bH (\y^{(t-1)} - \onet\bby^{(t-1)} ) \notag \\
    = & \ave ( \bI_n\otimes\bbH + \mu \bI_{nd})^{-1} (\bH - \bI_n \otimes \bbH) (\y^{(t-1)} - \onet\bby^{(t-1)} ).
\end{align*}

Taken together with the identity $\| \frac1n \bm 1_n^\top \otimes \bI_d \|_2 = \frac{1}{\sqrt{n}}$,
the assumption $\| \bH_j - \bbH \|_2 \leq \beta$,
and the bound $\| ( \bH + \mu \bI_{nd})^{-1} \|_2 \leq \frac{1}{\sigma + \mu}$ and $\Big\| \big( \bI_{nd} + \mu \bI_{n}\otimes\bbH^{-1} \big)^{-1}  \Big\|_2 \leq \frac{L}{L + \mu}$,
we can further bound \eqref{eq:quadratic_consensus_y_W_3} by
\begin{align}
    \sqrt n \| \bbx^{(t-1)} - \bbx^+ \|_2
    \leq& \frac{1}{\sigma + \mu} \frac{\beta}{\sigma + \mu} \| \bs^{(t-1)} - \nabla f(\y^{(t-1)}) \|_2 \notag \\
    & + \Big( \frac{L}{L + \mu} \frac{\beta}{\sigma + \mu} + \frac{\beta}{\sigma + \mu} \Big) \| \y^{(t-1)} - \onet\bby^{(t-1)} \|_2 . \label{eq:dane_quadratic_convergence_error_1}
\end{align}

\item Regarding the second term $\| \bbx^+ - \y^\opt \|_2$, we provide a slightly improved bound compared to \cite{shamir2014communication}. In view of \eqref{eq:dane_quadratic_aux_opt_cond},
\begin{align}
    \| \bbx^+ - \y^\opt \|_2
    =& \Big\| \bby^{(t-1)} - \y^\opt - \frac1n \sum_j (\bH_j + \mu \bI_d)^{-1} \nabla f(\bby^{(t-1)}) \Big\|_2 \notag \\
    =& \Big\| \Big( \bI - \frac{1}{n} \sum_{i=1}^n ( \bH_i + \mu \bI)^{-1} \overline \bH \Big) ( \bby^{(t-1)} - \y^\opt ) \Big\|_2 \notag \\
    \leq& \Big\| \bI - \frac{1}{n} \sum_{i=1}^n ( \bH_i + \mu \bI)^{-1} \overline \bH \Big\|_2  \| \bby^{(t-1)} - \y^\opt \|_2 . \label{eq:quadratic_dane_convergence_poly}
\end{align}

Then, we use the triangle inequality to break the convergence rate in \eqref{eq:quadratic_dane_convergence_poly} into two parts:
\begin{align}
& \Big\| \bI - \frac{1}{n} \sum_{i=1}^n ( \bH_i + \mu \bI)^{-1} \overline \bH \Big\|_2  \nonumber \\
    \leq& \Big\| \bI - ( \overline \bH + \mu \bI)^{-1} \overline \bH \Big\|_2
    + \Big\| \frac{1}{n} \sum_{i=1}^n \Big( ( \bH_i + \mu \bI)^{-1} - ( \overline \bH + \mu \bI)^{-1} \Big) \overline \bH \Big\|_2.
    \label{eq:theta_quadratic-two-parts}
\end{align}
When $ \overline \bH \succeq \sigma \bI_d$, it is straightforward to check that the first term of \eqref{eq:theta_quadratic-two-parts} is upper bounded by
\begin{equation*}%
 \Big\| \bI - ( \overline \bH + \mu \bI)^{-1} \overline \bH \Big\|_2 \leq 1 - \frac{\sigma}{\sigma + \mu}.
\end{equation*}
Regarding the second term of \eqref{eq:theta_quadratic-two-parts}, let $\bm{\Delta}_i := \bH_i - \overline \bH$ and use the definition of $\beta$,  
one derives
\begin{align}
	\big\| (\overline \bH + \mu \bI)^{-1} \bm{\Delta}_{i} \big\|_2 \leq \big\| (\overline \bH + \mu \bI)^{-1}  \big\|_2 \cdot \big\| \bm{\Delta}_{i} \big\|_2 \leq  \frac{\beta}{\sigma + \mu} < 1
	\label{eq:Neumann-UB}
\end{align}
under our hypothesis $\beta< \mu + \sigma$. In addition, 
\begin{align}
&\quad\; \Big\| \frac{1}{n} \sum_{i=1}^n \Big( ( \bH_i + \mu \bI)^{-1} - ( \overline \bH + \mu \bI)^{-1} \Big) \overline \bH \Big\|_2 \nonumber \\
	&   = \Big\| \frac{1}{n} \sum_{i=1}^n \Big( \sum_{m=0}^{\infty} (-1)^m [( \overline \bH + \mu \bI)^{-1} \bm{\Delta}_i]^{m}( \overline \bH + \mu \bI)^{-1} - ( \overline \bH + \mu \bI)^{-1} \Big) \overline \bH \Big\|_2 \label{eq:Neumann} \\
&  = \Big\| \frac{1}{n} \sum_{i=1}^n \Big( \sum_{m=2}^{\infty} (-1)^m [( \overline \bH + \mu \bI)^{-1} \bm{\Delta}_i]^{m}( \overline \bH + \mu \bI)^{-1} \Big) \overline \bH \Big\|_2 \label{eq:fajdksfjaldsf} \\
&   \leq \frac{1}{n} \sum_{i=1}^n \sum_{m=2}^{\infty}  \| ( \overline \bH + \mu \bI)^{-1} \|_2^m \cdot \| \bm{\Delta}_i \|_2^{m} \cdot \big\| (\bI + \mu \overline \bH^{-1})^{-1} \big\|_2 \nonumber\\
&   \leq  \sum_{m=2}^{\infty}  (\sigma + \mu)^{-m} \beta^{m} \frac{L}{L + \mu}  = \frac{L}{L + \mu} \frac{\beta^2}{(\sigma + \mu) (\sigma + \mu - \beta)} .\nonumber
\end{align}
	Here, the line \eqref{eq:Neumann} is an expansion based on the Neumann series (whose convergence is guaranteed by \eqref{eq:Neumann-UB})
\begin{align*}
(\bm{H}_{i}+\mu\bm{I})^{-1} & =(\overline{\bm{H}}+\mu\bm{I}+\bm{\Delta}_{i})^{-1}=\big(\bm{I}+(\overline{\bm{H}}+\mu\bm{I})^{-1}\bm{\Delta}_{i}\big)^{-1}(\overline{\bm{H}}+\mu\bm{I})^{-1}\\
                            & =\Bigg\{ \sum_{m=0}^{\infty}(-1)^{m}\big[ (\overline{\bm{H}}+\mu\bm{I})^{-1}\bm{\Delta}_{i} \big]^m \Bigg\} (\overline{\bm{H}}+\mu\bm{I})^{-1}.
\end{align*}
The identity \eqref{eq:fajdksfjaldsf} holds since $\sum_{i=1}^n \bm{\Delta}_i = \bm{0}$, and hence the summation in \eqref{eq:fajdksfjaldsf} effectively starts at $m = 2$. 

Putting the above two bounds together back in \eqref{eq:theta_quadratic-two-parts}, we arrive at
\begin{align}
    \Big\| \bI - \frac{1}{n} \sum_{i=1}^n ( \bH_i + \mu \bI)^{-1} \overline \bH \Big\|_2 
    \leq & \theta_1
    = 1 - \frac{\sigma}{\sigma + \mu} + \frac{L}{L + \mu} \frac{\beta^2}{(\sigma + \mu)(\sigma + \mu - \beta)} . \label{eq:dane_quadratic_theta_bound}
\end{align}

\end{enumerate}

Putting together \eqref{eq:dane_quadratic_convergence_error_1} and \eqref{eq:dane_quadratic_theta_bound}, and plugging back into \eqref{eq:quadratic_convergence_error_tri},
we can bound the convergence error by:
\begin{align}
    \sqrt n \big\| \overline \by^{(t)} - \by^{\mathsf{opt}} \big\|_2
    =& \sqrt n \big\| \overline \bx^{(t-1)} - \by^{\mathsf{opt}} \big\|_2 \notag \\
    \leq& ~\theta_1 \sqrt n \big\| \overline \by^{(t-1)} - \by^{\mathsf{opt}} \big\|_2
    + \frac{1}{\sigma + \mu} \frac{\beta}{\sigma + \mu} \big\| \bs^{(t-1)} - \nabla f(\y^{(t-1)}) \big\|_2 \notag \\
    & + \Big( \frac{L}{L + \mu} \frac{\beta}{\sigma + \mu} + \frac{\beta}{\sigma + \mu} \Big) \big\| \y^{(t-1)} - \onet\bby^{(t-1)} \big\|_2 . \label{eq:quadratic_dane_convergence_convergence}
\end{align}

\subsection{Consensus error}%
\label{sub:dane_quadratic_consensus}
Using the identity $\overline \by^{(t)} = \ave \by^{(t)}$ and the update rule \eqref{eq:quadratic_dane_update_x}, we can demonstrate that
\begin{align}
    & \left\| \by^{(t)} - \onet \overline \by^{(t)} \right\|_2 \notag \\
    =& \left\|  \Big(\bI_{nd} - \frac{1}{n} \bm{1}_n \bm{1}_n^{\top} \otimes \bm{I}_d  \Big) \by^{(t)}  \right\|_2 \nonumber \\
    =& \left\| \Big(\bI_{nd} - \frac{1}{n} \bm{1}_n \bm{1}_n^{\top} \otimes \bm{I}_d \Big) \WK \left( \by^{(t-1)} - (\bH + \mu \bI_{nd})^{-1}\bs^{(t-1)}\right) \right\|_2 \notag \\
    \leq& \left\|  \left(\bW^K  - \frac{1}{n}\bm{1}_n \bm{1}_n^{\top} \right) \otimes \bm{I}_d \right\|_2
    \left\| \by^{(t-1)} - \onet\bby^{(t-1)} - \Big(\bI_{nd} - \frac{1}{n} \bm{1}_n \bm{1}_n^{\top} \otimes \bm{I}_d \Big)
    \left( (\bH + \mu \bI_{nd})^{-1}\bs^{(t-1)}\right)  \right \|_2 \label{eq:quadratic_consensus_y_W_00}  \\
    \leq& \alpha \| \y^{(t-1)} - \onet\bby^{(t-1)} \|_2
    + \alpha \left\| \Big(\bI_{nd} - \frac{1}{n} \bm{1}_n \bm{1}_n^{\top} \otimes \bm{I}_d \Big)
    (\bH + \mu \bI_{nd})^{-1}\bs^{(t-1)}  \right \|_2 , \label{eq:quadratic_consensus_y_W_0}
\end{align}
where \eqref{eq:quadratic_consensus_y_W_00} is due to the following equality: 
\[
\Big(\bI_{nd} - \frac{1}{n} \bm{1}_n \bm{1}_n^{\top} \otimes \bm{I}_d \Big) \WK 
= \left[ \left(\bW^K  - \frac{1}{n}\bm{1}_n \bm{1}_n^{\top} \right) \otimes \bm{I}_d \right] 
    \Big(\bI_{nd} - \frac{1}{n} \bm{1}_n \bm{1}_n^{\top} \otimes \bm{I}_d \Big),
\]
which holds because the property of the averaging operator $\left(\frac1n \bm{1}_{n}\bm{1}_{n}^{\top}\otimes\bm{I}_{d} \right)$,
\[
\left(\frac1n \bm{1}_{n}\bm{1}_{n}^{\top}\otimes\bm{I}_{d} \right)
\Big(\bI_{nd} - \frac{1}{n} \bm{1}_n \bm{1}_n^{\top} \otimes \bm{I}_d \Big)
= \Big[ \frac1n \bm{1}_{n}\bm{1}_{n}^{\top}
\Big(\bI_{d} - \frac{1}{n} \bm{1}_n \bm{1}_n^{\top} \otimes \bm{I}_d \Big) \Big] \otimes \bI_n
= \bm0,
\]
 and the fact that $(\bm{A}\otimes \bm{B})(\bm{C}\otimes \bm{D})=(\bm{A}\bm{C}) \otimes (\bm{B}\bm{D})$.

We rearrange the second term in \eqref{eq:quadratic_consensus_y_W_0} as
\begin{align}
    & \left\| \Big(\bI_{nd} - \frac{1}{n} \bm{1}_n \bm{1}_n^{\top} \otimes \bm{I}_d \Big)
    (\bH + \mu \bI_{nd})^{-1}\bs^{(t-1)}  \right \|_2 \notag \\
    =& \Big\| \Big(\bI_{nd} - \frac{1}{n} \bm{1}_n \bm{1}_n^{\top} \otimes \bm{I}_d \Big)
    (\bH + \mu \bI_{nd})^{-1} \Big( \bs^{(t-1)} - \nabla f(\y^{(t-1)}) \Big)  \notag \\
    &+ \Big(\bI_{nd} - \frac{1}{n} \bm{1}_n \bm{1}_n^{\top} \otimes \bm{I}_d \Big) (\bH + \mu \bI_{nd})^{-1}
    \Big( \nabla f(\y^{(t-1)}) - \nabla f(\onet\bby^{(t-1)}) \Big) \notag \\
    &+ \Big(\bI_{nd} - \frac{1}{n} \bm{1}_n \bm{1}_n^{\top} \otimes \bm{I}_d \Big) (\bH + \mu \bI_{nd})^{-1}
    \Big( \nabla f(\onet\bby^{(t-1)}) - \nabla f(\onet\y^\opt) \Big) \notag \Big\|_2 \\
    =& \Big\| \Big(\bI_{nd} - \frac{1}{n} \bm{1}_n \bm{1}_n^{\top} \otimes \bm{I}_d \Big)
    (\bH + \mu \bI_{nd})^{-1} \Big( \bs^{(t-1)} - \nabla f(\y^{(t-1)}) \Big) \notag \\
    &+ \Big(\bI_{nd} - \frac{1}{n} \bm{1}_n \bm{1}_n^{\top} \otimes \bm{I}_d \Big) 
    (\bH + \mu \bI_{nd})^{-1}
    (\bI_n\otimes\bbH) ( \y^{(t-1)} - \onet\bby^{(t-1)} )  \notag \\
    &+ \Big(\bI_{nd} - \frac{1}{n} \bm{1}_n \bm{1}_n^{\top} \otimes \bm{I}_d \Big) 
    \Big( (\bH + \mu \bI_{nd})^{-1} - (\bI_n\otimes\bbH + \mu \bI_{nd})^{-1} \Big)
    (\bI_n\otimes\bbH) ( \onet\bby^{(t-1)} - \onet\y^\opt ) \Big\|_2 . \notag
\end{align}
Using similar trick as in \eqref{eq:quadratic_consensus_y_W_3}, the above quantity can be further upper bounded as
\begin{align}
    & \left\| \Big(\bI_{nd} - \frac{1}{n} \bm{1}_n \bm{1}_n^{\top} \otimes \bm{I}_d \Big)
    (\bH + \mu \bI_{nd})^{-1}\bs^{(t-1)}  \right \|_2 \notag \\
    \leq & \Big\| \bI_{nd} - \frac{1}{n} \bm{1}_n \bm{1}_n^{\top} \otimes \bm{I}_d \Big\|_2
    \big\| (\bH + \mu \bI_{nd})^{-1} \big\|_2 \big\| \bs^{(t-1)} - \nabla f(\y^{(t-1)}) \big\|_2 \notag \\
    &+ \Big\| \bI_{nd} - \frac{1}{n} \bm{1}_n \bm{1}_n^{\top} \otimes \bm{I}_d \Big\|_2
    \Big\| ( \bH + \mu \bI_{nd})^{-1} \Big\|_2  \big\| \bI_n \otimes \bbH \big\|_2
    \| \y^{(t-1)} - \onet\bby^{(t-1)} \|_2 \notag \\
    &+ \sqrt n \Big\| \bI_{nd} - \frac{1}{n} \bm{1}_n \bm{1}_n^{\top} \otimes \bm{I}_d \Big\|_2
    \Big\| ( \bH + \mu \bI_{nd})^{-1} (\bI_n \otimes \bbH - \bH) ( \bI_{nd} + \mu \bI_n \otimes \bbH^{-1})^{-1} \Big\|_2
    \| \bby^{(t-1)} - \y^\opt \|_2.
    \label{eq:quadratic_convergence_2}
\end{align}

Combine \eqref{eq:quadratic_consensus_y_W_0} and \eqref{eq:quadratic_convergence_2},
we conclude that
\begin{align}
    \big\| \by^{(t)} - \onet \overline \by^{(t)} \big\|_2
    \leq& \left( \alpha + \frac{\alpha L}{\sigma + \mu} \right) \big\| \by^{(t-1)} - \onet \overline \by^{(t-1)} \big\|_2
    + \frac{\alpha}{\sigma + \mu} \big \| \bs^{(t-1)} - \nabla f(\y^{(t-1)}) \big \| _2 \nonumber \\
    & + \frac{\alpha L}{L + \mu} \frac{\beta}{\sigma + \mu}  \sqrt n \big\| \overline \by^{(t-1)} - \by^{\mathsf{opt}} \big\|_2 .
	\label{eq:quadratic_dane_linear_sys_1}
\end{align}

\subsection{Gradient estimation error}
\label{sub:dane_gradient_error}

In view of the fundamental theorem of calculus and the definition of $\beta$, it holds that 
$$\left\| \nabla (f - f_j) (\x) - \nabla (f - f_j) (\y) \right\|_2
= \left\| \left[ \int_0^1 \nabla^2 (f - f_j) \big( c \x + (1-c) \y \big) dc \right] (\x - \y) \right\|_2
\leq \beta \| \x - \y \|_2. $$

To begin, the update formulas \eqref{eq:network_dane_mixing} and \eqref{eq:network_dane_s_update}
are equivalent to
\begin{align}
    \y^{(t)} =& \WK \x^{(t-1)}, \label{eq:network_dane_y_update} \\
    \bs^{(t)} =& \WK \bs^{(t-1)} + \nabla F(\y^{(t)}) - \nabla F(\y^{(t-1)}) .\label{eq:network_dane_s_full_update}
\end{align}
Note that, since
\begin{align*} 
\left( \bm{W} -\tfrac{1}{n}\bm{1}_n\bm{1}_n^{\top} \right)^K & = \left( \bm{W} -\tfrac{1}{n}\bm{1}_n\bm{1}_n^{\top} \right)\left( \bm{W} -\tfrac{1}{n}\bm{1}_n\bm{1}_n^{\top} \right)  \cdots \left( \bm{W} -\tfrac{1}{n}\bm{1}_n\bm{1}_n^{\top} \right)\\
& = \left( \bm{W}^2 -\tfrac{1}{n}\bm{1}_n\bm{1}_n^{\top} \right)\cdots \left( \bm{W} -\tfrac{1}{n}\bm{1}_n\bm{1}_n^{\top} \right) = \bm{W}^{K}-\tfrac{1}{n}\bm{1}_n\bm{1}_n^{\top},
\end{align*}
we have the mixing rate of $\bm{W}^K$ is
$$\alpha : =   \| \bm{W}^K  -\tfrac{1}{n}\bm{1}_n\bm{1}_n^{\top} \| =   \| \bm{W}  -\tfrac{1}{n}\bm{1}_n\bm{1}_n^{\top} \|^K = \alpha_0^K. $$

In view of the equivalent update rule \eqref{eq:network_dane_s_full_update},
\begin{align*}
    \| \bs^{(t)} - \nabla f(\y^{(t)}) \|_2
    =& \Big\| \WK \bs^{(t-1)} + \nabla F(\y^{(t)}) - \nabla F(\y^{(t-1)}) - \nabla f(\y^{(t)}) \Big\|_2 \\
    =& \Big\| \WK \Big( \bs^{(t-1)} - \nabla f(\y^{(t-1)}) \Big) + \WK \nabla f(\y^{(t-1)}) \\
     &+ \nabla F(\y^{(t)}) - \nabla F(\y^{(t-1)}) - \nabla f(\y^{(t)}) \Big\|_2 \\
    =& \Big\| \WK \Big( \bs^{(t-1)} - \nabla f(\y^{(t-1)}) \Big)
     + \nabla (F - f) (\y^{(t)}) \\
     & + \WK \nabla f(\y^{(t-1)}) - \nabla F(\y^{(t-1)}) \Big\|_2
\end{align*}
Subtract and add
$\Big( \mean \Big) \Big( \bs^{(t-1)} - \nabla f(\y^{(t-1)}) \Big)$,
$\nabla (f - F) (\onet \bby^{(t)})$ and $\nabla (f - F) (\onet \y^{\mathsf{opt}})$
to the previous equation, and rearrange terms,
\begin{align}
    \| \bs^{(t)} - \nabla f(\y^{(t)}) \|_2
    =& \Big\| \Big[ \WK - \mean \Big] \Big( \bs^{(t-1)} - \nabla f(\y^{(t-1)}) \Big) \notag \\
     & + \nabla (F - f) (\y^{(t)}) - \nabla (F - f) (\onet \by^{\mathsf{opt}}) \notag \\
     & + \WK \Big(\nabla f(\y^{(t-1)}) - \nabla f(\onet \y^{\mathsf{opt}}) \Big) - \Big[ \nabla F(\y^{(t-1)}) - \nabla F( \onet \y^{\mathsf{opt}} ) \Big] \notag \\
     & + \Big[ \mean \Big] \Big( \bs^{(t-1)} - \nabla f(\y^{(t-1)}) \Big) \Big\|_2 \notag \\
    \leq& \alpha \big\| \bs^{(t-1)} - \nabla f(\y^{(t-1)}) \big\|_2
    + \beta \| \y^{(t)} - \onet\by^{\mathsf{opt}} \|_2 \notag \\
     & + \Big\| \WK \Big(\nabla f(\y^{(t-1)}) - \nabla f(\onet \y^{\mathsf{opt}}) \Big) - \Big[ \nabla F(\y^{(t-1)}) - \nabla F( \onet \y^{\mathsf{opt}} ) \Big] \notag \\
     &\quad\quad + \left[ \mean \right] \left( \bs^{(t-1)} - \nabla f(\y^{(t-1)}) \right) \Big\|_2 .\label{eq:dane_gradient_tracking_1}
\end{align}
Using the facts $ \Big[ \mean \Big] \bs^{(t-1)} = \Big[ \mean \Big] \nabla F(\y^{(t-1)})$
and $\Big[ \mean \Big] \nabla (F - f)(\onet \y^{\mathsf{opt}}) = \mathbf 0$, the last term of \eqref{eq:dane_gradient_tracking_1} becomes
\begin{align}
     & \Big\| \Big[ \WK - \mean \Big] \Big(\nabla (f - F) (\y^{(t-1)}) - \nabla (f - F) (\onet \bby^{(t-1)}) \Big) \notag \\
     & + \Big(\nabla (f - F) (\onet \bby^{(t-1)}) - \nabla (f - F) (\onet \y^{\mathsf{opt}}) \Big) \notag \\
     & + \Big[ \WK - \bI_{nd} \Big] \Big(\nabla F(\y^{(t-1)}) - \nabla F(\onet \bby^{(t-1)}) \Big) \Big\|_2 \notag  \\
    \leq& \Big\| \WK - \mean \Big\|_2 \| \nabla (f - F) (\y^{(t-1)}) - \nabla (f - F) (\onet \bby^{(t-1)}) \|_2 \notag \\
     & + \| \nabla (f - F) (\onet \bby^{(t-1)}) - \nabla (f - F) (\onet \y^{\mathsf{opt}}) \|_2 \notag \\
     & + \Big\| \WK - \bI_{nd} \Big\|_2 \| \nabla F(\y^{(t-1)}) - \nabla F(\onet \bby^{(t-1)}) \|_2 \notag \\
    \leq& \alpha \beta \| \y^{(t-1)} - \onet \bby^{(t-1)} \|_2
     + \beta \sqrt n \| \bby^{(t-1)} - \y^{\mathsf{opt}} \|_2
     + (\alpha + 1) L \| \y^{(t-1)} - \onet \bby^{(t-1)} \|_2. \label{eq:dane_gradient_tracking}
\end{align}
We used $\Big\| \WK - \bI_{nd} \Big\|_2 = \Big\| \WK - \ave + \ave - \bI_{nd} \Big\|_2 \leq \Big\| \WK - \ave \Big\|_2  + \Big\| \ave - \bI_{nd} \Big\|_2 \leq \alpha + 1$ to obtain the last inequality.

Combining \eqref{eq:dane_gradient_tracking_1} and \eqref{eq:dane_gradient_tracking}, we obtain the bound
\begin{align}
    \| \bs^{(t)} - \nabla f(\y^{(t)}) \|_2
    \leq& \alpha \big\| \bs^{(t-1)} - \nabla f(\y^{(t-1)}) \big\|_2
     + \beta \| \y^{(t)} - \onet \bby^{(t)} \|_2 + \beta \sqrt n \| \bby^{(t)} - \y^{\mathsf{opt}} \|_2 \notag \\ 
        &+ \big( \alpha \beta + (\alpha + 1) L\big) \| \y^{(t-1)} - \onet \bby^{(t-1)} \|_2
     + \beta \sqrt n \| \bby^{(t-1)} - \y^{\mathsf{opt}} \|_2 .
     \label{eq:gradient_tracking}
\end{align}

\subsection{Linear system}
\label{sub:dane_quadratic_linear_system}

Recall the definitions $\eta = \frac{1}{\sigma + \mu}$, $\gamma = \frac{L}{L + \sigma}$
and the error vector \eqref{eq:dane_quadratic_theta_def}.
Combining \eqref{eq:quadratic_dane_convergence_convergence}, \eqref{eq:quadratic_dane_linear_sys_1} and  \eqref{eq:gradient_tracking} leads to the matrix $\bG$ defined in \eqref{eq:dane_quadratic_original_g}.

\section{Proof of Lemma \ref{lemma:dynamic_system}}
\label{proof:dynamic_system}

The proof follows the same procedures as the proof of Lemma~\ref{lemma:dynamic_system_qudratic}.
(i) In Appendix \ref{sub:dane_convergence_error}, we bound the convergence error $\sqrt n \| \bby^{(t)} - \y^\opt \|_2$;
(ii) in Appendix~\ref{sub:dane_consensus}, we bound the parameter consensus error $\| \y^{(t)} - \onet\bby^{(t)} \|_2$;
(iii) finally,
using the bound we obtained in Appendix~\ref{sub:dane_gradient_error} of the gradient estimation error,
we create induction inequalities of $\| \y^{(t)} - \onet\bby^{(t)} \|_2$,
$\sqrt n \|\bby^{(t)} - \y^\opt \|_2$ and $L^{-1} \| \bs_j^{(t)} - \nabla f(\y^{(t)}) \|_2$
in Appendix \ref{sub:dane_linear_system} to conclude the proof. For consistency and simplicity, we use the same definitions of $\x^+$ in \eqref{eq:def_x_+}, $\eta = \frac{1}{\sigma + \mu}$, and $\gamma = \frac{L}{L + \sigma}$ as in the proof of Lemma~\ref{lemma:dynamic_system_qudratic}.

\subsection{Convergence error}
\label{sub:dane_convergence_error}

We continue to decompose the convergence error as \eqref{eq:quadratic_convergence_error_tri}, and bound the two terms respectively.

\begin{enumerate}
    \item For the term $ \| \bbx^{(t-1)} - \bbx^+ \|_2$,
        we first subtract \eqref{eq:dane_opt_cond_x} from \eqref{eq:dane_opt_cond_y+}, which gives
\begin{align}
\nabla f_j(\x_j^{(t-1)}) - \nabla f_j(\x_j^+) & + \mu (\x_j^{(t-1)} - \bx_j^+)
= \nabla f(\y_j^{(t-1)}) - \bs_j^{(t-1)} \nonumber \\
& \quad + \nabla (f - f_j) (\bby^{(t-1)}) - \nabla (f - f_j) (\y_j^{(t-1)}) + \mu (\y_j^{(t-1)} - \bby^{(t-1)}) ,\notag 
\end{align}
then use the strong convexity of $f_j(\cdot)$ and the definition of $\beta$ to bound both sides,
\begin{align*}
  &  \| \nabla f_j(\x_j^{(t-1)}) - \nabla f_j(\x_j^+) + \mu (\x_j^{(t-1)} - \bx_j^+) \|_2
    \geq (\sigma + \mu) \| \x_j^{(t-1)} - \x_j^+ \|_2,   \\
  &  \big\| \nabla f(\y_j^{(t-1)}) - \bs_j^{(t-1)}
    + \nabla (f - f_j) (\bby^{(t-1)}) - \nabla (f - f_j) (\y_j^{(t-1)}) + \mu (\y_j^{(t-1)} - \bby^{(t-1)}) \big\|_2 \\
    \leq& (\beta + \mu) \| \y_j^{(t-1)} - \bby^{(t-1)} \|_2 + \| \nabla f(\y_j^{(t-1)}) - \bs_j^{(t-1)} \|_2 .
\end{align*}
Therefore, combining the above two inequalities, we have
\begin{align}
    \| \x_j^{(t-1)}  - \x^+_j \|_2
    \leq& \frac{1}{\sigma + \mu} \| \nabla f (\y_j^{(t-1)} ) - \bs_j^{(t-1)}  \|_2 + \frac{\beta + \mu}{\sigma + \mu} \| \by_j^{(t-1)}  - \bby^{(t-1)}  \|_2 . \label{eq:dane_convergence_error_0}
\end{align}

Subtracting the optimality conditions in \eqref{eq:dane_opt_cond_sum},
\begin{align*}
    \bm 0
    &\in \frac1n \sum_j \nabla f_j(\x_j^{(t-1)}) - \frac1n \sum_j \nabla f_j(\x^+_j) + \mu (\bbx^{(t-1)} - \bbx^+) \\
    &= \frac1n \sum_j \big( \nabla f_j(\x_j^{(t-1)} ) - L \x_j^{(t-1)}  \big) - \frac1n \sum_j \left( \nabla f_j(\x^+_j) - L \x_j^+ \right)
    + (L + \mu) (\bbx^{(t-1)}  - \bbx^+) .
\end{align*}
Note the gradient of the function $L \x - \nabla f_j(\x)$ is a $(L - \sigma)$-Lipschitz function.
Taking the $\ell_2$ norm and plugging in \eqref{eq:dane_convergence_error_0}, we have
\begin{align}
    \| \bbx^{(t-1)} & - \bbx^+ \|_2
   \leq \frac{1}{L + \mu} \Big\| \frac1n \sum_j \left( \big[ L \x_j^{(t-1)} - \nabla f_j(\x_j^{(t-1)}) \big] - \big[L \x_j^+ - \nabla f_j(\x^+_j) \big] \right) \Big\|_2 \notag \\
    \leq& \frac{1}{L + \mu} \frac1n \sum_j \Big\| \big[ L \x_j^{(t-1)} - \nabla f_j(\x_j^{(t-1)}) \big] - \big[L \x_j^+ - \nabla f_j(\x^+_j) \big] \Big\|_2 \notag \\
    \leq& \frac{L - \sigma}{L + \mu} \frac1n \sum_j  \big\| \x_j^{(t-1)} - \x^+_j \big\|_2 \notag \\
    \leq& \frac{L - \sigma}{L + \mu} \frac{1}{\sigma + \mu} \frac1n \sum_j \big\| \nabla f (\y_j^{(t-1)} ) - \bs_j^{(t-1)}  \big\|_2
    + \frac{L - \sigma}{L + \mu} \frac{\beta + \mu}{\sigma + \mu} \frac1n \sum_j \big\| \y_j^{(t-1)}  - \bby^{(t-1)}  \big\|_2,
    \label{eq:dane_convergence_error_1}
\end{align}
where the last line follows \eqref{eq:dane_convergence_error_0}.
\item For the second term $\| \bbx^+ - \y^\opt \|_2$, 
    because of the assumption $\big( \frac{\beta}{\sigma + \mu} \big)^2 \leq \frac{\sigma}{\sigma + 2 \mu}$,
    we can invoke \cite[Theorem 3.1]{fan2019communication}, which is a careful analysis of the error of DANE, and bound the error as
\begin{align}
    \| \bbx^+ - \y^\opt \|_2
    \leq& \frac{\frac{\beta}{\sigma + \mu} \sqrt{\sigma^2 + 2 \sigma \mu} + \mu}{\sigma + \mu}  \| \bby - \y^\opt \|_2
    := \theta_2 \| \bby^{(t-1)} - \y^\opt \|_2 .\label{eq:dane_convergence_error_2}
\end{align}
\end{enumerate}

Putting together \eqref{eq:dane_convergence_error_1} and \eqref{eq:dane_convergence_error_2}, and plugging back into \eqref{eq:quadratic_convergence_error_tri},
we can bound the convergence error by:
\begin{align}
    \sqrt n \| \bby^{(t)} - \y^\opt \|_2
    =& \sqrt n \| \bbx^{(t-1)} - \y^\opt \|_2 \notag \\
    \leq&
    \theta_2 \sqrt n \| \bby^{(t-1)} - \y^\opt \|_2
    + \frac{1}{L + \mu} \frac{L}{\sigma + \mu} \| \nabla f (\y^{(t-1)}) - \bs^{(t-1)} \|_2 \notag \\
    & + \frac{\beta + \mu}{L + \mu} \frac{L}{\sigma + \mu} \| \y^{(t-1)} - \onet\bby^{(t-1)} \|_2.
    \label{eq:dane_convergence_convergence}
\end{align}

\subsection{Consensus error}%
\label{sub:dane_consensus}

Let $\bH_j^{(t)} = \int_0^1 \nabla^2 f_j \big( c \x_j^{(t)} + (1-c) \y_j^{(t)} \big) \dd c$
and $\bH^{(t)} = \mathrm{diag} (\bH_1^{(t)}, \bH_2^{(t)}, \ldots, \bH_n^{(t)})$.
Via the fundamental theorem of calculus, we can solve for $\x_j^{(t-1)}$ from the optimality condition \eqref{eq:dane_opt_cond_y+} as
\begin{align}
    \x_j^{(t-1)} = \y_j^{(t-1)} - (\bH_j^{(t-1)} + \mu \bI_d)^{-1} \bs_j^{(t-1)}.
\end{align}

Similar to \eqref{eq:quadratic_consensus_y_W_0},
we decompose the consensus error as
\begin{align}
    \| \y^{(t)} - \onet\bby^{(t)} \|_2
    \leq& \alpha \| \y^{(t-1)} - \onet\bby^{(t-1)} \|_2 + \alpha \Big\| \Big(\bI_{nd} - \mean \Big) (\bH^{(t-1)} + \mu \bI_{nd})^{-1}\bs^{(t-1)} \Big\|_2
    \label{eq:dane_consensus_1}
\end{align}
Then, we bound \eqref{eq:dane_consensus_1}.
Adding and subtracting terms and using the triangle inequality,
\begin{align}
    & \Big\| \Big(\bI_{nd} - \mean \Big) (\bH^{(t-1)} + \mu \bI_{nd})^{-1}\bs^{(t-1)} \Big\|_2 \notag \\
    \leq& \Big\| \Big(\bI_{nd} - \mean \Big) (\bH^{(t-1)} + \mu \bI_{nd})^{-1} \Big(\bs^{(t-1)} - \nabla f(\y^{(t-1)}) + \nabla f(\y^{(t-1)}) - \nabla f(\onet\bby^{(t-1)}) \Big) \Big\|_2 \notag \\
    &+ \Big\| \Big(\bI_{nd} - \mean \Big) (\bH^{(t-1)} + \mu \bI_{nd})^{-1}  \nabla f(\onet\bby^{(t-1)}) \Big\|_2
    \label{eq:dane_consensus_2}
\end{align}
We can bound the first term in \eqref{eq:dane_consensus_2} as
\begin{align}
    & \Big\| \Big(\bI_{nd} - \mean \Big) (\bH^{(t-1)} + \mu \bI_{nd})^{-1} \Big(\bs^{(t-1)} - \nabla f(\y^{(t-1)}) + \nabla f(\y^{(t-1)}) - \nabla f(\onet\bby^{(t-1)}) \Big) \Big\|_2 \notag \\
    \leq& \Big\| \Big( \bI_{nd} - \mean \Big) (\bH^{(t-1)} + \mu \bI_{nd})^{-1} \Big\|_2 \big\| \bs^{(t-1)} - \nabla f(\y^{(t-1)}) + \nabla f(\y^{(t-1)}) - \nabla f(\onet\bby^{(t-1)}) \big\|_2  \notag \\
    \leq& \frac{1}{\sigma + \mu} \Big( \| \bs^{(t-1)} - \nabla f(\y^{(t-1)}) \|_2 + \| \nabla f(\y^{(t-1)}) - \nabla f(\onet\bby^{(t-1)}) \|_2 \Big)  \notag \\
    \leq& \frac{1}{\sigma + \mu} \Big( \| \bs^{(t-1)} - \nabla f(\y^{(t-1)}) \|_2 + L \| \y^{(t-1)} - \onet\bby^{(t-1)} \|_2 \Big)
    \label{eq:dane_consensus_3}
\end{align}
Then, for the second term in \eqref{eq:dane_consensus_2},
\begin{align}
    & \Big\| \Big(\bI_{nd} - \mean \Big) (\bH^{(t-1)} + \mu \bI_{nd})^{-1} \nabla f(\onet\bby^{(t-1)}) \Big\|_2 \notag \\
    =& \Big\| \Big(\bI_{nd} - \mean \Big) \Big((\bH^{(t-1)} + \mu \bI_{nd})^{-1} - \big( (L + \mu) \bI_{nd} \big)^{-1} \Big) \nabla f(\onet\bby^{(t-1)}) \Big\|_2 \notag \\
    \leq& \Big\| (\bH^{(t-1)} + \mu \bI_{nd})^{-1} (L \bI_{nd} - \bH^{(t-1)}) \big( (L + \mu) \bI_{nd} \big)^{-1} \nabla f(\onet\bby^{(t-1)}) \Big\|_2 \notag \\
    \leq& \frac{L - \sigma}{L + \mu} \frac{L}{\sigma + \mu} \sqrt n \| \bby^{(t-1)} -   \y^\opt  \|_2
    \label{eq:dane_consensus_4}
\end{align}

Therefore, by combing \eqref{eq:dane_consensus_1}, \eqref{eq:dane_consensus_2}, \eqref{eq:dane_consensus_3} and \eqref{eq:dane_consensus_4},
we can bound the consensus error by:
\begin{align}
\| \y^{(t)} - \onet \bby^{(t)} \|_2
\leq& \Big( \alpha + \frac{\alpha L}{\sigma + \mu} \Big) \| \y^{(t-1)} - \onet\bby^{(t-1)} \|_2 \notag \\
&+ \frac{\alpha }{\sigma + \mu} \| \nabla f(\y^{(t-1)}) - \bs^{(t-1)} \|_2
+ \frac{\alpha L}{L + \mu} \frac{L}{\sigma + \mu} \sqrt n \| \bby^{(t-1)} -  \y^\opt  \|_2.
\label{eq:dane_convergence_consensus}
\end{align}

\subsection{Linear system}
\label{sub:dane_linear_system}

Combining \eqref{eq:gradient_tracking}, \eqref{eq:dane_convergence_consensus},
\eqref{eq:dane_convergence_convergence},
we reach the matrix claimed in \eqref{eq:dane_original_g}.

\renewcommand{\norm}[1]{\|#1\|_2}

\section{Proof of Lemma \ref{lemma:svrg_dynamic}}
\label{proof:svrg_dynamic}

The proof follows similar procedures as the proof of Lemma~\ref{lemma:dynamic_system_qudratic}.
(i) In Appendix \ref{appendix:svrg_convergence}, we bound the expected function value convergence errors $\EE \big[ \sum_{j=1}^n \big( f(\bx_j^{(t)}) - f(\y^\opt) \big) \big]$ and $\EE \big[ \sum_{j=1}^n \big( f(\by_j^{(t)}) - f(\y^\opt) \big) \big] $;
(ii) in Appendix~\ref{appendix:svrg_consensus}, we bound the expected parameter consensus error $\EE \| \y^{(t)} - \onet\bby^{(t)} \|_2^2$;
(iii) in Appendix~\ref{appendix:svrg_gradient_tracking}, we bound the expected parameter consensus error $\EE \| \y^{(t)} - \onet\bby^{(t)} \|_2^2$;
(iv) finally,
we create induction inequalities of $\EE \big[ \sum_{j=1}^n \big( f(\bx_j^{(t)}) - f(\y^\opt) \big) \big]$, $\EE \big[ \sum_{j=1}^n \big( f(\by_j^{(t)}) - f(\y^\opt) \big) \big]$,
$\EE \| \y^{(t)} - \onet\bby^{(t)} \|_2^2$ and $\EE \| \y^{(t)} - \onet\bby^{(t)} \|_2^2$ to conclude the proof.
Expectations in this section are conditioned on $\bx^{(t-1)}$, $\by^{(t-1)}$ and $\bs^{(t-1)}$, if not specified.

\subsection{Function value convergence error}
\label{appendix:svrg_convergence}

First, we bound the function value convergence error of $\y^{(t)}$ using the previous estimate $\x^{(t-1)}$.
By the strong convexity of $f(\cdot)$ and the assumption of $\alpha \le 1/\kappa$,
\begin{align}
    \sum_{j=1}^{n} f(\y_{j}^{(t)}) 
    \le& nf(\bby^{(t-1)}) + \frac{L}{2}\norm{\by^{(t)} - \onet \bby^{(t)}}^2 \notag\\
    \le& nf(\bbx^{(t-1)}) + \frac{\alpha^2L}{2}\norm{\bx^{(t-1)} - \onet \bbx^{(t)}}^2 \notag\\
    \le& nf(\bbx^{(t-1)}) + \frac{\sigma}{2}\norm{\bx^{(t-1)} - \onet \bbx^{(t)}}^2 \notag\\
    =& \sum_{j=1}^{n} \paren{f(\bbx^{(t-1)}) + \innprod{\nabla f(\bbx^{(t-1)}), \bx_j^{(t-1)}-\bbx^{(t-1)}}+\frac{\sigma}{2}\norm{\bx_j^{(t-1)} -  \bbx^{(t)}}^2} \notag \\
    \leq& \sum_{j=1}^{n} f(\x_{j}^{(t-1)}) . \label{eq:svrg_convergence_1}
\end{align}

Next, we bound the function value convergence error after local update, $\sum_{j=1}^n \big( f(\by_j^{(t)}) - f(\y^\opt) \big)$.
By constructing the following helper function,
we can connect local updates of \texttt{Network-SVRG} to that of D-SVRG \cite{cen2019convergence},
which is the counterpart of SVRG in the master/slave setting. For agent $j$ at the $t$th time,
we define the corrected sample loss function as
\begin{align}
    \tilde\ell^{(j)}(\x; \bz)
    = \ell(\x; \bz)
    + \left\langle \bs_j^{(t)}-\nabla f(\by_j^{(t)}), \x-\y_j^{(t)} \right\rangle. \notag
\end{align}
Then, define the corrected local and global loss functions as
\begin{align}
    h_{i}^{(t,j)}(\x)
    & = \frac1m \sum_{\bz\in \cM_i} \tilde\ell^{(j)}(\x; \bz) 
    = f_i(\x)
    + \left\langle \bs_j^{(t)}-\nabla f(\by_j^{(t)}), \x-\y_j^{(t)} \right\rangle, \notag \\
    h^{(t,j)}(\x)
    & = \frac1n \sum_i h_{i}^{(t,j)}(\x)
    = f(\x)
    + \left\langle \bs_j^{(t)}-\nabla f(\by_j^{(t)}), \x-\y_j^{(t)} \right\rangle.
    \label{eq:svrg_helper_function}
\end{align}
Here, $h^{(t,j)}(\cdot)$ and $h_{i}^{(t,j)}(\cdot)$ are $\sigma$-strongly convex and $L$-smooth functions,
and $\big\| h_{i}^{(t,j)}(\x) - h^{(t,j)}(\x) \big\|_2 \leq \beta$ by the definition of $\beta$.
Let ${h_*^{(t,j)}}$ denote the optimum value of $h^{(t,j)}(\cdot)$.

The key observation is that the local update \eqref{eq:svrg_local_v} at agent $j$ is the same as the update at agent $j$ when applying D-SVRG to optimize $h^{(t,j)}$ initialized with $\y_j^{(t)}$. This is true because $\forall \bz \in \cM_j$, the sample gradient and global gradient used in D-SVRG updates at $\y_j^{(t)}$ satisfy
\begin{align}
& \nabla \tilde\ell^{(j)}(\bu; \bz) - \nabla \tilde\ell^{(j)}(\bu'; \bz )
= \nabla \ell( \bu'; \bz ) - \nabla \ell( \bu; \bz ) , \quad\mbox{and}\quad \nabla h^{(t,j)}(\y_j^{(t)}) = \bs_j^{(t)} , \notag
\end{align}
which agree with \eqref{eq:svrg_local_v}.
Therefore, we can apply \cite[Theorem 1]{cen2019convergence}
to bound the optimization error of optimizing $h^{(t,j)}$
\begin{align}
    \EE\Big[ h^{(t,j)}(\bx_j^{(t)})- {h_*^{(t,j)}} \Big] < \nu \Big( h^{(t,j)}(\by_j^{(t)})- {h_*^{(t)}} \Big),
    \label{eq:svrg_h_descent}
\end{align}
where $\x_j^{(t)}$ is the output at agent $j$ produced by running one iteration of Alg.~\ref{alg:network_svrg},
which is also the output of running one iteration of D-SVRG at the same agent,
$\nu$ is the convergence rate of D-SVRG,
which can be bounded by $\nu \leq 1 - \frac12 \frac{\sigma - 2\beta}{\sigma - 3\beta}$
when choosing step size $\delta = \frac{1}{40 L} \big( 1 - \frac{4 \beta}{\sigma} \big)$
and the number of local updates $S = 160 \frac{L}{\sigma} \big( 1 - \frac{4\beta}{\sigma } \big)^{-2}$.

Next, we relate function value descent of $h^{(t,j)}$ to the function value descent of $f$.
Plug in \eqref{eq:svrg_helper_function} and rearrange terms,
\begin{align}
    f(\bx_j^{(t)}) - f(\y^\opt) \notag
    =& h^{(t,j)}(\bx_j^{(t)}) - (1-\nu)f(\y^\opt) - \nu f(\y^\opt) - \inner{\bs_j^{(t)}-\nabla f(\by_j^{(t)})}{\bx_j^{(t)}-\by_j^{(t)}} \notag \\
    =& h^{(t,j)}(\bx_j^{(t)}) - (1-\nu) h^{(t,j)}(\by^{\mathsf{opt}}) - \nu f(\y^\opt) \notag \\
    &- \inner{\bs_j^{(t)}-\nabla f(\by_j^{(t)})}{\bx_j^{(t)}-\by_j^{(t)}-(1-\nu)\paren{\by^{\mathsf{opt}}-\by_j^{(t)}}} \notag \\
    \leq& h^{(t,j)}(\bx_j^{(t)}) - (1-\nu)  {h_*^{(t,j)}} - \nu f(\y^\opt) \notag \\
    &- \inner{\bs_j^{(t)}-\nabla f(\by_j^{(t)})}{\bx_j^{(t)}-\by_j^{(t)}-(1-\nu)\paren{\by^{\mathsf{opt}}-\by_j^{(t)}}} \notag \\
    =& h^{(t,j)}(\bx_j^{(t)}) -  {h_*^{(t,j)}} + \nu \Big( {h_*^{(t,j)}} - f(\y^\opt) \Big) \notag \\
     & - \inner{\bs_j^{(t)} - \nabla f(\by_j^{(t)})}{\bx_j^{(t)}-\by_j^{(t)}-(1-\nu)\paren{\by^{\mathsf{opt}}-\by_j^{(t)}}}, \notag
\end{align}
where we used $h^{(t,j)}(\y^\opt) \geq {h_*^{(t,j)}}$ and $\nu \leq 1$ to reach the last inequality.

Taking expectation on both sides and combining with \eqref{eq:svrg_h_descent},
we reach the following function value descent of $f(\cdot)$:
\begin{align}
    \ex{f(\bx_j^{(t)}) - f(\y^\opt)}
    \le& \nu\paren{ h^{(t,j)}(\by_j^{(t)}) -  {h_*^{(t,j)}}} + \nu \Big( {h_*^{(t,j)}} - f(\y^\opt) \Big) \notag \\
    & - \EE \Big[ \inner{\bs_j^{(t)}-\nabla f(\by_j^{(t)})}{\bx_j^{(t)}-\by_j^{(t)}-(1-\nu) \paren{\by^{\mathsf{opt}}-\by_j^{(t)}}} \Big] \notag \\
    =& \nu\paren{f(\by_j^{(t)}) - f(\y^\opt)} - \ex{\inner{\bs_j^{(t)}-\nabla f(\by_j^{(t)})}{\bx_j^{(t)}-\by^{\mathsf{opt}}-\nu(\by_j^{(t)}-\by^{\mathsf{opt}})}} , \notag
\end{align}
where the last line follows from \eqref{eq:svrg_helper_function}.
Summing the previous inequality over all agents and using matrix notations,
we obtain the following inequality 
\begin{align}
    \EE \Bigg[ \sum_{j=1}^n f(\bx_j^{(t)}) - f(\y^\opt) \Bigg]
    \le & \nu \Bigg[ \sum_{j=1}^n f(\y_j^{(t)}) - f(\y^\opt) \Bigg]
        - \ex{\innprod{\bs^{(t)} - \nabla f(\by^{(t)}),\bx^{(t)}- \onet\by^{\mathsf{opt}}}} \notag \\
        & + \nu \ex{\innprod{\bs^{(t)} - \nabla f(\by^{(t)}), \by^{(t)}- \onet\by^{\mathsf{opt}}}}. \label{eq:svrg_convergence_5}
\end{align}

Our next step is to carefully bound the last two error terms in \eqref{eq:svrg_convergence_5}.
\begin{align}
    & \Big|\innprod{\bs^{(t)} - \nabla f(\by^{(t)}),\bx^{(t)}- \onet\by^{\mathsf{opt}}}\Big| \notag \\
    \le& \norm{\bs^{(t)} - \nabla f(\by^{(t)})}\norm{\bx^{(t)}- \onet\by^{\mathsf{opt}}} \notag \\
    \le& \Big( \alpha \norm{\bs^{t-1}- \nabla f(\by^{(t-1)})} + 2L\norm{\by^{(t-1)}- \onet\bby^{(t-1)}} \notag \\
    & + 2\beta \norm{\by^{(t-1)}- \onet\by^{\mathsf{opt}}} + \beta \norm{\by^{(t)}- \onet\by^{\mathsf{opt}}} \Big) \norm{\bx^{(t)}- \onet\by^{\mathsf{opt}}} \notag \\
    \le& \frac12 \alpha L^{-1} \norm{\bs^{t-1} - \nabla f(\by^{(t-1)})}^2 + \alpha^{-1}L \norm{\by^{(t-1)}- \onet\bby^{(t-1)}}^2+ \frac32 \alpha L \norm{\bx^{(t)}- \onet\by^{\mathsf{opt}}}^2 \notag \\
    & + \beta \| \by^{(t-1)}- \onet\by^{\mathsf{opt}} \|_2^2 + \frac\beta2 \| \by^{(t)}- \onet\by^{\mathsf{opt}} \|_2^2
    + \frac{3\beta}{2} \| \bx^{(t)}- \onet\by^{\mathsf{opt}}\|_2^2, \label{eq:svrg_convergence_2}
\end{align}
where the first inequality is due to \eqref{eq:svrg_gradient_error},
and the last inequality is obtained by Cauchy-Schwarz inequality.
Similar to \eqref{eq:svrg_convergence_5},
because of the strong convexity of loss functions,
we have 
\[
    \| \y^{(t)} - \onet\y^\opt \|_2^2
    \leq \frac2\sigma \sum_j \Big( f(\y_j^{(t)}) - f(\y^\opt) \Big).
\]
Then, we can further bound \eqref{eq:svrg_convergence_2} as
\begin{align}
    \Big| \innprod{\bs^{(t)} - \nabla f(\by^{(t)}),\bx^{(t)}-\by^\opt} \Big|
    \leq& \frac{1}{2}\alpha L^{-1} \| \bs^{t-1} - \nabla f(\by^{(t-1)})\|_2^2 + \alpha^{-1}L \| \by^{(t-1)}-\bby^{(t-1)}\|_2^2 \notag \\
    +& \frac{2\beta}{\sigma} \sum_{j=1}^n \Big(f(\by_j^{(t-1)})-f(\y^\opt) \Big) 
    +\frac{\beta}{\sigma} \sum_{j=1}^n \Big( f(\bx_j^{(t-1)})-f(\y^\opt) \Big) \notag \\
    +& \Big( \frac{3\beta}{\sigma}+3\kappa \alpha \Big) \sum_{j=1}^n \Big(f(\bx_j^{(t)})-f(\y^\opt) \Big). 
    \label{eq:svrg_convergence_3}
\end{align}
Similarly, we have the same bound applicable for the last term of \eqref{eq:svrg_convergence_5}:
\begin{align} 
\label{eq:svrg_convergence_4}
\Big|\innprod{\bs^{(t)} - \nabla f(\by^{(t)}), \by^{(t)}-\by^{\mathsf{opt}}} \Big|
\leq& \frac{1}{2} \alpha L^{-1} \norm{\bs^{t-1}- \nabla f(\by^{(t-1)})}^2 + \alpha^{-1}L\norm{\by^{(t-1)}-\bby^{(t-1)}}^2 \notag \\
+& \frac{2\beta}{\sigma} \sum_{j=1}^n \paren{f(\by_j^{(t-1)})-f(\y^\opt)} +\frac{\beta}{\sigma} \sum_{j=1}^n \paren{f(\bx_j^{(t-1)})-f(\y^\opt)} \notag \\
+& \paren{\frac{3\beta}{\sigma}+3\kappa \alpha} \sum_{j=1}^n \paren{f(\bx_j^{(t-1)})-f(\y^\opt)},
\end{align}
where the last term is due to \eqref{eq:svrg_convergence_1}.

Put together \eqref{eq:svrg_convergence_2}, \eqref{eq:svrg_convergence_3} and \eqref{eq:svrg_convergence_4} and taking expectation,
we reach the following bound
\begin{align} \label{eq:svrg_bound_system}
\ex{\sum_{j=1}^n\paren{f(\bx_j^{(t)}) - f(\y^\opt)}}
\le& \Big( \nu \big( 1+3\alpha\kappa+\frac{4\beta}{\sigma} \big)+ \frac\beta\sigma \Big) \sum_{j=1}^n\paren{f(\bx_j^{(t-1)}) - f(\y^\opt)} \notag \\
+& \alpha L^{-1} \norm{\bs^{t-1}- \nabla f(\by^{(t-1)})}^2 + 2\alpha^{-1}L\norm{\by^{(t-1)}-\bby^{(t-1)}}^2 \notag \\
+& \frac{4\beta}{\sigma} \sum_{j=1}^n \paren{f(\by_j^{(t-1)})-f(\y^\opt)}
+\paren{\frac{3\beta}{\sigma} +3\kappa \alpha} \ex{\sum_{j=1}^n \paren{f(\bx_j^{(t)})-f(\by^\opt)}}. 
\end{align}
Rearranging terms, we proved the advertised bound.

\subsection{Consensus error}
\label{appendix:svrg_consensus}

We first bound the consensus error $\| \y^{(t)} - \onet\bby^{(t)} \|_2^2 / (\alpha L)$.
Similar to \eqref{eq:quadratic_consensus_y_W_0},
\begin{align}\label{eq:svrg_bound_consensus_1}
    \norm{\by^{(t)}-\onet \bby^{(t)}}^2
    \le& \alpha^2 \norm{\bx^{(t-1)}-\onet \bbx^{(t-1)}}^2 \notag \\
    =& \alpha^2 \norm{\bx^{(t-1)} - \onet \by^{\mathsf{opt}}}^2 - n \alpha^2 \| \y^\opt - \bbx^{(t-1)} \|_2\notag \\
    \leq&  \alpha^2 \norm{\bx^{(t-1)}-\onet \by^{\mathsf{opt}}}^2. 
\end{align}
Then, using the strong convexity of $f(\cdot)$,
\begin{align}
    \norm{\by^{(t)}-\onet \bby^{(t)}}^2
    \leq& \alpha^2 \sum_{j=1}^n \norm{\bx_j^{(t-1)}-\by^{\mathsf{opt}}}^2 \notag \\
    \le& \frac{2\alpha^2}{\sigma}\sum_{j=1}^n \paren{f(\bx_j^{(t-1)})-f(\y^\opt)}.
    \label{eq:svrg_bound_consensus_2}
\end{align}

\subsection{Gradient estimation error}
\label{appendix:svrg_gradient_tracking}

To bound the gradient estimation error, we note that
\begin{align}
    \norm{\bs^{(t)} - \nabla f(\by^{(t)})}
    =& \norm{\WK \bs^{t-1}+\nabla F(\by^{(t)})-\nabla F(\by^{(t-1)})- \nabla f(\by^{(t)})} \notag \\
    =& \Big\|\WK \Big( \bs^{t-1} - \nabla f(\by^{(t-1)}) \Big) + \WK \nabla f(\by^{(t-1)}) -\nabla f(\by^{(t-1)}) \notag \\
    & + \nabla F(\by^{(t)})-\nabla F(\by^{(t-1)})+\nabla f(\by^{(t-1)})- \nabla f(\by^{(t)})\Big\|_2 \notag \\
    \le& \Big\| \WK \Big(\bs^{t-1}- \nabla f(\by^{(t-1)}) \Big) \Big\|_2 + \Big\| \WK \nabla f(\by^{(t-1)}) - \nabla f(\by^{(t-1)}) \Big\|_2 \notag \\
       &+ \norm{\nabla (F-f)(\by^{(t)}) + \nabla (F-f)(\by^{(t-1)})}.
    \label{eq:svrg_gradient_tracking_1}
\end{align}

We then bound the three terms in \eqref{eq:svrg_gradient_tracking_1} respectively.
\begin{enumerate}
\item The first term can be bounded as
\begin{align}
    & \norm{\WK (\bs^{t-1}- \nabla f(\by^{(t-1)}))} \notag \\
    =& \Big\| \WK \big( \bs^{t-1}- \nabla f(\by^{(t-1)}) \big) - \Big(\mean\Big) \big( \bs^{t-1}- \nabla f(\by^{(t-1)}) \big) \Big\|_2 \notag \\
    & + \Big\| \Big(\mean\Big) \Big(\bs^{t-1}- \nabla f(\by^{(t-1)}) \Big) \Big\|_2 \notag \\
    \le& \alpha \norm{\bs^{t-1}- \nabla f(\by^{(t-1)})}
    + \Big\| \Big(\mean\Big) \big(\bs^{t-1}- \nabla f(\by^{(t-1)}) \big) \Big\|_2 \notag \\
    =& \alpha \norm{\bs^{t-1}- \nabla f(\by^{(t-1)})}  + \Big\| \Big(\mean\Big) \Big(\nabla (F-f)(\by^{t-1}) - \nabla (F-f)(\by^{\mathsf{opt}}) \Big) \Big\|_2 \notag \\
    \le& \alpha \norm{\bs^{t-1}- \nabla f(\by^{(t-1)})} + \beta \norm{\by^{(t-1)}-\by^{\mathsf{opt}}},
    \label{eq:svrg_gradient_tracking_2}
\end{align}
where we used the fact $\Big\| \Big( \mean \Big) \Big\|_2 = 1$ and the definition of $\beta$ to reach the last inequality.

\item As for the second term in \eqref{eq:svrg_gradient_tracking_1}, we have
\begin{align}
    & \Big\| \WK \nabla f(\by^{(t-1)}) - \nabla f(\by^{(t-1)}) \Big\|_2 \notag \\
    \le& \Big\| \WK \nabla f(\by^{(t-1)}) - \Big( \mean \Big) \nabla f(\by^{(t-1)}) \Big\|_2 \notag\\
    & + \Big\| \Big( \mean \Big) \nabla f(\by^{(t-1)})-\nabla f(\by^{(t-1)}) \Big\|_2 \notag \\
    \le& 2\Big\| \Big( \mean \Big) \nabla f(\by^{(t-1)})-\nabla f(\by^{(t-1)}) \Big\|_2 \notag \\
    \le& 2\norm{\nabla f(\bby^{(t-1)})-\nabla f(\by^{(t-1)})} \notag  \\
    \le& 2L\norm{\by^{(t-1)}-\bby^{(t-1)}},
    \label{eq:svrg_gradient_tracking_3}
\end{align}
where the third inequality follows from the similar trick we used to obtain \eqref{eq:svrg_bound_consensus_1}.

\item Using the triangle inequality and the definition of $\beta$,
the last term in \eqref{eq:svrg_gradient_tracking_1} can be bounded by
\begin{align}
\norm{\nabla (F-f)(\by^{(t)}) + \nabla (F-f)(\by^{(t-1)})}
\leq \beta \norm{\by^{(t)}-\by^{\mathsf{opt}}} + \beta \norm{\by^{(t-1)}-\by^{\mathsf{opt}}}.
\label{eq:svrg_gradient_tracking_4}
\end{align}
\end{enumerate}

Combining \eqref{eq:svrg_gradient_tracking_1}, \eqref{eq:svrg_gradient_tracking_2},
\eqref{eq:svrg_gradient_tracking_3} and \eqref{eq:svrg_gradient_tracking_4},
the gradient estimation error can be bounded by
\begin{align}
    \norm{\bs^{(t)} - \nabla f(\by^{(t)})}
    \le& \alpha \norm{\bs^{t-1}- \nabla f(\by^{(t-1)})} + 2\beta \norm{\by^{(t-1)}-\by^{\mathsf{opt}}} \notag \\
    & + \beta \norm{\by^{(t)}-\by^{\mathsf{opt}}}+ 2L\norm{\by^{(t-1)}-\bby^{(t-1)}}. \label{eq:svrg_gradient_error}
\end{align}

Because of the strong convexity, $\norm{\by-\by^{\mathsf{opt}}}^2 \le \frac2\sigma \sum_{j=1}^n \big( f(\by_j) - f(\y^\opt) \big)$.
Combining with \eqref{eq:svrg_convergence_1},
we reached the following bound
\begin{align}\label{eq:svrg_bound_gradient}
    \norm{\bs^{(t)} - \nabla f(\by^{(t)})}^2
    \le& 4\alpha^2\norm{\bs^{t-1} - \nabla f(\by^{(t-1)})}^2 + \frac{32\beta^2}{\sigma}\sum_{j=1}^n \paren{f(\by_j^{(t-1)})-f(\y^\opt)} \notag \\
    &+ \frac{8\beta^2}{\sigma}\sum_{j=1}^n \paren{f(\bx_j^{(t-1)})-f(\y^\opt)} + 16L^2\norm{\by^{(t-1)}-\bby^{(t-1)}}^2.  
\end{align}

\subsection{Linear System}
Combining \eqref{eq:svrg_convergence_1}, \eqref{eq:svrg_bound_consensus_2}, \eqref{eq:svrg_bound_system},  and \eqref{eq:svrg_bound_gradient}, we obtain the claimed linear system.
\section{Proof of Lemma \ref{lemma:sarah_dynamic}}
\label{proof:sarah_dynamic}

Similar to the proof of Lemma \ref{lemma:svrg_dynamic}, we bound the following four terms:
    (i) Expected gradient convergence errors $\EE \norm{\nabla f(\bx^{(t)})}^2$ and $\EE \norm{\nabla f(\by^{(t)})}^2$ in Appendix~\ref{sub:gradient_convergence_error};
    (ii) Expected consensus error: $\EE \|\by^{(t)}-\onet \bby^{(t)} \|_2^2$ in Appendix~\ref{sub:consensus_error};
    (iii) Expected gradient estimation error: $\EE \| \bs^{(t)} - \nabla f(\by^{(t)}) \|_2^2$ in Appendix~\ref{sub:gradient_estimation_error}.
Then conclude the proof by creating induction inequalities.
Expectations in this section are also conditioned on $\bx^{(t-1)}$, $\by^{(t-1)}$ and $\bs^{(t-1)}$, if not specified.

\subsection{Gradient convergence error}%
\label{sub:gradient_convergence_error}

To bound the function gradient convergence error, we analyze the same helper function defined in \eqref{eq:svrg_helper_function},
where we can apply \cite[Theorem 2]{cen2019convergence} to bound the convergence error of $h^{(t,j)}(\cdot)$ as
\[
\ex{\norm{\nabla h^{(t,j)}(\bx_j^{(t)})}^2} < \nu \norm{\nabla h^{(t,j)}(\by_j^{(t)})}^2 ,
\]
where $\nu$ is the convergence rate of D-SARAH in \cite[Theorem 2]{cen2019convergence} following similar reasonings as Section~\ref{appendix:svrg_convergence}.
By setting $\delta = \frac2L \frac{1 - 8(\frac{\beta}{\sigma})^2}{9- 8 (\frac{\beta}{\sigma})^2}$
and $S = \frac{2 L}{\sigma} \frac{9- 8 (\frac{\beta}{\sigma})^2}{\big( 1 - 8(\frac{\beta}{\sigma})^2 \big)^2} $,
$\nu$ can be bounded by $\nu\leq \frac12 \frac{1}{1- 4 (\frac{\beta}{\sigma})^2}$.

Then, plugging in \eqref{eq:svrg_helper_function} and taking expectation,
we have
\begin{align*}
 \ex{\norm{\nabla f(\bx_j^{(t)})}^2}    %
    =& \ex{\norm{\nabla  h^{(t,j)}(\bx_j^{(t)}) - \bs_j^{(t)}+\nabla f(\by_j^{(t)})}^2} \\
    =& \ex{\norm{\nabla  h^{(t,j)}(\bx_j^{(t)})}^2} +\norm{\bs_j^{(t)}-\nabla f(\by_j^{(t)})}^2 - 2\ex{\innprod{\nabla  h^{(t,j)}(\bx_j^{(t)}),\bs_j^{(t)}-\nabla f(\by_j^{(t)})}} \\
    =& \ex{\norm{\nabla  h^{(t,j)}(\bx_j^{(t)})}^2} - \norm{\bs_j^{(t)}-\nabla f(\by_j^{(t)})}^2  - 2\ex{\innprod{\nabla f(\bx_j^{(t)}),\bs_j^{(t)}-\nabla f(\by_j^{(t)})}} \\
    \le& \nu \norm{\nabla f(\by_j^{(t)})-\nabla f(\by_j^{(t)})+\bs_j^{(t)}}^2 - \norm{\bs_j^{(t)}-\nabla f(\by_j^{(t)})}^2  - 2\ex{\innprod{\nabla f(\bx_j^{(t)}),\bs_j^{(t)}-\nabla f(\by_j^{(t)})}} \\
    =& \nu \norm{\nabla f(\by_j^{(t)})}^2 - 2\nu\innprod{\nabla f(\by_j^{(t)}),\bs_j^{(t)}-\nabla f(\by_j^{(t)})}  - 2\ex{\innprod{\nabla f(\bx_j^{(t)}),\bs_j^{(t)}-\nabla f(\by_j^{(t)})}},
\end{align*}
where we apply D-SARAH's convergence result in the fourth step. Summing the previous inequality over all agents, we have
\begin{align*}
 \ex{\norm{\nabla f(\bx^{(t)})}^2}
    \le& \nu \norm{\nabla f(\by^{(t)})}^2 - 2\nu{\innprod{\nabla f(\by^{(t)}),\bs^{(t)}-\nabla f(\by^{(t)})}}  - 2 \ex{\innprod{\nabla f(\bx^{(t)}),\bs^{(t)}-\nabla f(\by^{(t)})}} \\
    \le& \nu \norm{\nabla f(\by^{(t)})}^2 + 2\nu \norm{\nabla f(\by^{(t)})}\norm{\bs^{(t)}-\nabla f(\by^{(t)})}  + 2\EE \big[ \norm{\nabla f(\bx^{(t)})}\norm{\bs^{(t)}-\nabla f(\by^{(t)})} \big].
\end{align*}

Using the same method as bounding \eqref{eq:svrg_convergence_2},
\eqref{eq:svrg_convergence_3} and \eqref{eq:svrg_convergence_4},
we can prove
\begin{align*}
    2 \EE \big[ \norm{\nabla f(\bx^{(t)})}\norm{\bs^{(t)}-\nabla f(\by^{(t)})} \big]
    \le&\paren{\frac{3\beta}{\sigma} + 3\alpha\kappa} \EE\Big[\norm{\nabla f(\bx^{(t)})}^2\Big] + \frac{2\beta}{\sigma} \norm{\nabla f(\by^{(t-1)})}^2 \\
    & + \frac{\beta}{\sigma}\norm{\nabla f(\bx^{(t-1)})}^2+\frac{\alpha}{\kappa} \norm{\bs^{t-1}- \nabla f(\by^{(t-1)})}^2 \\
    &+ \frac{2L^2}{\alpha\kappa}\norm{\by^{(t-1)}-\bby^{(t-1)}}^2, \\
    2\nu \norm{\nabla f(\by^{(t)})}\norm{\bs^{(t)}-\nabla f(\by^{(t)})}
    \le& \nu\paren{\frac{4\beta}{\sigma}+3\alpha\kappa} {\norm{\nabla f(\bx^{(t-1)})}^2} + \frac{2\beta}{\sigma} \norm{\nabla f(\by^{(t-1)})}^2 \\
    & +\frac{\alpha}{\kappa} \norm{\bs^{t-1}- \nabla f(\by^{(t-1)})}^2 + \frac{2L^2}{\alpha\kappa}\norm{\by^{(t-1)}-\bby^{(t-1)}}^2.
\end{align*}

To sum up,
\begin{align}
    \ex{\norm{\nabla f(\bx^{(t)})}^2}
    \le& \left( \nu\paren{1+ \frac{4\beta}{\sigma} + 3\alpha\kappa} + \frac{\beta}{\sigma} \right) \norm{\nabla f(\bx^{(t-1)})}^2 \nonumber \\
    & + 3 \left(\frac{\beta}{\sigma} + \alpha\kappa \right) \ex{\norm{\nabla f(\bx^{(t)})}^2} + \frac{4\beta}{\sigma} \EE \norm{\nabla f(\by^{(t-1)})}^2 \nonumber \\
    & + \frac{2\alpha}{\kappa} \norm{\bs^{t-1}- \nabla f(\by^{(t-1)})}^2 + \frac{4L^2}{\alpha\kappa} \norm{\by^{(t-1)} - \onet \bby^{(t-1)}}^2. \label{eq:sarah_gradient_norm_bound_first}
\end{align}

We then show the proof for the term $ \norm{\nabla f(\by^{(t)})}^2$, which claims that the averaging process does not increase the sum of the squared norm of gradient when $\alpha\le 1/\kappa$. We denote the Hessian of the quadratic function $f(\cdot)$ by $\overline{\bH} = \nabla^2f(\cdot)$, and have
\begin{align}
\norm{\nabla f(\by^{(t)})}^2 
=& \sum_{j=1}^{n} \Big\|\nabla f(\bby^{(t)})+ \overline{\bH} ( \by_{j}^{(t)} - \bby^{(t)} ) \Big\|_2^2 \nonumber \\
\le&n\norm{\nabla f(\bby^{(t)})}^2+L^2\sum_{j=1}^{n}\norm{\by_j^{(t)}-\bby^{(t)}}^2 \nonumber\\
=&n\norm{\nabla f(\bbx^{(t-1)})}^2+L^2\norm{\WK \bx^{(t-1)} - \onet \bbx^{(t-1)}}^2 \nonumber\\
\le&n\norm{\nabla f(\bbx^{(t-1)})}^2+\alpha^2L^2\norm{\bx^{(t-1)}-\onet \bbx^{(t-1)}}^2 \nonumber\\
\le&n\norm{\nabla f(\bbx^{(t-1)})}^2+ \alpha^2 \kappa^2 \sum_{j=1}^{n}\norm{\overline{\bH} ( \bx_j^{(t-1)}- \bbx^{(t-1)}) }^2 \nonumber\\
\leq& \sum_{j=1}^{n}\Big\|\nabla f(\bbx^{(t-1)})+ \overline{\bH} ( \bx_{j}^{(t-1)} - \bbx^{(t-1)} ) \Big\|_2^2 =  \norm{\nabla f(\bx^{(t-1)})}^2. \label{eq:sarah_gradient_norm_bound}
\end{align}

\subsection{Consensus error}%
\label{sub:consensus_error}

By the property of $\bW^K$ and the strong convexity of $f$, we have
\begin{align}
\norm{\by^{(t)}-\onet \bby^{(t)}}^2 \le& \alpha^2 \norm{\bx^{(t-1)}-\onet \bbx^{(t-1)}}^2 \notag \\
\le&  \alpha^2 \norm{\bx^{(t-1)}-\onet \by^{\mathsf{opt}}}^2 \notag \\
\leq& \frac{\alpha^2}{\sigma^2} \norm{\nabla f(\bx^{(t-1)})}^2 \label{eq:sarah_last}.
\end{align}

\subsection{Gradient estimation error}%
\label{sub:gradient_estimation_error}

Note that the bound \eqref{eq:svrg_gradient_error} derived for \texttt{Network-SVRG} still holds,
combining it with \eqref{eq:sarah_last} and the strong convexity of $f$,
we have
\begin{align} \label{eq:sarah_gradient_bound}
    \norm{\bs^{(t)} - \nabla f(\by^{(t)})}^2
    \le& 4 \alpha^2 \norm{\bs^{t-1}- \nabla f(\by^{(t-1)})}^2 
    + 16\Big(\frac{\beta}{\sigma} \Big)^2 \norm{\nabla f(\by^{(t-1)})}^2 \notag \\
    & + 4 \Big(\frac{\beta}{\sigma} \Big)^2 \norm{\nabla f(\bx^{(t-1)})}^2 
    + 16L^2\norm{\by^{(t-1)}-\bby^{(t-1)}}^2. 
\end{align}

\subsection{Linear System}
Combining \eqref{eq:sarah_gradient_norm_bound_first}, \eqref{eq:sarah_gradient_norm_bound}, \eqref{eq:sarah_last}, \eqref{eq:sarah_gradient_bound}, we obtain the claimed linear system.

\end{document}